\documentclass[review]{fcs}
\usepackage{utils/common_package}
\usepackage{utils/math_commands}
\usepackage{utils/tikz_commands}
\renewcommand{\thefootnote}{\fnsymbol{footnote}}

\title{A Survey of Geometric Graph Neural Networks: Data Structures, Models and Applications}
\author[1,2]{Jiaqi HAN\footnotemark[1]}
\author[1]{Jiacheng CEN\footnotemark[1]}
\author[1]{Liming WU\footnotemark[1]}
\author[1]{Zongzhao LI}
\author[3]{Xiangzhe KONG}
\author[3]{Rui JIAO}
\author[3]{Ziyang YU}
\author[4,5]{Tingyang XU}
\author[6]{Fandi WU}
\author[1]{Zihe WANG}
\author[1]{Hongteng XU}
\author[1]{Zhewei WEI}
\author[4,5]{Deli ZHAO}
\author[3]{Yang LIU\footnotemark[2]}
\author[4,5]{Yu RONG\footnotemark[2]}
\author[1]{Wenbing HUANG\footnotemark[2]}

\address[1]{Gaoling School of Artificial Intelligence, Renmin University of China, Beijing 100872, China.}
\address[2]{Department of Computer Science, Stanford University, CA 94305, USA.}
\address[3]{Department of Computer Science and Technology, Tsinghua University, Beijing 100084, China.}
\address[4]{DAMO Academy, Alibaba Group, Hangzhou 311121, China.}
\address[5]{Hupan Lab, Hangzhou 311121, China.}
\address[6]{Tencent AI Lab, Shenzhen 518100, China.}

\fcssetup{
  received       = {December 28, 2024},
  accepted       = {February 24, 2025},
  corr-email     = {liuyang2011@tsinghua.edu.cn;\\ yu.rong@hotmail.com; hwenbing@126.com.},
  issue = {19},
  volume = {11},
  doi  = {10.1007/s11704-025-41426-w},
}
\begin{abstract}
Geometric graphs are a special kind of graph with geometric features, which are vital to model many scientific problems. Unlike generic graphs, geometric graphs often exhibit physical symmetries of translations, rotations, and reflections, making them ineffectively processed by current Graph Neural Networks (GNNs). To address this issue, researchers proposed a variety of geometric GNNs equipped with invariant/equivariant properties to better characterize the geometry and topology of geometric graphs. Given the current progress in this field, it is imperative to conduct a comprehensive survey of data structures, models, and applications related to geometric GNNs. In this paper, based on the necessary but concise mathematical preliminaries, we formalize geometric graph as the data structure, on top of which we provide a unified view of existing models from the geometric message passing perspective. Additionally, we summarize the applications as well as the related datasets to facilitate later research for methodology development and experimental evaluation. We also discuss the challenges and future potential directions of geometric GNNs at the end of this survey. 
\end{abstract}

\begin{document}
\renewcommand{\thefootnote}{\fnsymbol{footnote}}
\footnotetext[1]{Authors contribute equally. Work done by Jiaqi Han during his visit to Renmin University of China.}
\footnotetext[2]{Corresponding Authors: Wenbing Huang, Yang Liu, Yu Rong.}
\renewcommand{\thefootnote}{\arabic{footnote}}

\section{Introduction}\label{sec:introduction}

Many scientific problems particularly in physics and biochemistry require to process data in the form of \emph{geometric graphs}~\citep{bronstein2021geometric}. Distinct from typical graph data, geometric graphs additionally assign each node a special type of node feature in the form of geometric vectors. For example, a molecule/protein can be regarded as a geometric graph, where the 3D position coordinates of atoms are the geometric vectors; in a general multi-body physical system, the 3D states (positions, velocities or spins) are the geometric vectors of the particles. Notably, geometric graphs exhibit symmetries of translations, rotations and/or reflections. This is because the physical law controlling the dynamics of the atoms (or particles) is the same no matter how we translate or rotate  the  physical system from one place to another.  When tackling this type of data, it is essential to incorporate the inductive bias of symmetry into the design of the model, which motivates the study of geometric Graph Neural Networks (GNNs).

\begin{figure}
    \centering
    \includegraphics[width=\linewidth]{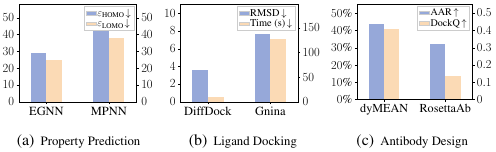}
    \caption{Performance comparisons between geometric GNNs and traditional methods on molecular property prediction, protein-ligand docking, and antibody design. Notably, the methods based on geometric GNNs, including EGNN~\citep{satorras2021en}, DiffDock~\citep{corso2023diffdock}, and dyMEAN~\citep{kong2023end}, remarkably outperform traditional MPNN~\citep{gilmer2017neural}, Gnina~\citep{mcnutt2021gnina}, and RossetaAb~\citep{adolf2018rosettaantibodydesign}, on the datasets of QM9~\citep{ramakrishnan2014quantum}, PDBBind~\citep{liu2017forging}, and SAbDab~\citep{dunbar2014sabdab}, respectively, verifying the effectiveness and efficiency of geometric GNNs over various tasks.}
    \label{fig:performance}
\end{figure}

Constructing GNNs that permit such symmetry constraints has long been challenging to methodological design. Pioneer approaches like DTNN~\citep{schutt2017quantum}, DimeNet~\citep{Klicpera2020Directional} and GemNet~\citep{klicpera2021gemnet}, transform the input geometric graph into distance/angle/dihedral-based scalars that are invariant to rotations or translations, constituting the family of invariant GNNs. Noticing the limit on the expressivity of invariant GNNs, EGNN~\citep{satorras2021en} and PaiNN~\citep{schutt2021equivariant} additionally involve geometric vectors in message passing and node update to preserve the directional information in each layer, leading to equivariant GNNs. With group representation theory as a helpful tool, TFN~\citep{thomas2018tensor}, SE(3)-Transformer~\citep{fuchs2020se} and SEGNN~\citep{brandstetter2022geometric} generalizes  invariant scalars and equivariant vectors by viewing them as steerable vectors parameterized by high-degree spherical tensors, giving rise to high-degree steerable GNNs. Built upon these fundamental approaches, geometric GNNs have made remarkable success in various applications of diverse systems, including physical dynamics simulation~\citep{fuchs2020se,satorras2021en}, molecular property prediction~\citep{batzner20223,liao2023equiformer}, protein structure prediction~\citep{minkyung2021rosettafold}, protein generation~\citep{watson2023novo,ingraham2023illuminating}, and RNA structure ranking~\citep{raphael2021geometric}. \cref{fig:performance} illustrates the superior performance of geometric GNNs against traditional methods on the representative tasks.

To facilitate the research of geometric GNNs, this work presents a systematic survey focusing both on the methods and applications\footnote{This work is an extended survey of our previous short version~\citep{han2022geometrically}.}, which is structured as the following sections: In~\cref{sec:basics}, we introduce necessary preliminaries on group theory and the formal definition of equivariance/invariance; In~\cref{sec:structure}, we propose geometric graph as a universal data structure that will be leveraged throughout the entire survey as a bridge between real-world data and the models, \emph{i.e.}, geometric GNNs; In~\cref{sec:geomgnn}, we summarize existing models into invariant GNNs (\cref{sec:Invariant Graph Neural Networks}) and equivariant GNNs (\cref{sec:Equivariant Graph Neural Networks}), while the latter is further categorized into scalarization-based models (\cref{sec:scalarization}) and high-degree steerable models (\cref{sec:high-order}); Besides, we also introduce geometric graph transformers in~\cref{sec:transformers}; In~\cref{sec:app}, we provide a comprehensive collection of the applications that have witnessed the success of geometric GNNs on particle-based physical systems, molecules, proteins,  complexes, and other domains like crystals and RNAs.

\begin{figure*}[!t]
    \centering
    \includegraphics[width=\textwidth]{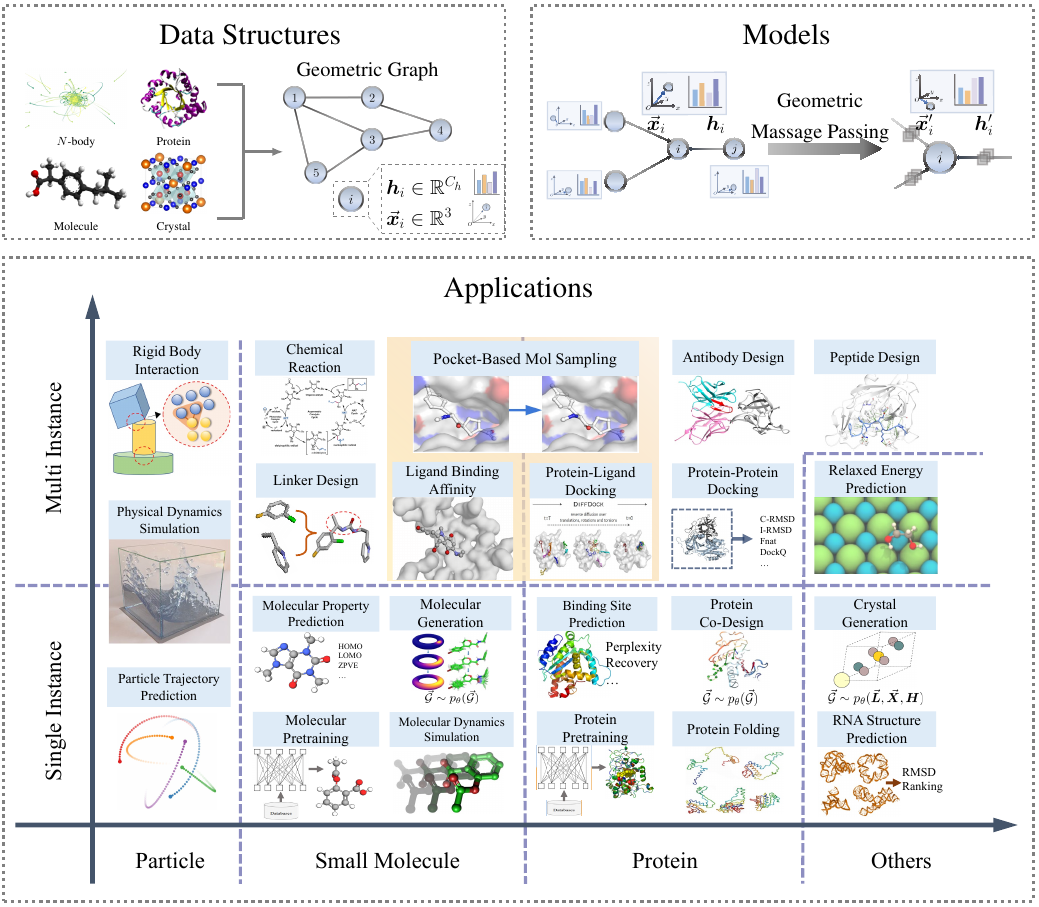}
    \caption{Illustration of the complete input-output pipeline from data structures, models to applications. Note that most figures here for illustrating different applications are edited based on previous papers~\citep{han2022learning,sanchez2020learning,kipf2018neural,huang20223dlinker,guan2023d,corso2023diffdock,jing2022torsional,wu2023equivariant,kong2023conditional,senior2020improved,ocp_dataset,raphael2021geometric,kong2024full}. The term ``instance'' indicates a self-interacted system composed of multiple particles/atoms, such as a molecule or a protein. Pocket-Based Molecule Sampling, Ligand-Binding Affinity Prediction, and Protein-Ligand Docking are denoted with yellow shading to imply that all these tasks take as the multi-instance format ``Molecule+Protein". }
    \label{fig:title}
\end{figure*}

The goal of this survey is to provide a general overview throughout data structure, model design, and applications (see \cref{fig:title}), which constitutes an entire input-output pipeline that is instructive for machine learning practitioners to employ geometric GNNs on various scientific tasks. 
Recently, several related surveys have been proposed, which place main focus on methodology of geometric GNNs~\citep{duval2023hitchhiker}, pretrained GNNs for chemical data~\citep{xia2023systematic}, representation learning for molecules~\citep{guo2023graph,atz2021geometric}, and general application of artificial intelligence in diverse types of scientific systems~\citep{zhang2023artificial}. In contrast to all of them, this survey places an emphasis on geometric graph neural networks, not only encapsulating theoretical foundations of geometric GNNs but also delivering an exhaustive summary of the related applications in domains across physics, biochemistry, and material science. Meanwhile, we discuss future prospects and interesting research directions in~\cref{sec:fp}. We also release the Github repository that collects  the reference, datasets, codes, benchmarks, and other resources related to geometric GNNs. 

\section{The Basic Notion of Symmetry}
\label{sec:basics}
In this section, we will compactly introduce the basic notions related to symmetry. Readers can skip this section and get straight to the methodology part in \cref{sec:structure} if they are familiar with the theoretical background.

\subsection{Transformation and Group}

By defining symmetry, we indicate that an object of interest keeps invariant under a set of transformations. For instance, the distance between any two points in space remains constant, regardless of how we simultaneously rotate or translate these two points. Mathematically, a set of transformations forms a \emph{group} (more details are referred to~\citep{esteves2020theoretical}).

\begin{definition}[Group]
A group $G$ is a set of transformations with a binary operation ``$\cdot$'' satisfying these properties: 
(i) it is closed, namely, $\forall a,b\in G, a\cdot b \in G$;
(ii) it is associative, namely, $\forall a, b, c\in G, (a\cdot b)\cdot c = a \cdot (b\cdot c)$;
(iii) there exists an identity element $e\in G$ such that $\forall a\in G, a\cdot e = e\cdot a=a$;
(iv) each element must have an inverse, namely, $\forall a\in G, \exists b\in G, a\cdot b=b\cdot a =e$, where the inverse $b$ is denoted as $a^{-1}$. 
\end{definition}

We below provide some examples commonly used in the applications  of this paper:
\begin{itemize}[leftmargin=*]
    \item $\mathrm{E}(d)$ is an Euclidean group~\citep{cederberg2004course} consisting of rotations, reflections and translations, acting on $d$-dimension vectors.
    \item $\mathrm{T}(d)$ is a subgroup of Euclidean group that consists of translations. 
    \item $\mathrm{O}(d)$ is an orthogonal group that consists of rotations and reflections, acting on $d$-dimension vectors.
    \item $\mathrm{SO}(d)$ is a special orthogonal group that only consists of rotations.
    \item $\mathrm{SE}(d)$ is a special Euclidean group that consists of only rotations and translations.
    \item Lie Group is a group whose elements form a differentiable manifold. Actually, all the groups above are specific examples of Lie Group.
    \item $\mathrm{S}_N$ is a permutation group whose elements are permutations of a given set consisting of $N$ elements.
\end{itemize}

\subsection{Group Representation}
While the group operation ``$\cdot$'' is defined abstractly above, it can be realized as matrix multiplication, with the help of group representation. A representation of $G$ is a group homomorphism $\rho(g): G \mapsto \mathrm{GL}(\gV)$ that takes as input the group element $g\in G$ and acts on the general linear group of some vector space $\gV$, satisfying $\rho(g)\rho(h) = \rho(g\cdot h), \forall g, h\in G$. When $\gV=\R^d$, then $\mathrm{GL}(\gV)$ contains all invertible $d\times d$ matrices and $\rho(g)$ assigns a matrix to the element $g$. 

For the orthogonal group $\mathrm{O}(d)$, one of its common group representations is defined by orthogonal matrices $\mO\in\R^{d\times d}$ subject to $\mO^{\top}\mO = \mI$; for $\mathrm{SO}(d)$, the group representation is restricted to orthogonal matrices of determinant 1, denoted as $\mR$. The case of translation group $\mathrm{T}(d)$ is a bit tedious and can be derived in the projective space using homogeneous coordinates; here, for simplicity, we directly define translation as vector addition other than matrix multiplication. Note that the representation of a group is not unique, which will be further illustrated in \cref{sec:high-order}.  

\subsection{Equivariance and Invariance}

Let $\gX$ and $\gY$ be the input and output vector spaces, respectively. The function $\phi: \gX\rightarrow\gY$ is called equivariant with respect to $G$ if when we apply any transformation to the input, the output also changes via the same transformation or under a certain predictable behavior. In form, we have:
\begin{definition}[Equivariance]
The function $\phi: \gX\mapsto\gY$ is $G$-equivariant if it commutes with any transformation in $G$,
\begin{eqnarray}
\label{eq:equ}
\phi (g\cdot x) = g\cdot\phi( x), \forall g\in G,
\end{eqnarray}
which, by implementing the group operation $\cdot$ with group representation, can be rewritten as:
\begin{eqnarray}
\label{eq:eqwithrepresetation}
\phi (\rho_{\gX}(g) x) = \rho_{\gY}(g)\phi( x), \forall g\in G,
\end{eqnarray}
where $\rho_{\gX}$ and $\rho_{\gY}$ are the group representations in the input and output space, respectively.
\end{definition}

The choice of group representation facilitates the specialization of different scenarios. When both  $\rho_{\gX}$ and $\rho_{\gY}$ are trivial representations, namely, $\rho_{\gX}(g)=\rho_{\gY}(g)=\mI$,\footnote{Note that the identity transformation $\mI$ could have different dimensions in the input space $\gX$ and output space $\gY$.} $\phi$ becomes a trivial function; notably, when $\rho_{\gY}(g)=\mI$, $\phi$ is called \emph{an invariant function}, demonstrating that invariance is just a special case of equivariance.

It is able to verify that equivariance induces the following desirable properties. (i) \textbf{Linearity:} any linear combination of equivariant functions is still equivariant. (ii) \textbf{Composability:} the composition of two equivariant functions (if they can be composed) yields an equivariant function. Therefore, equivariance for each layer of a network implies that a whole network is equivariant. (iii) \textbf{Inheritability:} if a function is equivariant with respect to group ${G}_1$ and group ${G}_2$, then this function must be equivariant with respect to the direct product of these two groups, \emph{i.e.} ${G}_1\times{G}_2$ under a corresponding definition of product group operation or group representation. This implies that proving equivariance of each transformation individually is sufficient to prove equivariance of joint transformations.

In the following context, the variable $x$ is instantiated as a geometric graph, the group transformation $\rho(g)$ becomes the transformation of geometric graphs, and the function $\phi$ is designed as an invariant/equivariant GNN.

\section{Data Structure: from Graph to Geometric Graph}
\label{sec:structure}

This section formally defines graph and geometric graph, and depicts how they differ from each other. \cref{tab:notations} summarizes the notations we used throughout this paper. 

\begin{table}[!t]
\caption{Basic notations and definitions throughout this survey.}
\vspace{2pt}
\label{tab:notations}
\renewcommand\arraystretch{1.2}
\resizebox{\linewidth}{!}{
\begin{tabular}{cm{0.8\linewidth}<{\raggedright}}
    \toprule
    \textbf{Notation} & \textbf{Description}\\
    \midrule
    \addlinespace[0.08pt]
    \rowcolor{gray!20}\multicolumn{2}{c}{Data Structure}\\
    \addlinespace[0.08pt]\midrule
    $\gG\coloneqq(\mA, \mH)$ & A graph $\gG$ containing $N$ nodes, with adjacency matrix $\mA\in\R^{N\times N}$ and node feature matrix $\mH\in\R^{N \times C_h}$.\\ \hline
    $\vec{\gG}\coloneqq(\mA, \mH, \vec{\mX})$ & A geometric graph $\vec{\gG}$ containing $N$ nodes, with adjacency matrix $\mA$ and node feature matrix $\mH$ as above, and additionally a 3D coordinate matrix $\vec{\mX}\in\R^{N \times 3}$.\\ \hline
    $\gN_i$   & The neighborhood of node $i$. \\   \hline 
    $\vh_i\in\R^{C_h}$ & The scalar feature of node $i$.\\ \hline
    $\vec{\vx}_i\in\R^3$ & The 3D coordinate of node $i$.\\ \hline
    $\vec{\mV}_i\in\R^{3\times C}$ & The multi-channel 3D vector of node $i$.\\   \hline 
    $\vec{\mV}_i^{(l)}\in\R^{(2l+1)\times C_l}$ & The type-$l$ irreducible vector of node $i$.\\ \hline
    $\vec{\sV}_i^{(\sL)}\coloneqq\{\vec{\mV}_i^{(l)}\}_{l\in\sL}$ 
    & The set consisting of irreducible vectors of all types $l\in\sL$. \\ \hline    
    $\ve_{ij}\in\R^{C_e}$ 
    & The edge feature from node $j$ to $i$.\\
    \midrule
    \addlinespace[0.08pt]
    \rowcolor{gray!20}\multicolumn{2}{c}{Operator}\\
    \addlinespace[0.08pt]\midrule
    $G, g$ & The group $G$ and its group element $g$.\\ \hline
    $\rho_{\gX}(g)$ & The {group representation} $\rho_{\gX}(g)$ of the transformation $g$ in the vector space $\gX$.\\ \hline
    $\times, \otimes$ & The operators between two vectors including cross product $\times$ and Kronecker product $\otimes$\\ \hline
    $\otimes_{\text{cg}}, \otimes_{\text{cg}}^{\mW}, \otimes_{\text{cg}}^{\sW}$ & {Clebsch-Gordan (CG) tensor product}, optionally with a {learnable parameter} $\mW$ and a {learnable parameter set} $\sW$.\\ \hline
    $Y^{(l)}(\vec{\vx})\in\R^{2l+1}$ & The type-$l$ vector constructed by spherical harmonics of $\vec{\vx}\in S^2$: $Y^{(l)}(\vec{\vx})=[Y^{(l)}_{-l},Y^{(l)}_{-l+1},\cdots, Y^{(l)}_{l-1}, Y^{(l)}_{l}]$.\\ \hline
    $\sY^{(\sL)}(\vec{\vx})\coloneqq \{Y^{(l)}(\vec{\vx})\}_{l\in\sL}$ & A set consisting of spherical harmonics of all types $l\in\sL$.\\ \hline
    $\mD^{(l)}(g)$ & The $l$-th degree Wigner-D matrix of the rotation transformation $g\in\mathrm{SO}(3)$.\\ 
    \midrule
    \addlinespace[0.08pt]
    \rowcolor{gray!20}\multicolumn{2}{c}{Neural Network}\\
    \addlinespace[0.08pt]\midrule
    $\phi, \psi, \varphi, \sigma$ & Functions implemented with MLP.\\   
    \bottomrule
\end{tabular}
}
\renewcommand\arraystretch{1}
\end{table}

\subsection{Graph}
\label{sec:graph}
Conventional studies on graphs~\citep{wu2020comprehensive,yuan2024index} usually focus on their relational topology. Examples include social networks, citation networks, etc. In the domain of AI-Driven Drug Design~(AIDD), they are usually referred to as 2D graphs~\citep{C7SC02664A}.

\begin{definition}[Graph]
    \label{def:graph}
    A graph is defined as $\gG\coloneqq(\mA, \mH)$ where $\mA\in[0,1]^{N\times N}$ is the adjacency matrix with $N$ being the number of nodes, and $\mH\in\sR^{N \times C_h}$ is the node feature matrix with $C_h$ being the dimension of the feature. 
\end{definition}
Concretely, the adjacency matrix $\mA$ takes the value 1 at its $(i,j)$-th entry $a_{ij}$ when node $i$ and $j$ are connected by an edge and $0$ otherwise. The $i$-th row of $\mH$, \emph{i.e.}, $\vh_i\in\sR^{C_h}$, represents the feature vector for node $i$, \emph{e.g.}, the one-hot embedding of the atomic number in a molecule graph.
Along with the definition of graph, we also describe some vital concepts derived. We denote the set of nodes as $\gV$ and the set of edges as $\gE$.
Correspondingly, the neighborhood of node $i$, marked as $\gN_i$, is specified to be $\gN_i\coloneqq\{j: (v_i, v_j)\in\gE\}$. The graph can additionally contain some edge features $\ve_{ij}\in\sR^{C_e}$ for edge $(v_i, v_j)$. 

\textbf{Transformations on graphs: $g\cdot\gG$.} One can arbitrarily change the order of nodes without changing the topology of the graph. With the language of group representation, the permutation transformation of a graph is denoted as $g\cdot\gG\coloneqq(\mP_g\mA\mP_g^\top, \mP_g\mH)$, where $\mP_g$ is the representation of the transformation $g\in \text{S}_N$ (\emph{i.e.} the permutation matrix\footnote{The permutation of $\mA$ can also be written in the form of group representation by first vectorizing $\mA$ as $\mathrm{Vec}(\mA)$ and then conducting $(\mP_g\otimes\mP_g)\mathrm{Vec}(\mA)$. Here $\otimes$ defines the Kronecker product, and $\mP_g\otimes\mP_g$ is the 2-order representation of the permutation matrix.}). We denote the equivalence in terms of permutation as $\gG\simeq g\cdot\gG$.

As a concrete example, molecules can be viewed as graphs, where the nodes $v_i$ are instantiated as the atoms, and the node features $\mH$ are the one-hot encoding of the atomic numbers, a row for each atom. The edges $\mA$ are either the existence of chemical bonds or constructed based on relative distance between atoms under a cut-off threshold, and the respective edge features $\ve_{ij}$ can be assigned as the type of the chemical bond and/or the relative distance. 

\subsection{Geometric Graph}
\label{sec:geograph}

In many applications, the graphs we tackle contain not only the topological connections and node features, but also certain \emph{geometric information}. Again, in the example of a molecule, we may additionally be informed of some geometric quantities in the Euclidean space, \emph{e.g.}, the positions of the atoms in 3D coordinates\footnote{Although we mainly focus on 3D space, most of our analyses can be extended to $d$-dimensional space where $d$ is an arbitrary integer.}. Such quantities are of particular interest in that they encapsulate rich directional information that depicts the geometry of the system. With the geometric information, one can go beyond working on limited perception of the graph topology, but instead to a broader picture of the entire configuration of the system in 3D space, where important information, such as the relative orientation of the neighboring nodes and directional quantities like velocities, could be better exploited. Hence, in this section, we begin with the definition of geometric graphs, which are usually referred to as 3D graphs~\citep{bronstein2021geometric}.

\begin{definition}[Geometric Graph]
\label{def:gg}
    A geometric graph is defined as $\vec{\gG}\coloneqq(\mA, \mH, \vec{\mX})$, where $\mA\in[0,1]^{N \times N}$ is the adjacency matrix, $\mH\in\sR^{N \times C_h}$ is the node feature matrix with dimension $C_h$, and $\vec{\mX}\in\sR^{N \times 3}$ are the 3D coordinates of all nodes.
\end{definition}

The $i$-th rows of $\mH$ and $\vec{\mX}$, namely, $\vh_i\in\R^{C_h}$ and $\vx_i\in\R^3$ denote the feature and 3D coordinate of node $v_i$, respectively. 
In the above definition, we distinguish the coordinate matrix $\vec{\mX}$ from other quantities $\mA$ and $\mH$, and geometric graph $\vec{\gG}$ from graph $\gG$, with an over-right arrow ``$\rightarrow$'', indicating that they contain geometric and directional information. Note that there could be other geometric variables besides $\vec{\mX}$ in a geometric graph, such as velocity, force, and so on. Then the shape of $\vec{\mX}$ is extended 
from $N\times 3$ to $N\times 3\times C$
where $C$ denotes the number of channels. In this section, we assume $C=1$ for conciseness, while more complete examples are shown in \cref{sec:app}.

\textbf{Transformations on geometric graphs: $g\cdot\vec{\gG}$.} In contrast to graphs, transformations on geometric graphs are not limited to node permutation. We summarize the transformations of interest below:
\begin{itemize}
    \item \textbf{Permutation}, which is defined as $g\cdot\vec{\gG}\coloneqq(\mP_{g}\mA\mP_g^\top, \mP_g\mH,\mP_g\vec{\mX})$, where $\mP_g$ is the permutation matrix representation of $g\in \mathrm{S}_n$;
    \item \textbf{Orthogonal transformation}, which is defined as $g\cdot\vec{\gG}\coloneqq(\mA,\mH, \vec{\mX}\mO_g)$, where $\mO_g$ is the orthogonal matrix representation of $g\in\mathrm{O}(3)$, consisting of rotations and reflections;
    \item \textbf{Translation}, which is defined as $g\cdot\vec{\gG}\coloneqq(\mA,\mH,\vec{\mX} + \vec{\vt}_g)$, where $\vec{\vt}_g$ is the translation vector of $g\in\mathrm{T}(3)$;
\end{itemize}

\begin{figure}[htbp]
    \centering
    \includegraphics[width=\linewidth]{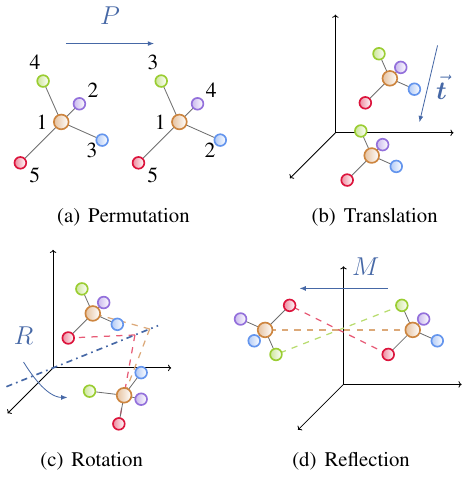}
    \caption{Examples of transformations on geometric graphs.}
    \label{fig:transformation}
\end{figure}

We always have the equivalence $\vec{\gG}\simeq g\cdot\vec{\gG}$. We can combine orthogonal transformation and translation into Euclidean transformation on geometric graphs, namely, $g\cdot\vec{\gG}\coloneqq(\mA,\mH, \vec{\mX}\mO_g + \vec{\vt}_g)$ for $g\in\mathrm{E}(3)$. Here, the Euclidean group $\mathrm{E}(3)$ is a semidirect product~\citep{villar2021scalars} of orthogonal transformation and translation, denoted as $\mathrm{E}(3)=\mathrm{T}(3)\ltimes\mathrm{O}(3)$. We can similarly define $\mathrm{SE}(3)$ transformation by considering only rotation and translation. 
We sometimes call $\mH$ invariant features (or scalars), since they are independent to $\mathrm{E}(3)$ transformation, and call $\vec{\mX}$ equivariant features (or vectors) that correlate to $\mathrm{E}(3)$ transformations. \cref{fig:transformation} demonstrates the example of transformation on geometric graph.

Geometric graphs are powerful and general tools to model a variety of objects in scientific tasks, including small molecules~\citep{schutt2018schnet,satorras2021en}, proteins~\citep{baek2023efficient,ingraham2023illuminating}, crystals~\citep{luo2023towards,jiao2023crystal}, physical point clouds~\citep{huang2022equivariant,han2022learning}, and many others. 

We will provide more details in \cref{sec:app}.

\section{Model: Geometric GNNs}
\label{sec:geomgnn}

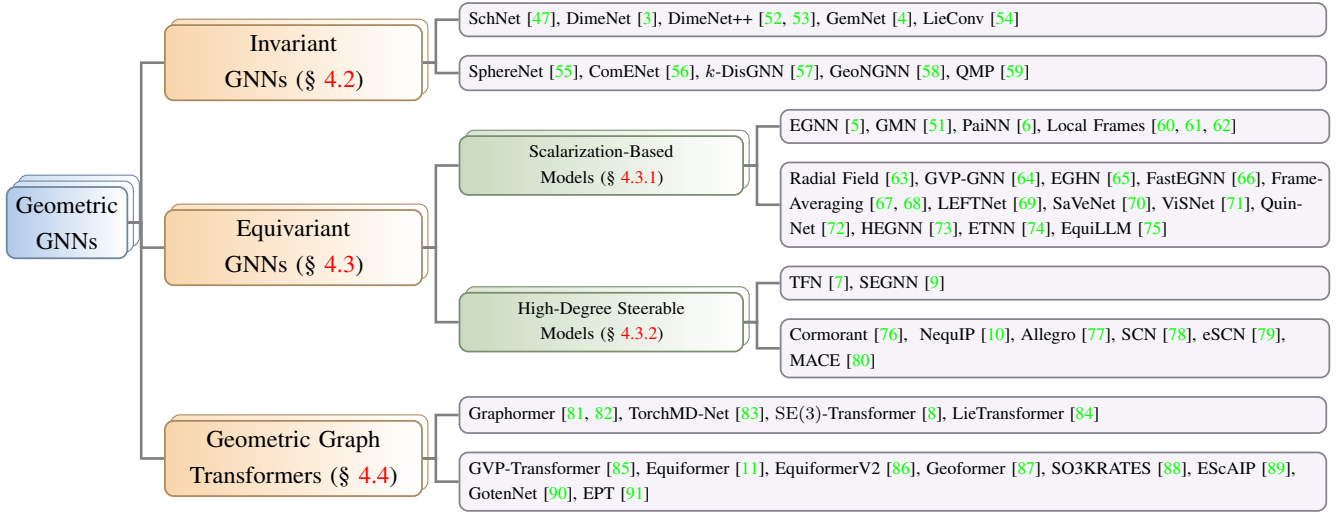
\begin{figure*}[!t]
    \centering
    \adjustbox{width=1\textwidth}{
\begin{forest}
for tree={
    forked edges,
    grow          = east,
    reversed      = true,
    anchor        = base west,
    parent anchor = east,
    child anchor  = west,
    base          = middle,
    font          = \normalsize,
    minimum width = 1em,
},
[Geometric \\ GNNs, TA-Layer-1,
    [\footnotesize{Invariant \\ GNNs} (\cref{sec:Invariant Graph Neural Networks}), TA-Layer-2,
        [SchNet~\citep{schutt2018schnet}{, }DimeNet~\citep{Klicpera2020Directional}{, }DimeNet++~\citep{gasteiger2020fast,zhu2023fastdimenet++}{, }GemNet~\citep{klicpera2021gemnet}{, }LieConv~\citep{finzi2020generalizing}, Leaf3]
        [SphereNet~\citep{liu2022spherical}{, }ComENet~\citep{wang2022comenet}{, }$k$-DisGNN~\citep{li2024distance}{, }GeoNGNN~\citep{li2024completeness}{, }QMP~\citep{yue2024a}, Leaf3]
    ]
    [\footnotesize{Equivariant \\ GNNs} (\cref{sec:Equivariant Graph Neural Networks}), TA-Layer-2
        [Scalarization-Based\\ Models (\cref{sec:scalarization}), TA-Layer-3
            [EGNN~\citep{satorras2021en}{, }GMN~\citep{huang2022equivariant}{, }PaiNN~\citep{schutt2021equivariant}{, }Local Frames~\citep{du2022se,kofinas2021rototranslated,kofinas2023latent}, Leaf4]
            [Radial Field~\citep{kohler2019equivariant}{, }GVP-GNN~\citep{jing2021learning}{, }EGHN~\citep{han2022equivariant}{, }FastEGNN~\citep{zhang2024improving}{, }Frame-Averaging~\citep{puny2021frame,duval2023faenet}{, }LEFTNet~\citep{du2024new}{, }SaVeNet~\citep{aykent2023savenet}{, }ViSNet~\citep{wang2024enhancing}{, }QuinNet~\citep{wang2023efficiently}{, }HEGNN~\citep{cen2024high}{, }ETNN~\citep{battiloro2025etnn}{, }EquiLLM~\citep{li2025large},Leaf4]
        ]
        [High-Degree Steerable\\ Models (\cref{sec:high-order}), TA-Layer-3
            [TFN~\citep{thomas2018tensor}{, }SEGNN~\citep{brandstetter2022geometric}, Leaf4]
            [Cormorant~\citep{anderson2019cormorant}{, } NequIP~\citep{batzner20223}{, }Allegro~\citep{musaelian2023learning}{, }SCN~\citep{zitnick2022spherical}{, }eSCN~\citep{passaro2023reducing}{, }MACE~\citep{batatia2022mace}, Leaf4]
        ]
    ]
    [\footnotesize{Geometric Graph\\ Transformers} (\cref{sec:transformers}), TA-Layer-2
        [Graphormer~\citep{ying2021transformers,shi2022benchmarking}{, }TorchMD-Net~\citep{tholke2022equivariant}{, }$\mathrm{SE}(3)$-Transformer~\citep{fuchs2020se}{, }LieTransformer~\citep{hutchinson2021lietransformer}, Leaf3]
        [GVP-Transformer~\citep{hsu2022learning}{, }Equiformer~\citep{liao2023equiformer}{, }EquiformerV2~\citep{liao2023equiformerv2}{, }Geoformer~\citep{wang2024geometric}{, }SO3KRATES~\citep{frank2024euclidean}{, }GotenNet~\citep{aykent2025rethinking}{, }EPT~\citep{jiao2024equivariant}, Leaf3]
    ]
]
\end{forest}
}
    \caption{The taxonomy of geometric GNNs introduced in~\cref{sec:geomgnn}.}
    \label{fig:tree_of_models}
\end{figure*}

In this section, we first recap the general form of Message Passing Neural Network (MPNN) on topological graphs. Then we introduce different types of geometric GNNs that extends the message passing paragidm of MPNNs to geometric graphs: invariant GNNs, equivariant GNNs, as well as geometric graph transformers. Finally, we briefly present the works that discuss the expressivity of geometric GNNs. \cref{fig:tree_of_models} presents the taxonomy of geometric GNNs in this section.

\subsection{Message Passing Neural Networks}
\label{sec:MPNN}
Graph Neural Networks (GNNs) are favorable to operate on graphs with the help of the message-passing mechanism, which facilitates the information propagation along the graph structure by updating node embeddings through neighborhood aggregation. To be specific, message-passing GNNs implement $\phi(\gG)$ on topological graphs $\gG$ by iterating the following message-passing process in each layer~\citep{gilmer2017neural},
\begin{align}
\label{eq:message}
    \vm_{ij} &= \phi_{\text{msg}}\left(\vh_i, \vh_j, \ve_{ij} \right), \\
\label{eq:update}
    \vh_i' &= \phi_{\text{upd}}\left( \vh_i, \{ \vm_{ij} \}_{j\in\gN_i} \right), 
\end{align}
where $\phi_{\text{msg}}\left(\cdot\right)$ and $\phi_{\text{upd}}\left(\cdot\right)$ are the message computation and feature update function, respectively. The node features $\vh_i,\vh_j$ and edge feature $\ve_{ij}$ is first synthesized by the message function to obtain the message $\vm_{ij}$. The messages within the neighborhood are then aggregated with one set function and leveraged to update the node features $\vh'_i$ combined with the input $\vh_i$.

GNNs defined by \cref{eq:message,eq:update} are always permutation equivariant but not inherently $\mathrm{E}(3)$-equivariant. 
When mentioning equivariance or invariance in what follows, this paper mainly discusses the latter unless otherwise specified.

\subsection{Invariant Graph Neural Networks}
\label{sec:Invariant Graph Neural Networks}
Moving forward to the geometric domain, there are various tasks that require the model we propose to be \emph{invariant} with regard to Euclidean transformations. For instance, for the task of molecular property prediction, the predicted energy should remain unchanged regardless of any rotation/translation of all atom coordinates. Embedding such inductive bias is crucial as it essentially conforms to the physical rule of our 3D world. 

In form, invariant GNNs update invariant features as $\mH' = \phi(\vec{\gG})$ with the function $\phi$ satisfying:
\begin{align}
    \phi(g\cdot\vec{\gG}) = \phi(\vec{\gG}), \forall g\in\mathrm{E}(3).
\end{align}
To design such function, invariant GNNs usually transform equivariant coordinates $\vec{\mX}$ to invariant scalars that are unaffected by Euclidean transformations. Early invariant GNNs can date back to DTNN~\citep{schutt2017quantum}, MPNN~\citep{gilmer2017neural} and MV-GNN~\citep{10.1093/bioinformatics/btac039}, where relative distances are applied for edge construction. Recent works further elaborate the use of various invariant scalars ranging from relative distances to angles or dihedral angles between edges, upon the message passing mechanism in \cref{eq:message,eq:update}. We introduce several representative works below.    

\textbf{SchNet}~\citep{schutt2018schnet}. This work designs a continues filter convolution conditional on relative distances $r_{ij}=\|\vec{\vx}_i-\vec{\vx}_j\|$. In particular, it re-implements \cref{eq:message} as 
\begin{align}
\vm_{ij} &=\sigma_{2}(r_{ij})\sigma_1(\vh_j),    
\end{align}
where the message is calculated as the multiplication between the continues convolution filter and the neighbor embedding, and the functions $\sigma$ are all Multi-Layer Perceptrons (MLPs). 

\textbf{DimeNet}~\citep{Klicpera2020Directional}. By observing that using relative distances alone is unable to encode directional information, DimeNet proposes directional message passing which takes as input not only relative distances but also angles between adjacent edges. The main component to compute the message embedding of each directional edge (from $j$ to $i$) is given by:
\begin{align}
\label{eq:dimenet}
    \vm'_{ji} &=\sigma_{\mathrm{msg}}(\vm_{ji}, \sum_{k\in\gN_j\backslash\{i\}}\sigma_{\mathrm{int}}(\vm_{kj},\ve_{\mathrm{RBF}}^{(ji)},\ve_{\mathrm{CBF}}^{(kji)})),  
\end{align}
where $\ve_{\mathrm{RBF}}^{(ji)}$ denotes the radial basis function representation of relative distance $d_{ji}$; $\ve_{\mathrm{CBF}}^{kji}$\footnote{Here CBF is short for Circular Bessel Function.} computes the joint representation of relative distance $d_{kj}$ and angle $\alpha_{(kj,ji)}$ between edge $(v_k,v_j)$ and $(v_j,v_i)$, with the help of spherical Bessel functions and spherical harmonics. In~\citep{Klicpera2020Directional}, \cref{eq:dimenet} is applied as an interaction block before an embedding block that derives the message $\vm_{ji}$ based on $\ve_{\mathrm{RBF}}^{(ji)}$ and hidden features $\vh_i$ and $\vh_j$. The updated messages $\vm'_{ji}$ of all neighbor nodes are then utilized to update hidden feature $\vh_i$.
A faster version of DimeNet is proposed later, dubbed DimeNet++~\citep{gasteiger2020fast,zhu2023fastdimenet++}.

\textbf{GemNet}~\citep{klicpera2021gemnet}. To achieve universal expressivity, GemNet further takes dihedral angles into account, formulating two-hop directional message passing based on quadruplets
of nodes. Basically, it replaces the message embeddings from \cref{eq:dimenet} in DimeNet~\citep{Klicpera2020Directional} with the following form:
\begin{align}
\label{eq:gemnet}
    \vm'_{ji} &=\sigma_{\mathrm{msg}}\left(\vm_{ji}, \textstyle\sum_{\substack{k\in\gN_i\backslash\{j\}\\l\in\gN_k\backslash\{i,j\}}}\vm_{jikl}\right),  \\
    \vm_{jikl}&=\sigma_{\mathrm{int}}\left(\vm_{lk},\ve_{\mathrm{RBF}}^{(lk)},\ve_{\mathrm{CBF}}^{(ikl)},\ve_{\mathrm{SBF}}^{(jikl)}\right),
\end{align}
where, $\ve_{\mathrm{RBF}}^{(lk)}$ and $\ve_{\mathrm{CBF}}^{(ikl)}$ are defined as above; $\ve_{\mathrm{SBF}}^{(jikl)}$\footnote{Here SBF is short for Spherical Bessel Function.} are calculated by, the spherical Bessel function of relative distance $d_{ji}$,
and spherical harmonics of angle $\alpha_{ji,ik}$ and dihedral angle $\alpha_{ji, kl}$. The input of \cref{eq:gemnet} additionally integrates hidden features $\vh_i$ and $\vh_j$ for more expressivity in its original formulation.  
Note that GemNet can be modified to enable equivariant output by multiplying the output with the associated direction, which belongs to scalarization based equivariant GNNs introduced in the next subsection. 

\textbf{LieConv~\citep{finzi2020generalizing}.}
 LieConv is formulated as follows.
\begin{align}
\label{eq:lie-m}
    \vm_{ij} &= \sigma\left(\log(u_i^{-1}u_j)\right)\vh_j, \\
\label{eq:lie-h}
    \vh_i' &= \textstyle\frac{1}{|\gN(i)|+1}\left(\vh_i + \sum_{j\in\gN(i)} \vm_{ij}\right),  
\end{align}
where $u_i\in G$ is a lift of $\vec{\vx}_i$, the logarithm $\log$ maps each group member onto the Lie Algebra $\mathfrak{g}$ that is a vector space, and $\sigma$ is a parametric MLP. Besides, \cref{eq:lie-h} conducts normalization by the division of the number of all nodes, \emph{i.e.} $|\gN(i)|+1$.
It is clear that LieConv only specifies the update of node features $\vh_i$ while keeping the geometric vectors $\vec{\vx}_i$ unchanged. That means LieConv is invariant.

In addition to the above models, SphereNet~\citep{liu2022spherical} is another prevailing invariant GNN. Similar to GemNet, SphereNet also exploits relative distances, angles, and torsion angles for geometric modeling, which is able to distinguish almost all 3D graph structures. Moreover, its proposed spherical message passing (SMP) enables both fast and accurate 3D molecular learning on large-scale molecules. ComENet~\citep{wang2022comenet} is another type of invariant model which incorporates 3D information completely and efficiently.
It ensures global completeness of model only with message passing in $1$-hop neighborhood to avoid time-consuming calculations like torsion in SphereNet or dihedral angles in GemNet. $k$-DisGNN~\citep{li2024distance} relies solely on invariant relative distance information, yet adopts high-order message-passing frameworks from traditional graph learning (\emph{e.g.}, $k$-WL or $k$-FWL), achieving completeness for $k = 2$. GeoNGNN~\citep{li2024completeness}, the geometric extension of the simplest subgraph GNN (NGNN~\citep{zhang2021nested}), effectively utilizes local subgraph information and also attains completeness with only distance features.
There are also some other studies~\citep{qin2022fast,zhu2018quaternion,zhang2020quaternion,yue2024a} exploiting the quaternion algebra to represent the 3D rotation group, which mathematically ensures SO(3) invariance during the inference. Specifically, QMP~\citep{yue2024a} constructs quaternion message-passing module to distinguish the molecular conformations caused by bond torsions.

\subsection{Equivariant Graph Neural Networks}
\label{sec:Equivariant Graph Neural Networks}

In contrast to invariant GNNs that only conduct the update of invariant features, equivariant GNNs simultaneously update both invariant features and equivariant features, given that many practical tasks (such as molecular dynamics simulation) requires equivariant output. More importantly, as proved in~\citep{joshi2023expressive}, equivariant GNNs are strictly more expressive than invariant GNNs particularly for sparse geometric graphs. 

In form, equivariant GNNs design the function over geometric graphs as $(\mH',\vec{\mX}') = \phi (\vec{\gG})$ satisfying:
\begin{align}
    \phi(g\cdot\vec{\gG}) = g\cdot\phi(\vec{\gG}), \forall g\in\mathrm{E}(3).
\end{align}
Specifically, through the lens of message-passing in \cref{eq:message,eq:update}, the geometric message is derived as
\begin{align}
\label{eq:geom_message}
    \vm_{ij}, \vec{\vm}_{ij}&=\phi_{\text{msg}}\left(\vh_i, \vh_j, \vec{\vx}_i,\vec{\vx}_j,\ve_{ij} \right).
\end{align}
In subsequent, the computed geometric messages $\vec{\vm}_{ij}$ are aggregated within the neighborhood $\gN_i$ specified by the connectivity or adjacency matrix of the graph, and updated by taking the input features into account. This update process is formally summarized as
\begin{align}
    \label{eq:geom:update}
\vh'_i,\vec{\vx}'_i&=\phi_{\text{upd}}\left(\vh_i, \{(\vm_{ij},\vec{\vm}_{ij})\}_{j\in\gN_i} \right).
\end{align}
The functions $\phi_{\text{msg}}$ and $\phi_{\text{upd}}$ should ensure that all invariant/equivariant output to be invariant/equivariant with respect to any $\mathrm{E}(3)$ transformation of the input. 

There are different ways to realize the specific form of $\phi_{\text{msg}}$ and $\phi_{\text{upd}}$. Below, we categorize current famous equivariant GNNs into two classes: scalarization-based models and high-degree steerable models.

\subsubsection{Scalarization-Based Models}
\label{sec:scalarization}

This line of works first translates 3D coordinates into invariant scalars, which is similar to the design of invariant GNNs, but it refines beyond invariant GNNs by further recovering the direction of the processed scalars for the update of equivariant features. 

\textbf{EGNN}~\citep{satorras2021en}. EGNN is one of the most famous scalarization based models, and it can be considered as an equivariant enhancement of two prior works, SchNet~\citep{schutt2018schnet} and Radial Field~\citep{kohler2019equivariant}.  For its message function $\phi_{\text{msg}}(\cdot)$, it first applies the relative distance for the update of invariant message, which is then multiplied back with the relative coordinate to derive directional message. The form of $\phi_{\text{msg}}(\cdot)$ is as follows:
\begin{align}
\label{eq:egnn_scalar_message}
    \vm_{ij} &= \sigma_1\left(\vh_i,\vh_j, \|\vec{\vx}_i-\vec{\vx}_j  \|^2, \ve_{ij}\right),\\
    \label{eq:egnn_vector_message}
    \vec{\vm}_{ij} &= (\vec{\vx}_i-\vec{\vx}_j)\sigma_2\left(\vm_{ij}\right),
\end{align}
while the update function $\phi_{\text{upd}}(\cdot)$ takes the following form,
\begin{align}
\label{eq:egnn_scalar_update}
    \vh'_i &= \sigma_3\left(\vh_i,  \textstyle\sum_{j\in\gN_i}\vm_{ij} \right),\\
    \label{eq:egnn_vector_update}
    \vec{\vx}'_i &= \vec{\vx}_i + \gamma\textstyle\sum_{j\in\gN_i}\vec{\vm}_{ij},
\end{align}
where $\sigma_1,\sigma_2,\sigma_3$ are all instantiated as Multi-Layer Perceptrons (MLPs), $\gamma$ is a predefined constant. 

\textbf{GMN}~\citep{huang2022equivariant}. In practice, each node is usually associated with multiple geometric features besides 3D position, such as velocity and force. Therefore, GMN proposes a multi-channel version of EGNN by defining a multi-channel vector $\vec{\mV}_i\in\R^{3\times C}$ for node $i$, where different channel (column) indicates different kind of geometric vector. In the message computation, the multi-channel vectors interact through inner product and are properly normalized for more training stability just before they are fed into the MLP, \emph{i.e.},
\begin{align}
\label{eq:gmn_scalar_message}
    \vm_{ij}&=\sigma_1\left(\vh_i, \vh_j, \frac{\vec{\mV}_{ij}^\top\vec{\mV}_{ij}}{\|\vec{\mV}_{ij}^\top\vec{\mV}_{ij} \|_F}, \ve_{ij} \right),\\
    \label{eq:gmn_vector_message}
\vec{\mM}_{ij}&=\vec{\mV}_{ij}\sigma_2\left(\vm_{ij}\right),
\end{align}
where $\vec{\mV}_{ij}$ is a translation-invariant directional matrix related to $\vec{\mV}_i$ and $\vec{\mV}_j$; for instance, if we have $\vec{\mV}_i=[\vec{\vx}_i,\dot{\vec{\vx}}_i]$ where $\dot{\vec{\vx}}_i\in\R^3$ defines the velocity, then we can either choose the direct subtraction $\vec{\mV}_{ij}=\vec{\mV}_i-\vec{\mV}_j$, or the concatenate form $\vec{\mV}_{ij}=[\vec{\mV}_i, \vec{\mV}_j]$ where the first channel of $\vec{\mV}_i$ and $\vec{\mV}_j$ is made translation invariant by subtracting the mean coordinate~\citep{han2022equivariant}. The update process is analogous to \cref{eq:egnn_scalar_update,eq:egnn_vector_update}, but extended to the multi-channel fashion as well.

\textbf{PaiNN}~\citep{schutt2021equivariant}. By initializing the multi-channel equivariant features to be zeros, namely, letting $\vec{\mV}_i=\vec{\mathbf{0}}\in\R^{3\times C}$, PaiNN iteratively updates $\vec{\mV}_i$ as well as invariant feature $\vh_i$ via the fixed relative position of the input coordinates $\vec{\vx}_{ij}=\vec{\vx}_i-\vec{\vx}_j$ in each layer, with the help of residual connection and gated non-linearity. We rewrite and somehow generalize the original form proposed by~\citep{schutt2021equivariant} using our consistent denotations. The messages are given by:
\begin{align}
 \vm_{ij} &= \sigma_1\left(\vh_j, \|\vec{\vx}_{ij}\|^2, \ve_{ij}\right),\\
    \vec{\mM}_{ij} &= \vec{\mV}_j\sigma_2\left(\vm_{ij}\right) + \vec{\vx}_{ij}\sigma_3\left(\vm_{ij}\right), 
\end{align}
and the update functions are calculated as:
\begin{align}
    \vm_i &= \vh_i + \textstyle\sum_{j\in\gN(i)}\vm_{ij}, \\
    \vec{\mM}_i &= \vec{\mV}_i + \textstyle\sum_{j\in\gN(i)}\vec{\mM}_{ij},\\
    \label{eq:PaiNN-h-update}
    \vh'_i &= \vm_i + \sigma_4\left(\vm_i,\|\vec{\mM}_i\| \right),\\
    \label{eq:PaiNN-X-update}
    \vec{\mV}'_i &= \vec{\mM}_i + \vec{\mM}_i\sigma_5\left(\vm_i,\|\vec{\mM}_i\| \right),
\end{align}
where, the functions $\sigma_1$-$\sigma_5$ are non-linear invariant scaling functions. In \cref{eq:PaiNN-h-update,eq:PaiNN-X-update}, ``$\|\cdot\|$'' outputs a multi-channel scalar each channel of which computes the vector norm of each channel of the input matrix.

\textbf{Local Frames~\citep{du2022se,kofinas2021rototranslated,kofinas2023latent}.} These methods construct local \emph{frames} (\emph{i.e.}, reference frames) $\vec\mF\in\sR^{3\times 3}$ that are equivariant to rotations and can be utilized to project the geometric information into invariant representations. In particular, LoCS~\citep{kofinas2021rototranslated} and Aether~\citep{kofinas2021rototranslated} leverage the angular position  $\vec\vw_i\in\sR^3$ of each node $i$ to construct node-wise local frames $\vec\mF_i=\mR(\vec\vw_i)$ where $\mR(\vec\vw_i)\in\sR^{3\times 3}$ is the corresponding rotation matrix of the angular position $\vec\vw_i$. ClofNet~\citep{du2022se} instead builds up edge-wise local frames $\vec\mF_{ij}=[\vec\va_{ij},\vec\vb_{ij},\vec\vc_{ij}]$, with
\begin{equation}
    \begin{aligned}
        \vec\va_{ij}=\frac{\vec\vx_{ic}-\vec\vx_{jc}}{\|\vec\vx_{ic}-\vec\vx_{jc}\|},\vec\vb_{ij}=\frac{\vec\vx_{ic}\times\vec\vx_{jc}}{\|\vec\vx_{ic}\times\vec\vx_{jc}\|}, \\
        \vec\vc_{ij}=\vec\va_{ij}\times\vec\vb_{ij}.
    \end{aligned}
\end{equation}
Here $\vec\vx_{ic}=\vec\vx_i-\vec\vx_c$ is translation-invariant by subtracting the center of mass $\vec\vx_c=\frac{1}{N}\sum_{i=1}^N\vec\vx_i$ so that the frame $\vec\mF_{ij}$ is also translation-invariant.

With local frames, the invariant message $\vm_{ij}$ is generated as
\begin{align}
    \vm_{ij}&=\sigma_1\left(\vh_i,\vh_j, \vec\mV_{ij}^\top\vec\mF_{ij}\right),
\end{align}
where $\vec\mV_{ij}$ is the translation-invariant geometric information between node $i$ and $j$, similar to the considerations in GMN (\cref{eq:gmn_vector_message}). ClofNet additionally considers to project the invariant message into an equivariant counterpart:
\begin{align}
    \vec{\mM}_{ij}=\vec{\mF}_{ij} \sigma_2\left(\vm_{ij} \right).
\end{align}

There are other works that exploit the scalarization technique to permit equivariance.  GVP-GNN~\citep{jing2021learning} first performs channel-wise linear projection of the input vector to align the channel dimension, and then computes the normalization of the projected vector as the scalar that is multiplied with the vector as the output vector. During this process, GVP-GNN does not pass the information from the input scalars, which is different from EGNN where the input scalars also influence the update of the vector.  EGHN~\citep{han2022equivariant}, built upon GMN, leverages a hierarchical encoder-decoder mechanism to represent the multi-body interaction with specially-designed equivariant pooling and unpooling modules. FastEGNN~\citep{zhang2024improving} addresses large-scale geometric graph scenarios by employing a small ordered set of virtual nodes, which minimizes the number of required edges and enhances computational efficiency. In LEFTNet~\citep{du2024new}, a local hierarchy of 3D isomorphism is proposed to evaluate the expressive power of equivariant GNNs and investigate the process of representing global geometric information from local patches. This work leads to two crucial modules for designing expressive and efficient geometric GNNs: local substructure encoding and frame transition encoding. SaVeNet~\citep{aykent2023savenet} enhances the numerical stability of the model by introducing gradually decaying directional noise during the training phase. ViSNet~\citep{wang2024enhancing} employs vector-scalar interactive message passing to implicitly extract various geometric features. QuinNet~\citep{wang2023efficiently} integrates many-body interactions, extending this modeling to include interactions of up to five bodies. Furthermore, HEGNN~\citep{cen2024high} leverages the inner product of high-degree steerable features to enhance scalar messaging, thereby achieving a balance between efficiency and effectiveness. Additionally, as scalars can be combined with various other invariant information, ETNN~\citep{battiloro2025etnn} further amplifies the expressiveness of the model by introducing deep topological learning constructs. EquiLLM~\citep{li2025large} enhances the representation of invariant scalars through knowledge injection from large language models, and can be flexibly generalized to various geometry tasks.

For all above methods, the scalarization process is implemented via the inner-product operator. In contrast to this, Frame-Averaging~\citep{puny2021frame} proposes to ensure equivariance via this averaging process: $\frac{1}{|G|}\sum_{g\in G}g\cdot\sigma(g^{-1}\cdot\vec{\vx})$, where $\sigma$ is an arbitrary MLP and the term $g^{-1}\cdot\vec{\vx}$ makes the input invariant. To deal with the case when the cardinality of $G$ is large, ~\cite{puny2021frame} instead conduct an average over a carefully selected subset that is obtained by the so-called frame function. The idea of Frame-Averaging is latter exploited in the field of material design~\citep{duval2023faenet}.

\subsubsection{High-Degree Steerable Models}
\label{sec:high-order}

For the aforementioned scalarization-based models, the node variables to be updated include invariant scalars $\vh_i$ and equivariant vectors $\vec{\vx}_i$ (or $\vec{\mV}_i$ for the multi-channel case), and the 3D rotation representation throughout the network is the rotation matrix $\mR_g$. It will be observed that scalars and vectors are respectively type-$0$ and type-$1$ steerable features, and the rotation matrix is the $1$-th degree matrix of a more general rotation representation. We will show that it is possible to derive high-degree representations of steerable features beyond scalars and vectors in equivariant GNNs. 

Prior to the introduction of high-degree models, we first introduce the concepts: \textbf{1.}  Wigner-D matrices~\citep{gilmore2008lie} to convert 3D rotations to group representations of different degree; \textbf{2.} spherical harmonics~\citep{muller2006spherical} to convert 3D vectors to steerable features of different type;  \textbf{3.} Clebsch-Gordan (CG) tensor product~\citep{griffiths2018introduction} to perform equivariant mapping between steerable features.

\textbf{Wigner-D matrices}.
In the general high-degree case, a widely studied genre of the representation for the rotation group SO($3$) is the irreducible representation~\citep{gilmore2008lie}:
\begin{align}
\label{eq:WD}
    \rho(g)\coloneqq \mD^{(l)}(g)\in\R^{(2l+1)\times(2l+1)}, g\in\text{SO}(3),
\end{align}
where $\mD^{(l)}(g)$ is the $l$-th degree Wigner-D matrix\footnote{Wigner-D matrices lie in the complex space, but they can be transformed to the real space under appropriate bases.}, and $l\in\sN=\{0,1,2,\dots\}$. In particular, $\mD^{(0)}(g)=1$ reduces to trivial representation and $\mD^{(1)}(g)=\mR_g$ takes the form of the rotation matrix.  The steerability of a type-$l$ feature $\vec{\vx}^{(l)}\in\R^{2l+1}$ is defined as $\mD^{(l)}(g)\vec{\vx}^{(l)}$, which naturally unifies the aforementioned invariant features and equivariant features by restricting $l=0$ and $l=1$, separately. Provided that there could be steerable features of multiple types and multiple channels, we provide a general form of steerable features:
\begin{align}
    \vec{\sV}^{(\sL)}\coloneqq \{\vec{\mV}^{(l)}\in\R^{(2l+1)\times C_l}\}_{l\in\sL},
\end{align}
where $\sL$ is the set consisting of all possible types and $C_l$ is the number of channels for type $l$. Since we are addressing geometric graphs in this paper, we will specify the steerable features of node $i$ as $\vec{\sV}_i^{(\sL)}$ and its type-$l$ component as $\vec{\mV}^{(l)}_i$.

\textbf{Spherical harmonics}.
We have defined how to steer type-$l$ features via Wigner-D matrices, but we do not know yet how to obtain type-$l$ features given 3D coordinates. Spherical harmonics are such tools to serve this purpose. Spherical harmonics are a set of Fourier basis on the unit sphere $S^2$. They map 3D vectors on the unit sphere $S^2$ into $(2l+1)$-dimensional vector space\footnote{Similar to Wigner-D matrices, the output of spherical harmonics are complex but can be transformed into real space under certain bases.}. That is,
\begin{align}
    Y^{(l)}(\vec{\vx}):S^2\mapsto\sR^{2l+1},
\end{align}
where $\vec\vx$ is a unit vector on the sphere, and the elements in $Y^{(l)}$ are usually used together and denoted as $[Y^{(l)}_{-l},Y^{(l)}_{-l+1},\cdots, Y^{(l)}_{l-1}, Y^{(l)}_{l}]$ where different element is called different order. It can also be generalized to take arbitrary 3D vector as input by properly normalizing the vector as $\frac{\vec\vx}{\|\vec\vx\|}$ prior to feeding into the spherical harmonics. This offers a unified view of transition to vector spaces of arbitrary type, where scalars correspond to $Y^{(0)}(\vec\vx)=1$ when $l=0$, and vectors correspond to $Y^{(1)}(\vec\vx)=\vec\vx\in\sR^3$ when $l=1$.
More importantly, spherical harmonics are equivariant in terms of Wigner-D matrices:
\begin{align}
    \label{eq:equ of sh}
    Y^{(l)}(\mR_g\vec{\vx})=\mD^{(l)}(g)Y^{(l)}(\vec{\vx}), g\in\text{SO}(3).
\end{align}
where $\mR_g\in\sR^{3\times3}$ is the rotation matrix and $D(g)\in\sR^{(2l+1)\times (2l+1)}$ refers to the $l$-degree Wigner-D matrix. To create multi-type multi-channel steerable features, we apply $Y^{(l)}$ over multiple copies for each type in $\sL$, yielding $\sY^{(\sL)}$.

\textbf{Clebsch-Gordan (CG) tensor product}.
Although spherical harmonics offer a way to design equivariant mapping from 3D coordinates (type-1 features) to type-$l$ features, they are unable to depict the interactions between steerable features of arbitrary types, which, however, is central to the design of equivariant functions when their input contains steerable features of various types. Fortunately, CG tensor product provides a tractable solution to this issue~\citep{griffiths2018introduction}. It derive $\vec{\mV}^{(l)}\in\R^{(2l+1)\times C}$ from two multi-channel steerable features $\vec{\mV}^{(l_1)}\in\R^{(2l_1+1)\times C_1}, \vec{\mV}^{(l_2)}\in\R^{(2l_2+1)\times C_2}$ by:
\begin{align}
\label{eq:cg_weight}
    \vec{\mV}^{(l)} = \left[\vec{\mV}^{(l_1)} \otimes_{\text{cg}}^{\mW} \vec{\mV}^{(l_2)}\right]^{(l)},
\end{align}
which can be expanded in details by:
\begin{align}
\label{eq:cg-w}
    v^{(l)}_{m,c} = \sum_{\substack{c_1=1\\c_2=1}}^{C_1,C_2}w_{c_1 c_2 c} \sum_{\substack{m_1=-l_1\\m_2=-l_2}}^{l_1,l_2}  Q^{(l,m)}_{(l_1, m_1)(l_2, m_2)} v_{m_1, c_1}^{(l_1)} v_{m_2, c_2}^{(l_2)},
\end{align}
where $v^{(l)}_{m,c}$ indicates the $m$-th order and $c$-th channel of $\vec{\mV}^{(l)}$; $Q^{(l,m)}_{(l_1, m_1)(l_2, m_2)}$ are the Clebsch-Gordan (CG) coefficients~\citep{griffiths2018introduction} and are zeros unless $|l_1-l_2|\leq l\leq l_1+l_2$; $w_{c_1c_2c}$ is the learnable parameter in the parameter matrix $\mW\in\R^{C_1\times C_2\times C}$, and when $\mW$ are all ones, \cref{eq:cg-w} reduces to the traditional non-parametric CG tensor product.

One promising property of CG tensor product is that it is $\mathrm{SO}(3)$-equivariant regarding Wigner-D matrices, implying that $\forall g\in\mathrm{SO}(3)$, 
\begin{equation}
\scriptstyle\mD^{(l)}(g)\vec{\mV}^{(l)}=\left[\left(\mD^{(l_1)}(g)\vec{\mV}^{(l_1)}\right) \otimes_{\text{cg}}^{\mW} \left(\mD^{(l_2)}(g)\vec{\mV}^{(l_2)}\right)\right]^{(l)}.
\end{equation}

For simplicity, the steerable variables in \cref{eq:cg_weight} are all of a single type. It is tractable to generalize \cref{eq:cg_weight} to the multi-type case by employing it over each combination of input-output type, and assigning different learnable parameters accordingly, which leads to a general form as follows:
\begin{align}
\label{eq:CG-set}
 \vec{\sV}^{(\sL)} = \vec{\sV}^{(\sL_1)} \otimes_{\text{cg}}^{\sW} \vec{\sV}^{(\sL_2)}.   
\end{align}

With the above building blocks, we below introduce several prevailing high-degree steerable models where the updated steerable variables for each node are $\vec{\sV}_i^{(\sL)}$.

\textbf{TFN~\citep{thomas2018tensor}.} With our formulation for the high-degree steerable operations, Tensor Field Network (TFN) computes the following equivariant point convolution:
\begin{align}
    \label{eq:tfn_message}
    \vec{\sM}^{(\sL)}_{ij} = \sY^{(\sL)}\left(\frac{\vec{\vx}_{ij}}{\|\vec{\vx}_{ij}\|}\right)\otimes_{\text{cg}}^\sW \vec\sV_j^{(\sL)},
\end{align}
where $\vec{\vx}_{ij}=\vec\vx_i-\vec\vx_j$ is the radial vector, and the element in $\sW$ is generated by a radial MLP $f(\|\vec{\vx}_{ij}\|)$ upon the distance $\|\vec{\vx}_{ij}\|$. Here $\vec{\vx}_i$ are fixed as the initial coordinates of the input data. The update of each node is implemented as a series of operations including aggregation:
\begin{align}
    \label{eq:tfn_aggregation}
    \vec{\sU}^{(\sL)}_i =\vec{\sV}^{(\sL)}_i + \textstyle\sum_{j\in\gN(i)}\vec{\sM}^{(\sL)}_{ij},
\end{align}
self-interaction:
\begin{align}
    \label{eq:tfn_self_interaction}
    \vec{\sV}^{(\sL)}_i = \{\vec\mU^{(l)}\mW^{(l)} \}_{l\in\sL},
\end{align}
where $\mW^{(l)}\in\sR^{c_l\times c_l}$ is the learnable channel-mixing matrix for each type $l$,
and node-wise non-linearity:
\begin{align}
    \label{eq:tfn_nonlinear}
    \vec{\sV}'^{(\sL)}=\left\{\vec\mV^{(l)}\sigma\left(\|\vec\mV^{(l)} \|_2 + \vb^{(l)} \right)\right\}_{l\in\sL},
\end{align}
where $\sigma\left(\cdot\right)$ is an activation function, ``$\|\cdot \|_2$'' is the $L_2$ vector norm over the order dimension (with size $(2l+1)$) of $\vec\mV^{(l)}$, and $\vb^{(l)}\in\sR^{c_l}$ is the bias for type $l$.

\textbf{SEGNN~\citep{brandstetter2022geometric}. } SEGNN enhances TFN from  equivariant point convolution to general equivariant message passing. Firstly, SEGNN involves high-degree geometric features from both node $i$ and $j$ in message computation by deriving $\vec{\sV}^{(\sL)}_{ij}=\vec{\sV}_i^{(\sL)}\oplus \vec{\sV}_j^{(\sL)}\oplus\{\|\vec{\vx}_{ij}\|^2\}$, where, again,   $\vec{\vx}_{ij}=\vec\vx_i-\vec\vx_j$ is the radial vector, and ``$\oplus$'' denotes concatenation along the channel dimension for the steerable features with the same type $l\in\sL$. For example, 
\begin{align}
    \vec{\sV}_1^{(\sL)}\oplus\vec{\sV}_2^{(\sL)}\coloneqq \{\vec{\mV}_1^{(l)}\| \vec{\mV}_2^{(l)} \}_{l\in\sL}.
\end{align}
Here ``$\|$'' stands for concatenation along the channel. Subsequently, the high-degree linear message passing specified in \cref{eq:tfn_message} is extended to a non-linear fashion via gated non-linearities~\citep{weiler20183d}:
\begin{align}
\label{eq:segnn_message1}
        \vec{\sV}^{(\sL)}_{ij}, \vg_{ij} &= \sY^{(\sL)}\left(\frac{\vec{\vx}_{ij}}{\|\vec{\vx}_{ij}\|}\right)\otimes_{\text{cg}}^{\sW_1} \vec\sV_{ij}^{(\sL)},\\
        \label{eq:segnn_message2}
        \vec{\sM}^{(\sL)}_{ij} &= \text{Gate}\left(\vec{\sV}^{(\sL)}_{ij},\text{Swish}(\vg_{ij}) \right),
\end{align}
where $\text{Gate}\left(\cdot\right)$ is the gated non-linearity introduced in~\citep{weiler20183d}, $\text{Swish}\left(\cdot\right)$ is the Swish activation~\citep{ramachandran2017searching}, and $\vg_{ij}$ is a scalar read out from the CG tensor product that will further be leveraged to control the scale in the non-linearity of \cref{eq:segnn_message2}. Notably, the CG product and non-linearity in \cref{eq:segnn_message1,eq:segnn_message2} are performed twice in the implementation of ~\citep{brandstetter2022geometric}. Analogous to the design of multi-layer perceptrons (MLPs), they are dubbed the steerable MLP.

The update function also employs the proposed steerable MLP. In detail,
\begin{align}
    \label{eq:segnn_update1}
        \vec{\sV}^{(\sL)}_{i}, \vg_{i} =&\left(\sum_{j\in\gN_i} \sY^{(\sL)}\left(\frac{\vec{\vx}_{ij}}{\|\vec{\vx}_{ij}\|}\right)\right) \notag\\
        &\otimes_{\text{cg}}^{\sW_2} \left(\vec{\sV}^{(\sL)}_{i} + \sum_{j\in\gN_i}\vec\sM_{ij}^{(\sL)}\right),\\
        \label{eq:segnn_update2}
        \vec{\sV}'^{(\sL)}_{i} = &\vec{\sV}^{(\sL)}_{i} + \text{Gate}\left(\vec{\sV}^{(\sL)}_{i},\text{Swish}(\vg_{i}) \right).
\end{align}

Besides those have been introduced above, there are still many methods to build equivariant models with high-degree steerable features. 
Cormorant~\citep{anderson2019cormorant} utilizes channel-wise CG product (a reduced and more efficient form of \cref{eq:cg_weight} that acts on each input channel independently) and channel concatenations to formulate one-body and two-body interactions among the input graph systems. 
NequIP~\citep{batzner20223} improves the convolutional layer in TFN~\citep{thomas2018tensor} by further introducing the radial Bessel functions and a polynomial envelope function used in DimeNet~\citep{Klicpera2020Directional} to get a better embedding of interaction distance, thereby improving the performance of the model. SCN~\citep{zitnick2022spherical} regards each node embedding as a set of spherical functions (\emph{i.e.}, the spherical harmonics), then conducts message passing by rotating the embeddings based on the 3D edge orientation, and finally updates the node embeddings via discrete spherical Fourier Transform. Its following work, eSCN~\citep{passaro2023reducing} proposes to reduce the computation complexity of the equivariant convolution on $\mathrm{SO}(3)$ with a mathematically equivalent one on $\mathrm{SO}(2)$. To enable higher body interaction beyond the two-body modeling in most previous papers, MACE~\citep{batatia2022mace} and Allegro~\citep{musaelian2023learning}, propose a simplified algorithm to construct the tensor product item, motivated by a new technology in physics called Atomic Cluster Expansion (ACE)~\citep{drautz2019atomic,dusson2022atomic,bochkarev2022efficient}.

An illustrative comparison of invariant GNNs, scalarization-based equivariant GNNs, and high-degree steerable equivariant GNNs is summarized in \cref{tab:models}.

\begin{table*}[t]
\caption{Illustrations of representative models for invariant GNNs, scalarization-based GNNs and high-degree steerable GNNs. Notably, these three types of models are able to process geometric features of different degrees.}
\label{tab:models}
\resizebox{\linewidth}{!}{
\begin{tabular}{
    cccc
}
\toprule
    & \makecell{Invariant GNNs \\ 
    (\emph{e.g.} SchNet~\citep{schutt2018schnet}) \\[0.8em] 
    \includegraphics[height = 5em]{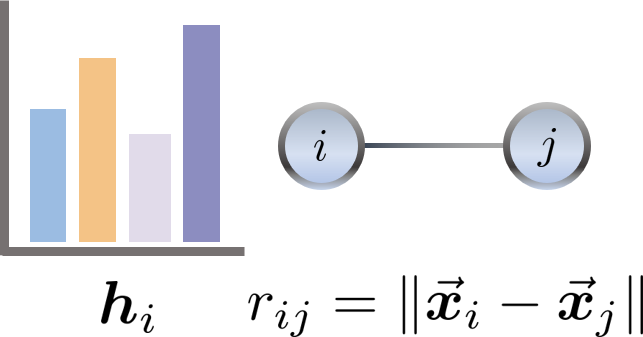}} 
    & \makecell{Scalarization-Based Models\\ 
    (\emph{e.g.} EGNN~\citep{satorras2021en}) \\[0.8em]  
    \includegraphics[height = 5em]{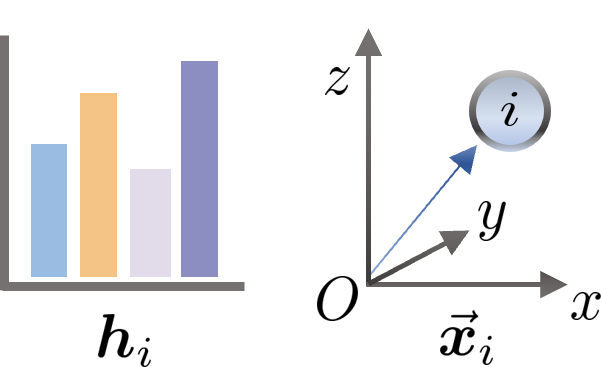}} 
    & \makecell{High-Degree Steerable Models \\ 
    (\emph{e.g.} TFN~\citep{thomas2018tensor}) \\[0.8em]  
    \includegraphics[height = 5em]{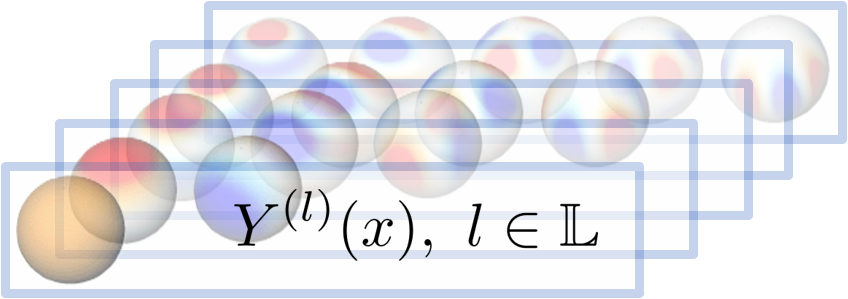}}\\
\midrule
\makecell{Message Computation \\ \includegraphics[height = 4em]{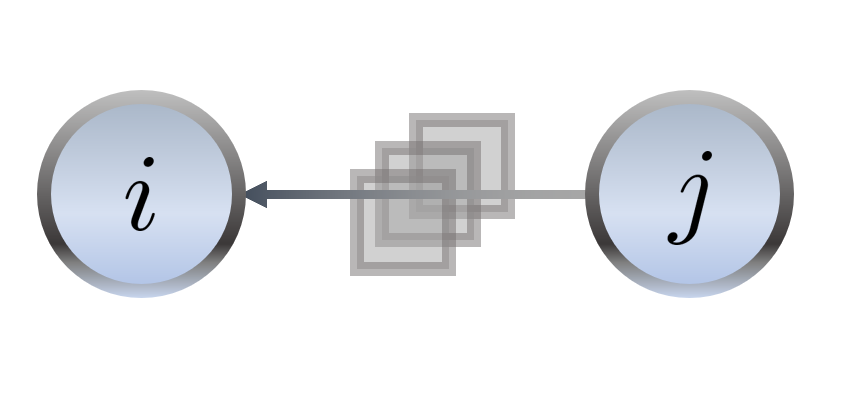}}
    & \large{$\vm_{ij} =\sigma_{2}(r_{ij})\sigma_1(\vh_j)$}
    & \large{$\begin{aligned}
        \vm_{ij} &= \sigma_1\left(\vh_i,\vh_j, \|\vec{\vx}_i-\vec{\vx}_j  \|^2, \ve_{ij}\right)\\
        \vec{\vm}_{ij} &= (\vec{\vx}_i-\vec{\vx}_j)\sigma_2\left(\vm_{ij}\right)
    \end{aligned}$}
    & \large{$\vec{\sM}^{(\sL)}_{ij} = \sY^{(\sL)}\left(\frac{\vec{\vx}_{ij}}{\|\vec{\vx}_{ij}\|}\right)\otimes_{\text{cg}}^\sW \vec\sV_j^{(\sL)}$}\\
\midrule
\makecell{Feature Update \\ \includegraphics[height = 4em]{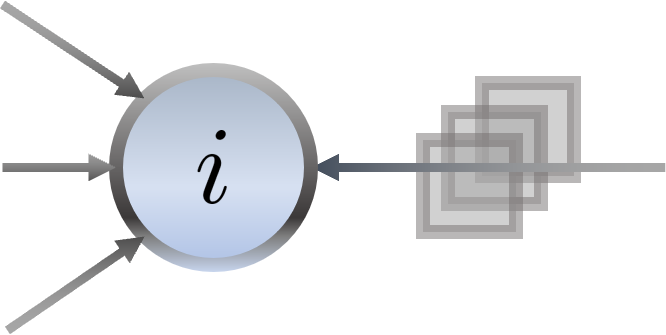}}
    & \large{$\vh'_i = \sigma_3\left(\vh_i,  \sum\nolimits_{j\in\gN_i}\vm_{ij} \right)$}
    & \large{$\begin{aligned}
        \vh'_i &= \sigma_3\left(\vh_i,  \sum\nolimits_{j\in\gN_i}\vm_{ij} \right)\\
        \vec{\vx}'_i &= \vec{\vx}_i + \gamma\sum\nolimits_{j\in\gN_i}\vec{\vm}_{ij}
    \end{aligned}$}
    & \large{$\vec{\sV}'^{(\sL)}_i =\vec{\sV}^{(\sL)}_i + \sigma\left(\vec{\sV}^{(\sL)}_i, \sum\nolimits_{j\in\gN(i)}\vec{\sM}^{(\sL)}_{ij}\right)$}\\
\bottomrule
\end{tabular}
}
\end{table*}

\subsection{Geometric Graph Transformers}
\label{sec:transformers}

Inspired by the significant success of Transformers~\citep{vaswani2017attention,yuan2025survey} in many areas, such as natural language processing and computer vision, there have been efforts to apply these self-attention-based architectures to data structure like graphs or even geometric graphs in the scope of this survey. Summarized in \cref{fig:tree_of_models}, these methods stem from different types of geometric representations, including invariant representation, scalarization-based equivariant representation, and high-degree steerable representation, which have been elaborated in~\cref{sec:geomgnn}. Below we discuss these Transformers in detail.

\textbf{Graphormer~\citep{ying2021transformers,shi2022benchmarking}.}  Graphormer has been firstly proposed as a powerful Transformer architecture operating on graphs, equipped with centrality encoding, spatial encoding, and edge encoding~\citep{ying2021transformers}. With its success on challenging 2D graph datasets, \emph{e.g.}, the OGB-LSC Challenge~\citep{hu2021ogb}, it has been subsequently extended to work on geometric graphs with special designs in computing the encodings. To be specific, the spatial encoding, which aims to measure the spatial relation between node $i$ and $j$ in $\vec{\gG}$, is chosen to be the Euclidean distance $\|\vec{\vx}_i-\vec{\vx}_j\|_2$ transformed by Gaussian basis functions~\citep{shuaibi2021rotation}. The centrality encoding is derived as a summation of the spatial encodings over the connected edges for each node. The encodings are then utilized in computing the self-attention, and layer normalization is also adopted for the intermediate features. Notably, all representations are E($3$)-invariant under the construction of Graphormer. In order to make it suitable for E($3$)-equivariant prediction tasks,~\citep{ying2021transformers} proposes to use a projection head as the final block, which aggregates the edge vectors, scaled by their corresponding attention weights to obtain a node-wise vector as output:
\begin{align}
    \vec{\vf}_i = \sum_{j\neq i}a_{ij}(\vec{\vx}_i-\vec{\vx}_j),
\end{align}
where $a_{ij}$ is the $\mathrm{E}(3)$-invariant attention weight between node $i$ and $j$.

\textbf{TorchMD-Net~\citep{tholke2022equivariant}.} TorchMD-Net is an equivariant Transformer that tackles general multi-channel geometric vectors in a scalarization-based manner, akin to PaiNN~\citep{schutt2021equivariant}. Yet, in the process of attention computation, only invariant representations $\vh_i$ and distances $\|\vec{\vx}_{ij}\|$ are involved. Specifically, the distance is firstly embedded by two MLPs $\sigma_{d_K}$ and $\sigma_{d_V}$ for the key and value, respectively:
\begin{align}
    \vd^K_{ij}=\sigma_{d_K}(\ve_{\text{RBF}}^{(ij)}),\quad \vd^V_{ij}=\sigma_{d_V}(\ve_{\text{RBF}}^{(ij)}),
\end{align}
where $\ve_{\text{RBF}}^{(ij)}$ is the radial basis function representation of distance $\|\vec{\vx}_{ij}\|$, similar to \cref{eq:dimenet}. The query, key, and value are given by linear transformations of the input scalar features:
\begin{align}
\vq_i=\vh_i\mW_Q, \vk_{ij}=\vh_i\mW_K\odot\vd_{ij}^K,  \vv_{ij}=\vh_j\mW_V\odot\vd^V_{ij},
\end{align}
where ``$\odot$'' is the element-wise product. Instead of traditionally adopted Softmax operator~\citep{vaswani2017attention}, TorchMD-Net simplifies to SiLU non-linearity:
\begin{align}
a_{ij}=\textstyle\sum_{C_h}\text{SiLU}\left(\vq_i\odot\vk_{ij}\right)\cdot\text{Cutoff}\left(\|\vec{\vx}_{ij}\|\right),
\end{align}
with $\text{Cutoff}\left(\cdot\right)$ being a cosine cutoff on the distance and the summation being over the channels of these invariant features. Finally, the output of the attention is yielded as
\begin{align}
\vh'_i=\left(\textstyle\sum_{j\in\gN_i}a_{ij}\vv_{ij}\right)\mW_O,
\end{align}
with $\mW_O$ being a linear transformation for the output.

\textbf{$\mathrm{SE}(3)$-Transformer~\citep{fuchs2020se}.} Different from Graphormer and TorchMD-Net that limit the representation to scalars and vectors with degree $l\in\{0, 1\}$, $\mathrm{SE}(3)$-Transformer employs attention mechanism on general steerable features with high degree. Following our notations introduced in~\cref{sec:high-order}, we describe the attention computation as follows.

The point-wise query $\vec{\sQ}_i^{(\sL)}$ and pairwise key $\vec{\sK}_{ij}^{(\sL)}$ and value $\vec{\sV}_{ij}^{(\sL)}$ are derived as:
\begin{equation}
\begin{aligned}
    \vec{\sQ}_i^{(\sL)} &= \mathbf{1}\otimes_{\text{cg}}^{\sW_Q}\vec{\sV}_i^{(\sL)},\\
    \vec{\sK}_{ij}^{(\sL)} &= \sY^{(\sL)}\left(\frac{\vec{\vx}_{ij}}{\|\vec\vx_{ij}\|}\right)\otimes_{\text{cg}}^{\sW_K}\vec{\sV}_j^{(\sL)},\\
    \vec{\sV}_{ij}^{(\sL)} &= \sY^{(\sL)}\left(\frac{\vec{\vx}_{ij}}{\|\vec\vx_{ij}\|}\right)\otimes_{\text{cg}}^{\sW_V}\vec{\sV}_j^{(\sL)}.
\end{aligned}
\end{equation}
The attention coefficient $a_{ij}$ is computed as a Softmax aggregation over the neighbors with message being the inner products of the queries and keys, ensuring rotation invariance:
\begin{align}
\alpha_{ij}=\frac{\exp\left(\vec{\sQ}_i^{(\sL)}\cdot \vec{\sK}_{ij}^{(\sL)}\right)}{\sum_{k\in\gN_i}\exp\left(\vec{\sQ}_i^{(\sL)}\cdot\vec{\sK}_{ik}^{(\sL)}\right)}.
\end{align}
The attention is then utilized to aggregate the values and update the node feature:
\begin{align}
\vec{\sV}'^{(\sL)}_i&=\mathbf{1}\otimes_{\text{cg}}^{\sW_1}\vec{\sV}_i^{(\sL)}+\textstyle\sum_{j\in\gN_i}\alpha_{ij}\vec{\sV}_{ij}^{(\sL)}.
\end{align}
With the invariant attention, the updated feature is easily guaranteed to satisfy $\mathrm{SE}(3)$-equivariance.

Besides, LieTransformer~\citep{hutchinson2021lietransformer} extends the idea of LieConv~\citep{finzi2020generalizing} by building attentions on top of lifting and sampling on Lie groups. GVP-Transformer introduced in~\citep{hsu2022learning} leverages GVP-GNN~\citep{jing2021learning} as the structural encoder and applies a generic Transformer over the extracted representation, exhibiting strong performance in learning inverse folding of proteins. Equiformer~\citep{liao2023equiformer} proposes to replace dot product attention in Transformers by MLP attention and non-linear message passing, building upon the space of high-degree steerable tensors. EquiformerV2~\citep{liao2023equiformerv2} further incorporates eSCN~\citep{passaro2023reducing} in the architecture for efficient modeling and introduces more technical enhancements like specially designed attention re-normalization and layer normalization for better empirical performance. Geoformer~\citep{wang2024geometric} develops an invariant module called Interatomic Positional Encoding (IPE) based on the invariant basis from ACE, in order to enhance the expressiveness of many-body contributions in the attention blocks. Recently, SO3KRATES~\citep{frank2024euclidean} proposed a technique aimed at leveraging the advantages of high-degree representations while simplifying the complexity inherent in tensor products. This approach focuses on the design of a model that utilizes only the paths that yield scalars in tensor products. Later, GotenNet~\citep{aykent2025rethinking} broadened the scope of the inner product form, creating a multi-channeled version and referring to models that employ this methodology as \emph{spherical-scalarization models}. GotenNet integrated the inner product with the original attention mechanism, resulting in an efficient equivariant transformer architecture.

As previous transformers typically focus on a specific domain, either proteins or small molecules. EPT~\citep{jiao2024equivariant} proposes a novel pretraining framework designed to harmonize the geometric learning of small molecules and proteins. It unifies the geometric modeling of multi-domain molecules via block-enhanced representation upon an PaiNN-based transformer framework.

\subsection{Theoretical Analysis on Expressivity}

In machine learning, an important criterion for measuring the expressiveness of a network is whether it has \emph{universal approximation property}. In the task of learning on geometric graphs, this is whether any function of geometric graphs can be approximated by geometric GNNs with arbitrary accuracy.

An initial attempt to explore this problem is conducted by~\citep{dym2020universality}, which proves the universality of the high-degree steerable model, \emph{i.e.},  TFN~\citep{thomas2018tensor}, over point clouds (namely fully-connected geometric graphs) by showing that TFN can fit any equivariant polynomials. GemNet~\citep{klicpera2021gemnet} further demonstrates that the universality holds with just spherical representations other than the full SO(3) representations that are required in the proof  of~\citep{dym2020universality}. 
Later, the GWL framework~\citep{joshi2023expressive} defines a geometric version of the Weisfeiler-Lehman (WL) test~\citep{weisfeiler1968reduction} to study the expressive power of geometric GNNs operating on sparse graphs from the perspective of discriminating geometric graphs, and discuses the difference of the expressivity between various invariant and equivariant GNNs, both theoretically and experimentally. One crucial conclusion drawn by the GWL paper is that GWL is strictly more powerful than invariant GWL, showing the advantage of equivariant GNNs against invariant GNNs. For fully-connected geometric graphs, invariant GWL has the same expressive power as GWL. More recently, HEGNN~\citep{cen2024high} has provided both theoretical and experimental insights into the necessity of employing high-degree steerable features on symmetric graphs. Specifically, under the strict equivariance constraint, the degradation of representations of certain degrees on symmetry graphs cannot be avoided unless it is circumvented by relaxing some conditions (\emph{e.g.} probabilistic symmetry breaking in SymPE~\citep{lawrence2025improving}). Furthermore, HEGNN establishes a connection between high-degree steerable features and Legendre polynomials, indicating that inner- product of sufficiently high-degree representations can recover all angular information present in geometric graphs.

There are other works that only investigate the universality of the message computation function~\citep{villar2021scalars,huang2022equivariant}. They explore the expressivity of the scalarization-based models (\emph{e.g.} EGNN), and \cite{villar2021scalars} confirms that the scalarization-based methods can  universally approximate any invariant/equivaraint functions of vectors. Besides, SGNN~\citep{han2022learning} generalizes from equivariance to subequivariance that depicts the case when part of the symmetry is broken by external force field, \emph{e.g.} gravity, and finally design an universal form of subequivariant functions. 

\section{Applications}
\label{sec:app}

In this section, we systematically review the applications related to geometric graph learning. We classify existing methods according to the system types they work on, which leads to the categorization of tasks on particle, (small) molecule, protein, molecule + molecule (Mol + Mol), molecule + protein (Mol + Protein), protein + protein, and other domains, as summarized in \cref{tab:allmodels}. We also provide a summary of all related datasets of single- and multiple-instance tasks in \cref{tab:dataset_single} and  \cref{tab:dataset_multi}, respectively.   \textbf{It is worth mentioning that our discussion primarily focuses on the methods utilizing geometric GNNs, although other methods, such as sequence-based approaches, may be applicable in certain applications}.
\newcommand{\rgr}{Prediction}
\newcommand{\cls}{Prediction}
\newcommand{\gen}{Generative}
\newcommand{\ngen}{Non-Generative}
\newcommand{\mix}{Mixed}

\begin{table*}[htbp]
\centering
\caption{The summary of various geometric GNNs for different tasks. The generative tasks indicates the ones addressable by generative models, otherwise referred to as the non-generative tasks. The ones can be solved with either generative or non-generative models are dubbed as the mixed tasks.}
\vspace{2pt}
\label{tab:allmodels}
\resizebox{0.95\linewidth}{!}{
\renewcommand{\arraystretch}{0.90}
\begin{tabular}
{m{0.15\linewidth}<{\raggedright}@{\hspace{0.5em}}m{0.28\linewidth}<{\raggedright}@{\hspace{0.5em}}m{0.15\linewidth}<{\raggedright}@{\hspace{0em}}m{0.78\linewidth}<{\raggedright}}

\toprule
\multicolumn{1}{c}{{\textbf{Data Type}}} & {\textbf{Task Name}} & {\textbf{Task Type}} & \hspace{1em}{\textbf{Methods}} \\
\midrule
\addlinespace[-0.1pt]
\rowcolor{gray!20}\multicolumn{4}{c}{Physics} \\
\addlinespace[-0.1pt]\midrule

\multirow{5}[2]{*}{Particle}
    & $N$-Body Simulation & \ngen & NRI~\citep{kipf2018neural}, IN~\citep{battaglia2016interaction}, E-NFs~\citep{satorras2021en}, EGNN~\citep{satorras2021en}, SEGNNs~\citep{brandstetter2022geometric}, GMN~\citep{huang2022equivariant}, EGHN~\citep{han2022equivariant}, HOGN~\citep{sanchez2019hamiltonian}, NCGNN~\citep{guo2023newton}, FastEGNN~\citep{zhang2024improving}, HEGNN~\citep{cen2024high}\\ 
    \cmidrule{2-4}
    & Scene Simulation & \ngen & SGNN~\citep{han2022learning}, GNS~\citep{sanchez2020learning}, 
    GNS*~\citep{allen2023graph}, 
    C-GNS~\citep{rubanova2022constraint},  HGNS\citep{wu2022learning}, DPI-Net~\citep{li2018learning}, HRN~\citep{mrowca2018flexible}, FIGNet~\citep{allen2023learning}, EGHN~\citep{han2022equivariant}, LoCS~\citep{kofinas2021rototranslated}, EqMotion~\citep{xu2023eqmotion}, ESTAG~\citep{wu2023equivariant}, SEGNO~\citep{liu2023physicsinspired}, FastEGNN~\citep{zhang2024improving}, HEGNN~\citep{cen2024high}, EquiLLM~\citep{li2025large}\\
\midrule
\addlinespace[-0.01pt]
\rowcolor{gray!20}\multicolumn{4}{c}{Biochemistry} \\
\addlinespace[-0.1pt]\midrule
\multirow{14}[7]{*}{Small Molecule}
    & Molecular Property Prediction & \ngen & Cormorant~\citep{anderson2019cormorant}, TFN~\citep{thomas2018tensor}, SE(3)-Transformer~\citep{fuchs2020se}, NequIP~\citep{batzner20223}, SEGNNs~\citep{brandstetter2022geometric}, LieConv~\citep{finzi2020generalizing}, Lietransformer~\citep{hutchinson2021lietransformer}, SchNet~\citep{schutt2018schnet}, DimeNet~\citep{Klicpera2020Directional}, GemNet~\citep{klicpera2021gemnet}, PaiNN~\citep{schutt2021equivariant}, TorchMD-Net~\citep{tholke2022equivariant}, Equiformer~\citep{liao2023equiformer}, SphereNet~\citep{coors2018spherenet}, EGNN~\citep{satorras2021en}, Graphormer~\citep{ying2021transformers,shi2022benchmarking}, SCN~\citep{zitnick2022spherical}, eSCN~\citep{passaro2023reducing}, GNN-LF~\citep{wangxiyuan2022graph}, LEFTNet~\citep{du2024new}, 
    SaVeNet~\citep{aykent2023savenet}, ViSNet~\citep{wang2024enhancing}, QuinNet~\citep{wang2023efficiently}, SO3KRATES~\citep{frank2024euclidean} , Gaunt~\citep{luo2024enabling}, 
    GotenNet~\citep{aykent2025rethinking}\\
    \cmidrule{2-4} 
    & Molecular Dynamics & \mix & E-CNF~\citep{kohler2020equivariant}, EGNN~\citep{satorras2021en}, NequIP~\citep{batzner20223}, GMN~\citep{huang2022equivariant}, EGHN~\citep{han2022equivariant}, NCGNN~\citep{guo2023newton}, ESTAG~\citep{wu2023equivariant}, EGNO~\citep{xu2024equivariant}, SEGNO~\citep{liu2023physicsinspired}, ITO~\citep{schreiner2023implicit}, E-ACF~\citep{midgley2024se}, GeoTDM~\citep{han2024geometric}, HEGNN~\citep{cen2024high}, StABlE~\citep{raja2024stability}, \cite{amin2025towards}\\
    \cmidrule{2-4}
    & Molecular Generation & \gen &GeoDiff~\citep{xu2022geodiff}, GeoLDM~\citep{xu2023geometric}, ConfVAE~\citep{xu2021an}, ConfGF~\citep{shi2021learning}, G-SchNet~\citep{gebauer2019symmetry}, cG-SchNet~\citep{gebauer2022inverse}, MDM~\citep{huang2023mdm}, MolDiff~\citep{peng2023moldiff}, 
    DGSM~\citep{luo2021predicting}, E-NFs~\citep{satorras2021enf}, EDM~\citep{hoogeboom2022equivariant}, GeoMol~\citep{ganea2021geomol}, Torsional~Diffusion~\citep{jing2022torsional}, MPerformer~\citep{wang2023mperformer},  EEGSDE~\citep{bao2023equivariant}, DMCG~\citep{zhu2022direct}, HierDiff~\citep{qiang2023coarse}, EquiFM~\citep{song2024equivariant}, CoarsenConf~\citep{reidenbach2024coarsenconf}, GeoBFN~\citep{song2024unified}, MolCRAFT~\citep{qu2024molcraft} \\ 
    \cmidrule{2-4}
    & Pretraining & \mix & 3D-EMGP~\citep{jiao2022energy}, GeoSSL-DDM~\citep{liu2023molecular}, GraphMVP~\citep{liu2022pretraining}, GNS-TAT~\citep{zaidi2023pretraining}, MGMAE~\citep{feng2022mgmae}, 
    3D-Infomax~\citep{stark20223d}, Uni-Mol~\citep{zhou2023unimol}, Transformer-M\citep{luo2023one}, MoleculeSDE~\citep{pmlr-v202-liu23h}, SliDe~\citep{ni2023sliced}, Frad~\citep{feng2023fractional}, DenoiseVAE~\citep{liu2025denoisevae}, MolSpectra~\citep{wang2025molspectra} \\
\midrule
\multirow{13}[6]{*}{Protein} 
    & Protein Property Prediction & \ngen & LM-GVP~\citep{wang2022lm}, DeepFRI~\citep{gligorijevic2021structure}, GearNet~\citep{zhang2022protein}, 3DCNN~\citep{torng20173d}, TM-align~\citep{zhang2005tm}, GVP-GNN~\citep{jing2021learning}, PAUL~\citep{eismann2020hierarchical}, EDN~\citep{eismann2023protein}, EnQA~\citep{chen20233d}, ScanNet~\citep{tubiana2022scannet}, EquiPocket~\citep{zhang2023equipocket}, PocketMiner~\citep{meller2023predicting} \\
    \cmidrule{2-4}
    & Protein Inverse Folding & \gen & GVP-GNN~\citep{jing2021learning}, \cite{ingraham2019generative}, ESM-IF1~\citep{hsu2022learning}, GCA~\citep{tan2022generative}, ProteinMPNN~\citep{dauparas2022robust}, PiFold~\citep{gao2022pifold}, LM-Design~\citep{pmlr-v202-zheng23a}, KW-Design~\citep{gao2023kw} \\ \cmidrule{2-4}
    & Protein Folding &\gen & AlphaFold~\citep{senior2020improved}, AlphaFold2~\citep{AlphaFold2021}, RosettaFold~\citep{minkyung2021rosettafold}, RosettaFold2~\citep{baek2023efficient}, 
    RFAA~\citep{krishna2023generalized}, EigenFold~\citep{jing2023eigenfold}, RFdiffusion~\citep{watson2023novo}, Chroma~\citep{ingraham2023illuminating}, ESMFold~\citep{lin2023evolutionary}, HelixFold-Single~\citep{fang2023method} \\
    \cmidrule{2-4}
    & Protein Co-Design & \gen & Chroma~\citep{ingraham2023illuminating}, RFdiffusion~\citep{watson2023novo}, PROTSEED~\citep{shi2023protein}, ReQFlow~\citep{yue2025reqflow} \\ \cmidrule{2-4}
    & Pretraining & \mix & ProtTrans~\citep{elnaggar2021prottrans}, 
    xTrimoPGLM~\citep{chen2024xtrimopglm}, 
    ProtGPT2~\citep{ferruz2022protgpt2}, 
    HJRSS~\citep{mansoor2021toward}, 
    GearNet~\citep{zhang2022protein}, 
    ProFSA~\citep{gao2023self},
    PromptProtein~\citep{wang2023multi},  DrugCLIP~\citep{gao2023drugclip}, ESM-1b~\citep{rives2019biological}, ESM2~\citep{lin2023evolutionary}, \cite{guo2022self}, PAAG~\citep{yuan2024annotation}\\
\midrule
\multirow{2}[1]{*}{Mol + Mol}
    & Linker Design & \gen & DiffLinker~\citep{igashov2024equivariant}, DeLinker~\citep{imrie2020deep}, 3DLinker~\citep{huang20223dlinker} \\
    & Chemical Reaction & \gen & OA-ReactDiff~\citep{duan2023accurate}, TSNet~\citep{jackson2021tsnet} \\ \midrule
\multirow{6}[4]{*}{Mol + Protein}
    & Ligand Binding Affinity Prediction & \ngen & TargetDiff~\citep{guan2023d}, MaSIF~\citep{gainza2020deciphering}, GET~\citep{kong2023generalist}, ProtNet~\citep{wang2022learning}, HGIN~\citep{zhao2023geometric}, BindNet~\citep{feng2023protein}, BADGER~\citep{jian2024general}, DeepTernary~\citep{xue2025se} \\
    \cmidrule{2-4}
    & Protein-Ligand Docking & \mix & EquiBind~\citep{stark2022equibind}, DiffDock~\citep{corso2023diffdock}, TankBind~\citep{lu2022tankbind}, DESERT~\citep{long2022zero}, FABind~\citep{pei2023fabind}, Re-Dock~\citep{huang2024re} \\
    \cmidrule{2-4}
    & Pocket-Based Mol Sampling & \mix & Pocket2Mol~\citep{peng2022pocket}, TargetDiff~\citep{guan2023d}, DiffBP~\citep{lin2022diffbp}, SBDD~\citep{luo2021a}, GraphBP~\citep{liu2022generating}, FLAG~\citep{zhang2023molecule}, DESERT~\citep{long2022zero}, D3FG~\citep{lin2024functional}, MolCRAFT~\citep{qu2024molcraft}, MolJO~\citep{qiu2024structure}, DiffBP~\citep{lin2022diffbp}, VoxBind~\citep{pinheiro2024structure}\\
\midrule
\multirow{8}[5]{*}{Protein + Protein}
    & Protein Interface Prediction & \ngen & DeepInteract~\citep{morehead2022geometric}, dMaSIF~\citep{sverrisson2021fast}, SASNet~\citep{townshend2019end} \\ \cmidrule{2-4}
    & Binding Affinity Prediction & \ngen & mmCSM-PPI~\citep{rodrigues2021mmcsm}, GeoPPI~\citep{liu2021deep}, GET~\citep{kong2023generalist} \\ \cmidrule{2-4}
    & Protein-Protein Docking & \mix & EquiDock~\citep{ganea2022independent}, HMR~\citep{wang2023learning}, HSRN~\citep{jin2022antibody}, DiffDock-PP~\citep{ketata2023diffdock}, SyNDock~\citep{ji2023syndock}, 
    AlphaFold-Multimer~\citep{evans2021protein}, dMaSIF~\citep{sverrisson2022physics},  ElliDock~\citep{yu2024rigid}, EBMDock~\citep{wu2024neural} \\
    \cmidrule{2-4}
    & Antibody Design & \mix & DiffAb~\citep{luo2022antigenspecific}, MEAN~\citep{kong2023conditional}, dyMEAN~\citep{kong2023end}, RefineGNN~\citep{jin2022iterative}, PROTSEED~\citep{shi2023protein}, 
    AbBERT\citep{gao2022incorporating}, ADesigner~\citep{tan2023cross}, AbODE\citep{verma2023abode}, AbDiffuser~\citep{martinkus2023abdiffuser}, tFold\citep{wu2024fast}, GeoAB~\citep{lin2024geoab}, RAAD~\citep{wu2024relation}, EquiLLM~\cite{li2025large} \\
    \cmidrule{2-4}
    & Peptide Design & \mix & HelixGAN~\citep{xie2023helixgan}, RFDiffusion~\citep{watson2023novo}, PepGLAD~\citep{kong2024full}, PPFlow~\citep{lin2024ppflow} \\ \midrule
\addlinespace[-0.01pt]
\rowcolor{gray!20}\multicolumn{4}{c}{Other Domains} \\
\addlinespace[-0.1pt]\midrule
\multirow{6}[4]{*}{Others}
    & Crystal Property Prediction &\ngen & CGCNN~\citep{xie2018crystal} , MEGNet~\citep{chen2019graph}, ALIGNN~\citep{choudhary2021atomistic}, ECN~\citep{kaba2022equivariant}, Matformer~\citep{yan2022periodic}, Crystal Twins~\citep{magar2022crystal} , MMPT~\citep{yu2023crystal}, CrysDiff~\citep{song2024diffusion} \\ \cmidrule{2-4}
    
    & Crystal Generation & \gen & CDVAE~\citep{xie2021crystal}, SyMat~\citep{luo2023towards}, DiffCSP~\citep{jiao2023crystal},  
    DiffCSP++~\citep{jiao2024space},
    MatterGen~\citep{zeni2023mattergen}, PXRDGen~\citep{li2024powder}, EquiCSP~\citep{lin2024equivariant}, FlowMM~\citep{millerflowmm}, CrysBFN~\citep{wu2025a} \\ \cmidrule{2-4}

    & RNA Structure Ranking & \ngen & ARES~\citep{raphael2021geometric}, PaxNet~\citep{zhang2022physics},
    EquiRNA~\citep{li2025sizegeneralizable} \\

\bottomrule
\end{tabular}
}
\end{table*}

\begin{table*}[htbp]
\centering
\caption{The summary of typical datasets and benchmarks for the single instance applications.}
\resizebox{0.9\textwidth}{!}{
\renewcommand{\arraystretch}{0.73}
\begin{tabular}{@{\hspace{0.5em}}p{0.35\textwidth}@{\hspace{0.5em}}
|@{\hspace{0.5em}}p{0.22\textwidth}@{\hspace{0em}}
@{\hspace{0em}}p{0.28\textwidth}@{\hspace{0em}}
@{\hspace{0.5em}}p{0.28\textwidth}@{\hspace{0em}} }
\toprule
Dataset    & \# sample & \hspace{0.5em} Task       & Benchmark \\
\midrule
\addlinespace[-0.1pt]
\rowcolor{gray!20}\multicolumn{4}{c}{Particle} \\ 
\addlinespace[-0.1pt]\midrule
    $N$-Body~\citep{kipf2018neural}	&70K	&$N$-body Simulation	&NRI~\citep{kipf2018neural} \\ \midrule 
    3D $N$-Body~\citep{satorras2021en}	&7K	&$N$-body Simulation	&EGNN~\citep{satorras2021en} \\ \midrule
    Constrained $N$-Body~\citep{huang2022equivariant}	&5.5K	&$N$-body Simulation	&GMN~\citep{huang2022equivariant} \\ \midrule
    Hierarchical $N$-Body~\citep{satorras2021en}	&9K	&$N$-body Simulation	&EGHN~\citep{satorras2021en} \\ \midrule
    Water3D~\citep{sanchez2020learning}	&0.8K	&Scene Simulation	&GNS~\citep{sanchez2020learning} \\ \midrule
    Kubric MOVi-A~\citep{greff2022kubric}	&0.02K	&Scene Simulation	&GNS*~\citep{allen2023graph} \\ \midrule
    Physion~\citep{bear2021physion}	&16K &Scene Simulation	&SGNN~\citep{han2022learning} \\ \midrule
    MIT Pushing~\citep{yu2016more}	&6K	&Scene Simulation	&FIGNet~\citep{allen2023learning} \\ \midrule
    FluidFall~\citep{li2018learning}	&3K	&Scene Simulation	&DPI-Net~\citep{li2018learning} \\ \midrule
    FluidShake~\citep{li2018learning}	&2K	&Scene Simulation	&DPI-Net~\citep{li2018learning} \\ \midrule
    BoxBath~\citep{li2018learning}	&3K	&Scene Simulation	&DPI-Net~\citep{li2018learning} \\ \midrule
    RiceGrip~\citep{li2018learning}	&5K	&Scene Simulation	&DPI-Net~\citep{li2018learning} \\ \midrule
\addlinespace[-0.1pt]
\rowcolor{gray!20}\multicolumn{4}{c}{Small Molecule} \\
\addlinespace[-0.1pt]\midrule
    \multirow{4}[2]{*}{QM9~\citep{ramakrishnan2014quantum}} &\multirow{4}[2]{*}{134K}  &Molecule Property Prediction &ATOM3D~\citep{townshend2021atomd}  \\ \cmidrule{3-4} & &Molecule Generation &GEOM-QM9~\citep{xu2021learning} \\ \cmidrule{3-4} & &Molecule Pretraining &3D-Infomax~\citep{stark20223d}   \\ \midrule
    \multirow{2}[1]{*}{MD17~\citep{chmiela2017machine}} &\multirow{2}[1]{*}{3.6M}  &Molecule Property Prediction &SchNet~\citep{schutt2018schnet} \\ \cmidrule{3-4} & &Molecule Dynamics &GMN~\citep{huang2022equivariant} \\ \midrule
    \multirow{2}[1]{*}{OCP~\citep{oc22_dataset}} &\multirow{2}[1]{*}{9.8M}  &Molecule Property Prediction &eSCN~\citep{passaro2023reducing} \\ \cmidrule{3-4} & &Molecular Dynamics &GemNet~\citep{klicpera2021gemnet} \\ \midrule
    Adk~\citep{seyler5108170molecular}	&4.1K	&Molecular Dynamics	&EGHN~\citep{satorras2021en}  \\ \midrule
    DW-4~\citep{kohler2020equivariant} &10K	&Molecular Dynamics	&EQ-Flow~\citep{kohler2020equivariant}  \\ \midrule
    LJ-13~\citep{kohler2020equivariant} &10K	&Molecular Dynamics	&EQ-Flow~\citep{kohler2020equivariant}  \\ \midrule
    Fast-folding proteins~\citep{lindorff2011fast} &5M	&Molecular Dynamics	&ITO~\citep{schreiner2023implicit}  \\ \midrule
    \multirow{4}[2]{*}{GEOM~\citep{axelrod2022geom}} &\multirow{4}[2]{*}{450K}  &Molecule Property Prediction &SchNet~\citep{schutt2018schnet} \\ \cmidrule{3-4} &  &Molecule Generation &GEOM-Drugs~\citep{xu2021learning} \\ \cmidrule{3-4} & &Molecule Pretraining &GMN~\citep{huang2022equivariant} \\ \midrule
    PCQM4Mv2~\citep{hu2021ogb} &3.3M &Molecule Pretraining & 3D PGT~\citep{wang2023automated} \\ \midrule
    QMugs~\citep{isert2022qmugs}	&665K &Molecule Pretraining  &3D-Infomax~\citep{stark20223d} \\ \midrule
    Uni-Mol~\citep{zhou2023unimol}	&209M &Molecule Pretraining &Uni-Mol~\citep{zhou2023unimol} \\ \midrule
\addlinespace[-0.1pt]
\rowcolor{gray!20}\multicolumn{4}{c}{Protein} \\ 
\addlinespace[-0.1pt]\midrule
    GENE Ontology~\citep{ashburner2000gene}  &33.5K   &Protein Property Prediction &GearNet~\citep{zhang2022protein} \\  \midrule
    ENZYME~\citep{bairoch2000enzyme} &18.5K  &Protein Property Prediction &GearNet~\citep{zhang2022protein} \\ \midrule
    \multirow{4}[2]{*}{CATH~\citep{orengo1997cath}} &\multirow{4}[2]{*}{189K}  &Protein Inverse Folding
    &GVP-GNN~\citep{wang2022lm}  \\ \cmidrule{3-4} & &Protein Pretraining &S2F~\citep{xue2022multimodal} \\ \cmidrule{3-4} & &Protein Co-Design &PROTSEED~\citep{shi2023protein} \\ \midrule
    \multirow{4}[2]{*}{SCOPe~\citep{chandonia2019scope}} &\multirow{4}[2]{*}{108K}  &Protein Inverse Folding
    &ProstT5~\citep{heinzinger2023prostt5} \\ \cmidrule{3-4} & &Protein Pretraining &ProSE~\citep{bepler2021learning} \\ \cmidrule{3-4}  & &Protein Property Prediction &TAPE~\citep{rao2019evaluating} \\ \midrule
    \multirow{4}[2]{*}{AlphaFoldDB~\citep{varadi2022alphafold}} &\multirow{4}[2]{*}{200M}  &Protein Folding
    &ESMFold~\citep{lin2023evolutionary}  \\ \cmidrule{3-4} & &Protein Inverse Folding &AlphaDesign~\citep{gao2022alphadesign} \\ \cmidrule{3-4} & &Protein Pretraining &GearNet~\citep{zhang2022protein} \\ \midrule
    \multirow{2}[1]{*}{UniProt~\citep{uniprot2023uniprot}}	&\multirow{2}[1]{*}{216M}	&Protein Pretraining	&Prottrans~\citep{elnaggar2021prottrans} \\ \cmidrule{3-4}
    & &Protein Property Prediction &DeepLoc~\citep{almagro2017deeploc} \\
    \midrule
    BFD~\citep{steinegger2018clustering}	&2100M	&Protein Pretraining	&Prottrans~\citep{elnaggar2021prottrans} \\ \midrule
    NetSurfP-2.0~\citep{klausen2019netsurfp}	&11.3K	&Protein Pretraining	&PEER~\citep{xu2022peer} \\ \midrule
    CASP~\citep{kryshtafovych2019critical}	&45.7K	&Protein Structure Ranking	&ATOM3D~\citep{townshend2021atomd} \\ \midrule
    \multirow{2}[1]{*}{PDB~\citep{berman2000protein}} &\multirow{2}[1]{*}{1.2M}  &Protein Residue Identity &ATOM3D~\citep{townshend2021atomd} \\ \cmidrule{3-4} & &Protein Folding &ESMFold~\citep{lin2023evolutionary} \\ 
\bottomrule
\end{tabular}%
}

\label{tab:dataset_single}%
\end{table*}%

\begin{table*}[htbp]
\centering
\caption{The summary of typical datasets and benchmarks for the multi-instance applications.}
\label{tab:dataset_multi}
\resizebox{0.95\textwidth}{!}{
\begin{tabular}{@{\hspace{0.5em}}p{0.32\textwidth}@{\hspace{0.5em}}
|@{\hspace{0.5em}}p{0.12\textwidth}@{\hspace{0em}}
@{\hspace{0em}}p{0.35\textwidth}@{\hspace{0em}}
@{\hspace{0.5em}}p{0.28\textwidth}@{\hspace{0em}} }
\toprule
Dataset    & \# sample & \hspace{0.5em} Task       & Benchmark \\
\midrule
\addlinespace[-0.1pt]
\rowcolor{gray!20}\multicolumn{4}{c}{Mol + Mol} \\ 
\addlinespace[-0.1pt]\midrule
    ZINC~\citep{sterling2015zinc} &727K	&Linker Design	&3DLinker~\citep{huang20223dlinker} \\ \midrule
    CASF~\citep{su2019comparative} &0.28K &Linker Design &DeLinker~\citep{imrie2020deep} \\ \midrule
    GEOM~\citep{axelrod2022geom} &450K	&Linker Design	&DiffLinker~\citep{igashov2024equivariant} \\ \midrule
    SN2-TS~\citep{jackson2021tsnet}	&0.11K &Chemical Reaction	&TSNet~\citep{jackson2021tsnet} \\ \midrule
    Transition1x~\citep{schreiner2022transition1x} &9.6M	&Chemical Reaction	&OA-ReactDiff~\citep{duan2023accurate} \\ \midrule
\addlinespace[-0.1pt]
\rowcolor{gray!20}\multicolumn{4}{c}{Mol + Protein} \\ 
\addlinespace[-0.1pt]\midrule
    \multirow{2}[1]{*}{CrossDocked 2020~\citep{francouer2020three}} &\multirow{2}[1]{*}{22.5M}  &Ligand Affinity &GNINA~\citep{francouer2020three} \\ \cmidrule{3-4} & &Pocket-Based Molecule Sampling &TargetDiff~\citep{guan2023d} \\ \midrule
    \multirow{2}[1]{*}{PDBBind~\citep{liu2017forging}} &\multirow{2}[1]{*}{23.5K}  &Ligand Affinity &ATOM3D~\citep{townshend2021atomd} \\ \cmidrule{3-4} & &Protein-Ligand Docking &EquiBind~\citep{stark2022equibind} \\ \midrule
\addlinespace[-0.1pt]
\rowcolor{gray!20}\multicolumn{4}{c}{Protein + Protein} \\ 
\addlinespace[-0.1pt]\midrule
    \multirow{2}[1]{*}{DIPS~\citep{townshend2019end}} &\multirow{2}[1]{*}{42.8K}  &Protein Interface Prediction &ATOM3D~\citep{townshend2021atomd} \\ \cmidrule{3-4} & &Protein-Protein Docking &EquiDock~\citep{ganea2022independent} \\ \midrule
    DIPS-plus~\citep{morehead2021dips} &42.1K &Protein Interface Prediction	&DeepInteract~\citep{morehead2022geometric} \\ \midrule
    Biogrid~\citep{stark2006biogrid} &1.7M	&Protein Interface Prediction &SYNTERACT~\citep{hallee2023protein} \\ \midrule
    \multirow{4}[2]{*}{DB5.5~\citep{vreven2015updates}} &\multirow{4}[2]{*}{0.23K}  &Protein Interface Prediction &ATOM3D-PIP~\citep{townshend2021atomd} \\ \cmidrule{3-4} & &Protein-Protein Docking &EquiDock~\citep{ganea2022independent} \\ \cmidrule{3-4} & &Binding Affinity Prediction &GET~\citep{kong2023generalist} \\ \midrule
    \multirow{1}[1]{*}{PDBBind~\citep{liu2017forging}}	&\multirow{1}[1]{*}{23.5K} &Binding Affinity Prediction &GeoPPI~\citep{liu2021deep} \\ \midrule
    SAbDab~\citep{dunbar2014sabdab}	&8.1K &Antibody Design	&RefineGNN~\citep{jin2022iterative} \\ \midrule
    RAbD~\citep{adolf2018rosettaantibodydesign}	&0.06K	    &Antibody Design &RefineGNN~\citep{jin2022iterative} \\ \midrule
    \multirow{2}[1]{*}{SKEMPI 2.0~\citep{jankauskaite2019skempi}}	&\multirow{2}[1]{*}{7.1K} &Antibody Design	&ATOM3D~\citep{townshend2021atomd} \\ \cmidrule{3-4}
    & &Binding Affinity Prediction & mmCSM-PPI~\citep{rodrigues2021mmcsm} \\ \midrule
    Cov-abdab~\citep{raybould2021cov}	&2.4K &Antibody Design	&RefineGNN~\citep{jin2022iterative} \\ \midrule
    PepBDB~\citep{wen2019pepbdb} &13K &Peptide Design &CAMP~\citep{lei2021deep} \\ \midrule
    LNR~\citep{tsaban2022harnessing} &0.09K &Peptide Design &PDAR~\citep{tsaban2022harnessing} \\ \midrule
    PepGLAD~\citep{kong2024full} & 6K &Peptide Design &PepGLAD~\citep{kong2024full} \\ \midrule
    PPFlow~\cite{lin2024ppflow} & 13K &Peptide Design & PPBench2024~\citep{lin2024ppflow} \\ \midrule
\addlinespace[-0.1pt]
\rowcolor{gray!20}\multicolumn{4}{c}{Others} \\ 
\addlinespace[-0.1pt]\midrule
    \multirow{2}[1]{*}{Materials Project~\citep{jain2013commentary}} &\multirow{2}[1]{*}{154K}  &Protein Crys. Property Prediction &CGCNN~\citep{xie2018crystal} \\ \cmidrule{3-4} & &Crystal Generation &CDVAE~\citep{xie2021crystal} \\ \midrule
    Perov-5~\citep{castelli2012new, castelli2012computational} &18.9K &Crys. Generation &CDVAE~\citep{xie2021crystal} \\ \midrule
    Carbon-24~\citep{carbon2020data} &10.1K &Crys. Generation &CDVAE~\citep{xie2021crystal} \\ \midrule
    ARVIS-DFT~\citep{choudhary2020joint}	&41K	&Crys. Property Prediction	&JARVIS-ML~\citep{choudhary2018machine} \\ \midrule
    FARFAR2-Puzzles~\citep{watkins2020farfar2} &18K  &RNA Struct. Ranking &ARES~\citep{townshend2021atomd} \\ \midrule
    rRNAsolo~\citep{li2025sizegeneralizable} &92K  &RNA Struct. Ranking &EquiRNA~\citep{li2025sizegeneralizable} \\ 
\addlinespace[-0.5pt]
\bottomrule
\end{tabular}
}
\end{table*}

\subsection{Tasks on Particles}
\label{sec:Particle}

The particle representation serves as an abstract and unified concept in the context of dynamic modeling in physics. Rigid bodies, elastic bodies and even fluid can be modeled as a set of particles~\citep{han2022learning}. 
Under such a particle-based modeling, a physical object of interest corresponds to a geometric graph $\vec{\gG}$ as specified in \cref{def:gg}, where different particles are modeled as different nodes, and physical interactions between particles such as attraction/repulsion force, collision, rolling, and sliding are denoted as edge connections.

\subsubsection{Physical Dynamics Simulation}
\label{sec:physical dynamics simulation}
Geometric GNNs have been widely applied  to characterize the process of general physical dynamics. One typical example is $N$-body simulation, which is originally proposed by~\citep{kipf2018neural} and targets at modeling the dynamics of a prototype system composed of $N$ interacting particles. While it is built under an ideal condition, an $N$-body system is capable of representing various physical phenomena across a spectrum encompassing quantum physics through to astronomy, by accommodating diverse interactions. Other examples include the simulation of physical scenes that involves more complex objects including fluids, rigid-bodies, deformable-bodies, and human motions.

\emph{Task definition:}
Given the initial state of the system represented by a geometric graph $\vec{\gG}^{(0)}$, the future states of all $N$ particles after a period of $k$ steps are predicted by a parametric function:
\begin{align}
    \vec{\mX}^{(t+k)} = \phi_{\theta}(\vec{\gG}^{(t)}).
    \label{equ:$N$-bodydef}
\end{align}
In contrast to the above single-state prediction setting, one may also conduct a ``roll-out''  simulation by recurrently taking the predicted output of current state as the input for the prediction of the next state. Furthermore, it can also be extended to the spatio-temporal setting by taking the historical geometric graphs within a window of size $w$ (namely $\vec{\gG}^{(t-w+1:t)}$) as input, rather than a single input frame (namely $\vec{\gG}^{(t)}$) in~\cref{equ:$N$-bodydef}.

\emph{Symmetry preserved:}
This is an E($3$)-equivariant task, as the transformation of the initial state results in the same transformation of the predicted state. It means $g\cdot\phi_\theta\left(\vec{\gG} \right)=\phi_\theta\left(g\cdot\vec{\gG} \right), \forall g\in\mathrm{E}(3)$.

\emph{Datasets:} 
The datasets used in current methods belong to the following classes:
\textbf{1) $N$-body dataset series.}
The original $N$-body dataset~\citep{kipf2018neural} presents an environment capable of simulating three types of system, including 1D phase-coupled oscillators, 2D springs and 2D charged balls. The authors in~\citep{fuchs2020se} further generalize $N$-body to encompass 3D cases. Recently, the work~\citep{huang2022equivariant} designs Constrained $N$-body by adding geometric constraints between particles, leading to a combination of diverse systems with isolated particles, sticks and hinges. Later, the systems derived by~\citep{han2022equivariant} further introduce the interactions between complex objects that are composed of multiple particles interconnected by rigid sticks.
\textbf{2) Scene simulation datasets.} The paper~\citep{li2018learning} proposes four simulation environments: FluidFall, FluidShake, BoxBath, and RiceGrip, where the former two focus on fluid modeling, the third one tests fluid-rigid interactions, and the final one involves modeling deformable objects with elastic/plastic properties. Similar to BoxBath, Water-3D created by~\citep{sanchez2020learning} randomly initializes the water states and constructs a high-resolution water scenario. Beyond the simulation of particle-level interaction in previous datasets, Kubric~\citep{greff2022kubric} and MIT Pushing~\citep{yu2016more} can be utilized to evaluate face interactions. Physion~\citep{bear2021physion} is a large-scale dataset that involves more realistic and diverse objects driven by more complex physical interactions, including gravity, friction, elasticity, and other factors.

\emph{Methods:}
Plenty of studies have been devoted to learning to simulate complex physical systems using GNNs, including Interaction Network~\citep{battaglia2016interaction}, NRI~\citep{kipf2018neural}, HRN~\citep{mrowca2018flexible}, DPI-Net~\citep{li2018learning}, HOGN~\citep{sanchez2019hamiltonian}, GNS~\citep{sanchez2020learning}, C-GNS~\citep{rubanova2022constraint}, HGNS~\citep{wu2022learning}, GNS*~\citep{allen2023graph}, and FIGNet~\citep{allen2023learning}. However, all these methods adopt typical GNNs that are unaware of full symmetry in 3D world, and only a subset of them considers translation-equivariance. Since the work of SE(3)-Transformer~\citep{fuchs2020se}, roto-translation equivariance is introduced upon the attention-based geometric GNNs to address the $N$-body problem. Later,  
EGNN~\citep{satorras2021en} proposes a more effective E($n$)-equivariant GNN by using the scalarization-based strategy as already detailed in \cref{sec:scalarization}. In contrast to EGNN, SEGNN~\citep{brandstetter2022geometric} proposes a general SE(3)-equivariant message passing by making use of high-order degree representations. Recently, GMN~\citep{huang2022equivariant} have developed multi-channel equivariant modeling specifically for constrained $N$-body systems consisting of sticks or hinges. 
Upon GMN, EGHN~\citep{han2022equivariant} designs equivariant pooling and equivariant unpooling to handle the complex system with a hierarchical structure.
In the meantime, SGNN~\citep{han2022learning} generalizes and relaxes the symmetry from equivariance to sub-equivariance, which plausibly grants it the capability to excel in scenarios influenced by other factors like gravity. 
As conventional approaches utilize a fixed velocity estimation throughout the time interval, NCGNN~\citep{guo2023newton} instead estimates velocities at multiple time points using Newton-Cotes numerical integration.
There are also other works that approach physical simulation based on the spatio-temporal setting. LoCS~\citep{kofinas2021rototranslated} utilizes GRU to record the memory of past frames and additionally incorporates rotation-invariance to improve
the model’s generalization ability; EqMotion~\citep{xu2023eqmotion} distills the history trajectories of each node into a multi-dimension vector and then designs an equivariant module and an interaction reasoning module to predict future frames; 
ESTAG~\citep{wu2023equivariant} employs equivariant discrete Fourier Transform along with the equivariant spatio-temporal attention mechanism to model the physical dynamics.
SEGNO~\citep{liu2024segno} incorporates the second-order graph neural ODE with equivariant property to reduce the roll-out error of long-term physical simulation.

\subsection{Tasks on Small Molecules}
\label{sec:Molecule}
By representing atom coordinates as node positions and bonds as edges, a molecule naturally becomes a geometric graph where $\vec{\mX} \in \sR^{N\times 3}$ represents the positions of $N$ atoms in the molecular, $\mH \in \sR^{N \times C_h}$ indicates the atom types or other properties of the atoms, and $\mA\in \{0, 1\}^{N\times N}$ represents the existence of bonds. Usually, the edge feature $e_{ij}\in\{0,1,2,3\}$ is defined by the bond type of the edge from node $i$ to $j$. In addition to chemical edges, the relative distance $d_{ij}$ between two atoms is also utilized for constructing k-NN spatial edges by selecting for each atom the $k$ nearest atoms as its neighbors, and the spatial edge feature is defined as  $\ve_{ij}=\sigma(d_{ij})$ where $\sigma$ is a non-linear function, such as RBF.

Prior to the use of geometric graph, a molecule could be typically represented by a 1D string (\emph{e.g.} SMILES\citep{downs1989review} and SMARTS\citep{lipinski2012experimental}) or a 2D topological graph, both of which lose sight of the geometric information of the molecule, resulting in defective performance for the tasks that involve crucial spatial interactions between atoms. Here, we only introduce the works that apply geometric graphs to represent molecules.  

\subsubsection{Molecular Property Prediction}

Molecular property prediction has been a fundamental task in computational biochemistry and machine learning. As pinpointed by MoleculeNet~\citep{C7SC02664A}, common properties can be subdivided into four categories: quantum mechanics, physical chemistry, biophysics and physiology. With the help of geometric GNNs, we are now able to additionally consider the molecular geometries which have been demonstrated to be crucial in determining the quantum chemistry properties of molecules. 

\emph{Task definition:} With the input molecule characterized as a geometric graph $\vec{\gG}$, the task is to learn a model $\phi_\theta$ to predict a scalar property $\vy$ and/or a vectorial property $\vec{\vy}$:
\begin{align}
    \vy, \vec{\vy} = \phi_\theta\left(\vec{\gG} \right).
\end{align}
While most works mainly focus on the single-task setting by predicting each individual type of property independently, it is also possible to leverage the multi-task setting by predicting multiple types of property simultaneously.

\emph{Symmetry preserved:} It is an SE($3$)-invariant task in terms of $\vy$ since it remains unaffected by any rotation or translation exerted on the molecule, \emph{i.e.}, $\phi_\theta\left(\vec{\gG} \right)=\phi_\theta\left(g\cdot\vec{\gG} \right), \forall g\in\mathrm{SE}(3)$. As for $\vec{\vy}$, we enforce SE($3$)-equivariance into the model: $g\cdot\phi_\theta\left(\vec{\gG} \right)=\phi_\theta\left(g\cdot\vec{\gG} \right), \forall g\in\mathrm{SE}(3)$.

\emph{Datasets:}
There are currently three popular data sources for the evaluation of this task, including QM9~\citep{ramakrishnan2014quantum}, MD17~\citep{chmiela2017machine} and Open Catalyst Project (OCP)~\citep{oc22_dataset}. The QM9 dataset contains 131K small organic molecules with up to nine heavy atoms from CONF, and each molecular is annotated with 13 property labels ranging from the highest occupied molecular orbital to the norm of the dipole moment. 
MD17 is a collection of molecular dynamic simulations for eight small organic molecules, whose goal is to predict both the energy and atomic forces of each molecule, given the atom coordinates in the non-equilibrium and slightly moving system. OCP consists of more than 100M atomic structures for catalysts to help address climate change, each composed of a molecule called adsorbate placed on a slab named catalyst. OCP provides two datasets OC20~\citep{ocp_dataset} and OC22~\citep{oc22_dataset} for benchmarking, and there are three kinds of tasks in OCP where Initial Structure to Relaxed Energy (IS2RE) taking an initial structure as input to predict the relaxed energy is a highly challenging task.

\emph{Methods:} 
Most of the methods introduced in \cref{sec:geomgnn} are evaluated on molecular property prediction tasks. Here, to avoid redundant introduction, we 
no longer describe each method in detail and only specify which of the three mentioned benchmarks they are evaluated on. Specifically, invariant GNNs (including
SchNet~\citep{schutt2018schnet}, DimeNet~\citep{Klicpera2020Directional}, SphereNet~\citep{coors2018spherenet}, and GemNet~\citep{klicpera2021gemnet}), equivariant GNNs (including Cormorant~\citep{anderson2019cormorant} and PaiNN~\citep{schutt2021equivariant}) and equivariant graph transformers (\emph{e.g.} TorchMD-Net~\citep{tholke2022equivariant} and Equiformer~\citep{liao2023equiformer}) employ both QM9 and MD17 for performance comparisons. Other methods like NequIP~\citep{batzner20223} are conducted on MD17, while EGNN~\citep{satorras2021en}, LieConv~\citep{finzi2020generalizing} and SE(3)-Transformer~\citep{fuchs2020se} are evaluated on QM9. SEGNN~\citep{brandstetter2022geometric}, Graphormer~\citep{ying2021transformers,shi2022benchmarking}, Equiformer~\citep{liao2023equiformer}, SCN~\citep{zitnick2022spherical}, and eSCN~\citep{passaro2023reducing} leverage more challenging benchmarks, namely, OC20 and even OC22 for performance assessment, revealing encouraging effectiveness of applying geometric GNNs to catalyst design.

\subsubsection{Molecular Dynamics Simulation}
Molecular Dynamics (MD) simulation aims to simulate the temporal evolution process of molecules driven by internal interactions between atoms within the same molecule, external interactions among different molecules, or environmental interactions from solvents and force fields.

\emph{Task definition:} Given an input molecular graph at time $t$, \emph{i.e.}, $\vec{\gG}^{(t)}$, this task simulates the dynamical evolution of the molecular over some time. In general, the future coordinates $\vec{\mX}^{(t+k)} (k>0)$ are estimated by
\begin{align}
    \vec{\mX}^{(t+k)} = \phi_{\theta}\left(\vec{\gG}^{(t)}\right).
\end{align}
Similar to general physical dynamics simulation in \cref{sec:physical dynamics simulation}, one may also conduct a roll-out prediction setting or the spatio-temporal input setting. Besides, in contrast to the direct trajectory prediction here, MD can be alternatively addressed with the methods designed for molecular property prediction as described in the last subsection. We can first predict the node-level force $\vec{\mF}\in\sR^{N\times 3}$ or the graph-level system energy $E\in\sR$ for the given state of the system $\vec{\gG}$, and then use these estimated quantities to update the molecular dynamics by solving the differential equations that describe molecular dynamics.

\emph{Symmetry preserved:} 
Clearly, the output coordinate matrix $\vec{\mX}^{(t+k)}$ is E($3$)-equivariant.

\emph{Datasets:}
MD17~\citep{chmiela2017machine}, AdK~\citep{seyler5108170molecular}, OCP~\citep{oc22_dataset}, DW-4~\citep{kohler2020equivariant}, fast-folding proteins~\citep{lindorff2011fast}, and LJ-13~\citep{kohler2020equivariant} are available datasets for MD simulation in the machine learning community.
MD-17~\citep{chmiela2017machine} which is usually used for molecular property prediction also contains the trajectories of eight molecules generated via DFT. The AdK equilibrium trajectory dataset simulated by CHARMM27 force field in the MDAnalysis software~\citep{gowers2016mdanalysis} involves the MD trajectory of apo adenylate kinase with explicit water and ions in NPT at 300 K and 1 bar, where the atom positions of the protein are saved every 240 $ps$ for a total of 1.004  $\mu s$. Besides the common relaxed energy prediction task, OCP releases a dataset split for MD, which computes short, high-temperature ab initio MD trajectories on a randomly sampled subset of the relaxed states. DW-4 is a relatively simple system consisting of only 4 particles embedded in a 2D space which are governed by an energy function between pairs of particles, while LJ-13 is given by the \emph{Leonnard-Jones} potential, consisting of 13 particles embedded in a 3D space. Both energy functions in DW-4 and LJ-13 satisfy $\mathrm{E}(3)$-equivariance. The fast-folding proteins dataset includes 12 structurally diverse proteins, such as Chignolin, Trp-Cage, and BBA. The simulations were conducted in explicit solvent, with frame spacing ranging from 100 $\mu s$ to 1 $ms$.

\emph{Methods:} 
As a multi-channel version of EGNN~\citep{satorras2021en}, GMN~\citep{huang2022equivariant} focuses specifically on the physical dynamics by considering the geometric constraints (such as chemical bonds) between atoms and achieves promising results on the MD simulation task in MD17. EGHN~\citep{han2022equivariant} develops an equivariant version of UNet~\citep{ronneberger2015u} equipped with equivariant pooling/unpooling layers to better reveal the hierarchy of large molecules such as proteins, leading to state-of-the-art performance on AdK dataset. NequIP~\citep{batzner20223} learns interatomic potentials and forces using high-order geometric tensors and $\mathrm{E}(3)$-equivariant convolution layers, achieving high data efficiency and quantum chemical level accuracy for MD17. By observing that GMN and other related geometric GNN methods only learn constant integration of the velocity, Newton–Cotes GNN~\citep{guo2023newton} predicts the integration based on several velocity estimations with Newton–Cotes formulas and proves its effectiveness theoretically and empirically. ESTAG~\citep{wu2023equivariant} reformulates dynamics simulation as a spatio-temporal prediction task by employing the trajectory in the past period to recover the Non-Markovian interactions. EGNO~\citep{xu2024equivariant} models the MD trajectory as a function over time using neural operators. SEGNO~\citep{liu2024segno} leverages the second-order continuity information to enhance the performance of 
GeoTDM~\citep{han2024geometric} further leverages the diffusion model to perform trajectory generation on molecular dynamics.

Considering the uncertainty of molecular dynamics at the quantum scale, some methods aim to fit the equilibrium distribution of molecules rather than predicting a single molecular conformation. By leveraging the continuous normalizing flows, E-CNF~\citep{kohler2020equivariant} predicts $\mathrm{SE}(3)$-equivariant molecular conformers through the invariant CoM prior density and equivariant vector fields, showing better generation capabilities compared to invariant flows. Later, E-ACF~\citep{midgley2024se} employs the augmented normalizing flow~\citep{huang2020augmented} to learn the target distribution of molecules from MD trajectories, which retains $\mathrm{SE}(3)$-equivariance by projecting the atomic Cartesian coordinates into the $\mathrm{SE}(3)$-invariant vector space. Furthermore, ITO~\citep{schreiner2023implicit} utilizes the score matching diffusion model for stochastic dynamics across multiple time-scales, with extended $\mathrm{SE}(3)$-equivariant PaiNN architecture \citep{schutt2021equivariant}, showcasing considerable generalization ability for different molecular scales.

\subsubsection{Molecular Generation}
Molecule generation plays a central role in drug discovery and material design. Its goal is to generate novel molecules with properties of interest by using machine learning. 

\emph{Task definition:} 
Basically, the methods for molecular generation learn a parametric probability distribution $p_{\theta}(\vec{\gG})$ from an observed dataset $\sD\coloneqq\{\vec{\gG}_i\}$. A novel molecular geometric graph is then sampled from the learned distribution:
\begin{align}
    \vec{\gG} \sim p_{\theta}(\vec{\gG}).
\end{align}
Instead of generating a whole geometric graph (namely de novo generation), there are part of methods investigating the conditional generalization paradigm by generating the 3D coordinates $\vec{\mX}$ given the 2D topological graph $\gG(\mH, \mA)$, forming the so-called conformation generation problem $\vec{\mX}\sim p_\theta(\vec{\mX}\mid \mH,\mA)$.

\emph{Symmetry preserved:} 
The generative model $p_{\theta}(\vec{\gG})$ should be E($3$)-invariant, \emph{i.e.}, $p_{\theta}(g\cdot\vec{\gG})=p_{\theta}(\vec{\gG}),\forall g\in\mathrm{E}(3)$. This is to ensure that the probability distribution is unaffected by the specific choice of the coordinate system to describe a molecule. In some methods as presented latter, $p_{\theta}(\vec{\gG})$ is marginalized from a joint distribution $p_{\theta}(\vec{\gG},\vec{\gG}^{(0)})=p_{\theta}(\vec{\gG}\mid \vec{\gG}^{(0)})p(\vec{\gG}^{(0)})$ where $p(\vec{\gG}^{(0)})$ denotes a certain initial distribution. In this scenario, the initial distribution $p(\vec{\gG}^{(0)})$ should be E($3$)-invariant and the likelihood distribution $p_{\theta}(\vec{\gG}\mid \vec{\gG}^{(0)})$ should be E($3$)-equivariant, to guarantee the E($3$)-invariance of $p_{\theta}(\vec{\gG})$~\citep{hoogeboom2022equivariant}. 

\emph{Datasets:}
QM9~\citep{ramakrishnan2014quantum} and GEOM~\citep{axelrod2022geom} are two prevailing datasets used for molecular generation. In particular, QM9 consisting of about 134K organic molecules contains the molecular 3D structures (\emph{e.g.} the coordinates of each atom in 3D space) and a wide range of chemical properties for each molecule. GEOM is a comprehensive dataset containing over 37 million molecular conformations, offering diverse conformation ensembles for each 2D molecular structure.

\emph{Methods:} 
Current methods can be divided into two classes: conformation generation and \emph{de novo} generation. 
Conformation generation is to generate 3D conformation given the 2D graph representation. Traditional methods~\citep{liberti2014euclidean} focus on the two-stage strategy: first predicting distances and then reconstructing coordinates, which yet could lead to unrealistic structures if the predicted distances are invalid. To avoid this issue, ConfVAE~\citep{xu2021an} reformulates the generation task as a bilevel optimization problem under the framework of VAE~\citep{kingma2013auto}, where the distance prediction and conformation generation are optimized jointly in an end-to-end manner. At the same time, ConfGF~\citep{shi2021learning} estimates the gradient fields of inter-atomic distances by using denoising score matching, and then samples the conformations via annealed Langevin dynamics. Later, DGSM~\citep{luo2021predicting} further extends ConfGF by modeling long-range interactions between non-bond atoms additionally. Instead of optimizing force field expensively, GeoMol~\citep{ganea2021geomol} predicts the local 3D geometries including bond distances and torsion angles simultaneously in an SE(3)-invariant way. Without predicting intermediate values like inter-atomic distances, DMCG~\citep{zhu2022direct} generates the 3D atomic coordinates by iteratively refining the initial coordinate predictions while accounting for invariance through its designed loss function. Due to the success of diffusion models, GeoDiff~\citep{xu2022geodiff} leverages graph field network to learn SE(3)-invariant distribution, and Torsional Diffusion~\citep{jing2022torsional} operates in torsion angle space rather than in Euclidean space.

As for de novo generation, a series methods have been proposed thanks to the fruitful progress of generative models~\citep{Wang2025DiffusionMF}. Built upon Schnet~\citep{schutt2018schnet}, G-SchNet~\citep{gebauer2019symmetry} introduces an autoregressive model to directly generate 3D molecular structures, while maintaining physical constraints. cG-SchNet~\citep{gebauer2022inverse} further extends G-SchNet to property-guided generation. Leveraging the generative capabilities of flow models, E-NFs~\citep{satorras2021enf} reformulates generation as the task of solving a continuous-time ODE, where the dynamics are predicted by EGNN~\citep{satorras2021en}. By harnessing the power of diffusion, EDM~\citep{hoogeboom2022equivariant} exploits $\mathrm{E}(3)$ equivariance by employing EGNN~\citep{satorras2021en} to enhance the diffusion process across both continuous and discrete features. GeoLDM~\citep{xu2023geometric} further maps the geometric features into the latent space where latent diffusion is performed. Rooted in EDM, EEGSDE~\citep{bao2023equivariant} formulates the generation process as an equivariant SDE and employs a meticulously designed energy function to guide the generation. Recently, MDM~\citep{huang2023mdm} takes into account inter-atomic forces at varying distances (\emph{e.g.} van der Waals forces), and injects variational noises to enhance performance for large molecules and improve generation diversity. To address atom-bond inconsistency problem, MolDiff~\citep{peng2023moldiff} introduces a joint atom-bond diffusion framework and bond guidance to make sure atoms are better suited for bonding. HierDiff~\citep{qiang2023coarse} adopts a hierarchical diffusion which first generates the coarse positions of molecular fragments and then fills in the fine-grained atomic geometry. EQUIFM~\citep{song2024equivariant} further explores de novo generation with flow matching, utilizing different probability paths for atom type and structure generation.

\subsubsection{Molecular Pretraining} 

Given that molecular labeling is expensive to obtain, pretraining molecular representation models without labels becomes fundamental and indispensable in real applications. These pretrained models can then be directly transferred or fine-tuned for specific downstream tasks, such as predicting binding affinity and molecular stability, thereby alleviating data scarcity and improving training efficiency. Previous research primarily focused on pretraining models utilizing non-geometric information, including SMILES notations~\citep{wang2019smiles}, chemical graphs~\citep{hu2020strategies}, functional groups~\citep{rong2020self}, etc. Recently, there has been a growing interest in self-supervised pretraining on the 3D geometric structure of molecules.

\emph{Task definition:}
Suppose $\phi_{\theta}(\vec{\gG})$ to be the representation model, and $\gL\left(\hat{\vy}(\vec{\gG}), \phi_{\theta}(\vec{\gG})\right)$ to be the self-supervised training objective where  $\hat{\vy}(\vec{\gG})$ denotes the pseudo label created based on the structure of $\vec{\gG}$. The representation model is optimized to minimize the self-supervised objective as
\begin{align}
    \theta = \argmin_{\theta}\gL\left(\hat{\vy}(\vec{\gG}), \phi_{\theta}(\vec{\gG})\right).
\end{align}

\emph{Symmetry preserved:} 
The representation model $\phi_{\theta}(\vec{\gG})$ is E($3$)-equivariant if $\hat{\vy}(\vec{\gG})$ is a steerable vector, and is E($3$)-invariant if $\hat{\vy}(\vec{\gG})$ consists of scalars.

{
\emph{Datasets:} 
PCQM4Mv2~\citep{hu2020ogb} is a comprehensive quantum chemistry dataset consisting of 3.37 million molecules derived from the OGB benchmark, which was originally curated as part of the PubChemQC project~\citep{doi:10.1021/acs.jcim.7b00083}.  QM9~\citep{ramakrishnan2014quantum} is another popular dataset that encompasses quantum chemistry structures and properties, featuring 134K molecules. QMugs\citep{isert2022qmugs} expands QM9 by offering a more extensive collection of drug-like molecules, totaling 665K molecules. GEOM~\citep{axelrod2022geom} is an energy-annotated molecular conformation dataset containing 37 million molecular conformations sourced from multiple datasets, such as QM9 and CREST program~\citep{pracht2020automated}. Uni-Mol~\citep{zhou2023unimol} constructs a conformation dataset containing 19 million molecules. It utilizes ETKGD with Merck Molecular Force Field optimization in RDKit to generate 11 random conformations for each molecule, resulting in a total of 209 million conformations.
}

\emph{Methods:} 
A variety of studies investigate the denoising objective, pretraining the model by recovering the original signal from a perturbed input.  Specifically, GeoSSL-DDM~\citep{liu2023molecular} formulates the denoising objective based on atomic distance.  Uni-Mol~\citep{zhou2023unimol} proposes position denoising and joint training between 3D molecular conformations and candidate protein binding pockets. 
GNS-TAT~\citep{zaidi2023pretraining} establishes a connection between coordinate denoising and the potential energy of molecular conformations. MGMAE~\citep{feng2022mgmae} proposes a reconstruction strategy to train on the heterogeneous atom-bond graph with a high mask ratio. 3D-EMGP~\citep{jiao2022energy} further proposes to predict the atomic pseudo force field which is estimated by an Riemann-Gaussian denoising distribution to ensure $\mathrm{E}(3)$-invariant pretraining loss. Apart from the denoising objective, GraphMVP~\citep{liu2022pretraining} leverages the correlation between 2D molecular graphs and 3D conformations, constructing a contrastive objective for the model pretraining. Similar to GraphMVP, Transformer-M~\citep{luo2023one} leverages positional encodings and attention biases to encode the 2D and 3D structures in one Transformer model. Meanwhile, 3D-Infomax~\citep{stark20223d} exploits this correspondence by attempting to maximize the mutual information between 2D molecular graph embeddings and learned representations of the corresponding 3D graphs. MoleculeSDE~\citep{pmlr-v202-liu23h} extends 3D-Infomax~\citep{stark20223d} and leverages group symmetric stochastic differential equation models to establish a connection between 3D geometries and 2D topologies, with a tighter MI bound. Frad~\citep{feng2023fractional} decomposes molecules into fragments to fix the rigid parts and pretrains the model via denoising on the flexible parts. SliDe~\citep{ni2023sliced} explores pretraining with denoising from a distribution that encodes physical principles. DenoiseVAE~\citep{liu2025denoisevae} utilizes a learnable noise generation strategy to adaptively acquire atom-specific noise distributions for different molecules, which results in more accurate force field learning.

\subsection{Tasks on Proteins}
\label{sec:Protein}
Proteins are large biomolecules that are composed of one or more long chains of amino acid residues. All proteinogenic amino acids share common structural features, including an $\alpha$-carbon to which an amino group, a carboxyl group, and a variable side chain are bonded. Most proteins fold into unique 3D structures that determine the function and activity of proteins in biological processes. Owing to the hierarchical structures of proteins, there are mainly two different ways to leverage geometric graph $\vec{\gG}$ to represent proteins. For one thing, we can treat each residue as a node, the positions of $\alpha$-carbons as the coordinate matrix $\vec{\mX}$ and the residue-level features as $\mH$. 
For another thing, we can apply the full-atom setting by considering each atom as a node, the positions of all atoms as $\vec{\mX}$ and atom-level features as $\mH$. In both ways, the edges can be created via either the chemical bonds or cut-off distances. There are plenty of works that develop machine learning methods to process proteins. While some of them focus on 1D residue sequences, this survey is mainly interested in the study of 3D structures and will demonstrate several relevant tasks in the following. 

\subsubsection{Protein Property Prediction} 

Similar to molecular property prediction, protein property prediction is a crucial $\mathrm{E}(3)$-invariant task in computational biology. 
Most previous works solely employ residue sequences to predict protein properties. Thanks to the development of geometric structure modeling, more and more attentions are paid to using geometric GNNs to estimate the functional property of proteins via exploring 3D structures. In terms of the prediction granularity, the task of protein property prediction is classified as protein-level, residue-level and atom-level prediction, with the details provided below. 

\textbf{Protein-Level Prediction}: Many tasks aim to predict the functions or certain scores given the protein structure. \textbf{(1) Enzyme Commission (EC)} number prediction~\citep{gligorijevic2021structure} is a prevailing protein-level classification task which aims to predict the catalyzed reaction class of the given enzyme. \textbf{(2) Gene Ontology (GO)} term prediction~\citep{gligorijevic2021structure} seeks to predict the functional classes concerning gene ontology given the protein structure, whose data is usually split into three tracks: molecular function (MF), biological process (BP), and cellular component (CC). \textbf{(3) Protein Structure Ranking} learns a quality score function of the given protein structure to estimate the structural similarity between the candidate protein and the native structure. It plays a vital role in computational biology, as it assists researchers in pinpointing the most accurate or biologically significant protein conformations from a collection of potential structures.
\textbf{(4) Protein Localization Prediction} targets at forecasting the subcellular locations of proteins~\citep{almagro2017deeploc}, which is essential to understand the function of a protein and helps investigate the pathogenesis of many human diseases~\citep{hung2011protein}. 
\textbf{(5) Fitness Landscape Prediction} primarily focuses on the prediction of the effects of residue mutations on the fitness of proteins. Typical target functions include $\beta$-lactamase~\citep{xu2022peer}, Adeno-Associated Virus (AAV), Thermostability~\citep{dallago2021flip} and Fluorescence and Stability~\citep{rao2019evaluating}. 

Abundant protein-level representation models are available in existing literature. DeepFRI~\citep{gligorijevic2021structure} and LM-GVP~\citep{wang2022lm} propose a two-stage architecture, which adopts language models to extract amino acid sequence information and graph-based model to learn the interactions between amino acids simultaneously. Notably, LM-GVP utilizes equivariant model GVP~\citep{jing2021learning} as the graph-based model. GearNet~\citep{zhang2022protein} proposes a relational graph convolution layer to better capture the 3D geometry of proteins, and exploits multi-view contrastive pretraining to better utilize unlabeled data. As for structure ranking, TM-Align~\citep{zhang2005tm} is a typical but not DL-based method, which is time-consuming. Thanks to the expressive ability of geometric GNN, ~\citep{jing2021learning,eismann2023protein,chen20233d} adopt equivariant GNN models such as GVP~\citep{jing2021learning} and TFN~\citep{thomas2018tensor} to fulfill model quality assessment (MQA). In addition, TFN~\citep{thomas2018tensor} is also used for ranking protein-protein complex in PAUL~\citep{eismann2020hierarchical}.

\textbf{Residue-Level Prediction}: Atom3D~\citep{townshend2021atomd} proposes Residue Identity (RES) prediction, which aims to predict the amino acid types at the center of a given local context. The performance on this task measures whether a model can capture the structural dependencies between individual amino acids, which is vital for protein engineering.

\textbf{Atom-Level Prediction}: The main form of atom-level prediction lies in pocket detection, which requires predicting whether an atom on the protein belongs to the binding site in terms of a potential ligand. Previous methods usually design algorithms to find and rank the cavities on the protein surface~\citep{krivak2015improving, le2009fpocket}, or voxelize the protein structure and use 3D-CNN for supervised training~\citep{jimenez2017deepsite, mylonas2021deepsurf}. Notably, a series of works are exploiting the geometric GNNs to achieve much better performance (ScanNet~\citep{tubiana2022scannet}, EquiPocket~\citep{zhang2023equipocket}, PocketMiner~\citep{meller2023predicting}).

\subsubsection{Protein Generation}

In terms of what to generate, the approaches for protein generation are categorized into protein folding (or protein structure prediction), protein inverse folding, and protein structure and sequence co-design.

\textbf{Protein Folding}  aims to generate folding structures given the amino acid sequences of the input protein. This task has significant implications in the field of drug design. The folding structure is generated by:
\begin{align}
    \vec{\mX}\sim p_\theta(\vec{\mX}\mid \vs),
\end{align}
where $\vs\in\R^{N}$ denotes the amino acid sequence based on the coordinates of all residues $\vec{\mX}\in\R^{N\times 3}$ (note that each row of $\vec{\mX}$ can include more than one 3D coordinate vector if full-atom coordinates are considered).

\emph{Symmetry preserved:} 
This is an equivariant task, implying that $p_\theta(\vec{\mX}\mid\vs)=p_\theta(\vec{\mX}\mO+\vec{\vt}\mid\vs)$ for an arbitrary orthogonal transformation $\mO$ and translation $\vec{\vt}$. Notably, some methods generate the distance matrix or other invariant forms of $\vec{\mX}$, reducing the task into a trivial generation problem without the equivariance constraint. 

\emph{Methods:} 
The AlphaFold series~\citep{senior2020improved,AlphaFold2021} and RoseTTAFold series~\citep{minkyung2021rosettafold,baek2023efficient} represent the forefront of contemporary techniques in protein folding. They employ a sophisticated multi-track architecture capable of processing multi-sequence alignments (MSA), amino acid pair-wise distance maps, and geometric structures, each with remarkable efficiency. Building upon these advancements, RoseTTAFold2~\citep{baek2023efficient} extends the capabilities of both AlphaFold2~\citep{AlphaFold2021} and RoseTTAFold~\citep{minkyung2021rosettafold} by refining the attention mechanism and enhancing the three-track architecture, resulting in notable performance improvements. Moreover, RFAA~\citep{krishna2023generalized} further extends RoseTTAFold's versatility to encompass the design of various biomolecules beyond proteins, including nucleic acids, small molecules, and metals.
In contrast, ESMFold~\citep{lin2022language} and HelixFold-Single~\citep{fang2023method} represent a departure from traditional methods by eschewing the requirement for MSA. Instead, it learns to predict protein structures directly from primary sequence data, significantly enhancing inference efficiency. Additionally, EigenFold~\citep{jing2023eigenfold} introduces a novel harmonic diffusion process that projects protein structures onto eigenmodes, thereby preventing the disassembly of adjacent nodes. 

\textbf{Protein Inverse Folding} aims to generate  amino acid sequences conditional on the folding structures of the input protein. Using the same denotations as the task of protein folding, the model $p_\theta$ generates the amino acid sequence $\vs\in\R^{N}$ of interest:
\begin{align}
    \vs\sim p_\theta(\vs\mid\vec{\mX}).
\end{align}

\emph{Symmetry preserved:} This is an invariant task, indicating that $p_\theta(\vs\mid\vec{\mX})=p_\theta(\vs\mid\vec{\mX}\mO+\vec{t})$ for an arbitrary orthogonal transformation $\mO$ and translation $\vec{t}$.

\emph{Methods:} Typical methods such as~\cite{ingraham2019generative} and~\cite{tan2022generative} take the invariant features including distance and dihedral angles as input, to ensure invariance during generation. More recently, based on GVP~\citep{jing2021learning} that is $\mathrm{E}(3)$-equivariant, ESM-IF~\citep{hsu2022learning} further incorporates more structure information for the generation, while keeping the output sequence invariant. Similarly, LM-Design~\citep{pmlr-v202-zheng23a} integrates structural embedding into language models to improve the performance of inverse folding. ProteinMPNN~\citep{dauparas2022robust} uses an invariant architecture to embed its backbone and predicts amino acid probabilities autoregressively while enforcing desired constraints. PiFold~\citep{gao2022pifold} additionally incorporate distance, angle, and direction features and proposes PiGNN to non-autoregressively generate the sequences. KW-Design~\citep{gao2023kw} integrates knowledge from pretrained sequence and structure models to refine the sequences generated by the baselines with a memory retrieval mechanism.

\textbf{Protein Structure and Sequence Co-Design} aims to generate both the amino acid sequences and folding structures, which is formally derived as: 
\begin{align}
    \vec{\mX}, \vs\sim p_\theta(\vec{\mX},\vs).
\end{align}

\emph{Symmetry preserved:} 
Clearly, this task is invariant with respect to $\vs$, and equivariant with respect to $\vec{\mX}$.

\emph{Methods:} 
Based on RoseTTAFold~\citep{minkyung2021rosettafold}, RFdiffusion~\citep{watson2023novo} incorporates Gaussian noise into coordinates and Brownian motion noise into orientations, subsequently denoises the structure step-by-step and recovers sequence using ProteinMPNN~\citep{dauparas2022robust}. Meanwhile, Chroma~\citep{ingraham2023illuminating} introduces a revolutionary programmable diffusion framework, empowering diverse conditional generation and precise targeting of properties through constraints such as symmetry, shape, and semantics. Both Chroma and RFDiffusion begin with structure generation and then conduct the subsequent sampling of the corresponding sequence through another module. Unlike these two works, PROTSEED~\citep{shi2023protein} designs the structure and sequence jointly by an encoder-decoder framework, where the encoder is trigonometry-aware to learn context features and the decoder is SE(3)-equivariant to express the sequence and structure.

\emph{Datasets:}
ATOM3D~\citep{townshend2021atomd} constructs multiple widely-used datasets tailored for protein design tasks. CASP~\citep{kryshtafovych2019critical} stands out as a renowned contest dedicated to protein structure prediction. In this competition, participants submit predicted structures for evaluation, particularly when the experimental structures are not publicly available. The community then assesses the quality of these submissions. Additionally, AlphaFoldDB~\citep{varadi2022alphafold}, SCOPe~\citep{chandonia2019scope} and  CATH~\citep{orengo1997cath} serve as valuable resources for protein design, providing datasets comprising protein structures alongside their corresponding sequences. SCOPe and CATH consist of segmented protein structure domains, while AlphaFoldDB boasts a repository of over 200 million complete structures predicted by AlphaFold2~\citep{AlphaFold2021}.
Moreover, with predictions stemming from ESMFold~\citep{lin2022language}, the ESM Metagenomic Atlas boasts a collection of about 772 million metagenomic protein structures.

\subsubsection{Protein Pretraining}

Similar to molecule pretraining task, protein pretraining also aims to learn representations of protein, which can be used in downstream tasks.

\emph{Task definition:} Generally, each input protein is modeled as a geometric graph $\vec{\gG}$ and the pretraining purpose is to learn a parametric model $\phi_{\theta}$ which can output high-quality representations $\mH\in \sR^{N\times d}$ of the input protein:
\begin{align}
    \mH = \phi_{\theta}(\vec{\gG}).
\end{align}

\emph{Symmetry preserved:}
It is equivariant for the output vectors in $\mH $, and invariant for the output scalars in $\mH$.

\emph{Datasets:}
For protein sequence pretraining methods, UniProt~\citep{uniprot2023uniprot} functions as a central repository for both protein sequence and functional information. It is organized into clusters by UniRef~\citep{suzek2015uniref}, with pairwise sequence identity thresholds typically set at 50\% and 100\% (referred to as UniRef50 and UniRef100) to eliminate redundancy. BFD~\citep{steinegger2018clustering}, on the other hand, represents a larger sequence dataset, formed by amalgamating UniProt with protein sequences sourced from metagenomic sequencing projects. Furthermore, NetSurfP-2.0~\citep{klausen2019netsurfp} furnishes labels for protein secondary structure prediction, delineated into 3-states and 8-states, offering valuable resources for supervised training.
In the realm of protein structure pretraining and classification, SCOPe~\citep{chandonia2019scope}, CATH~\citep{orengo1997cath}, and AlphaFoldDB~\citep{varadi2022alphafold} hold significant importance. They provide comprehensive repositories for protein structures, facilitating research and advancement in the field.

\emph{Methods:} 
Previous protein pretraining methods such as ESM-1b~\citep{rao2020transformer}, ESM2~\citep{lin2022language}, ProtTrans~\citep{elnaggar2021prottrans}, xTrimoPGLM~\citep{chen2024xtrimopglm} and ProtGPT2~\citep{ferruz2022protgpt2}, are based on sequence masking and prediction, inspired by the success of NLP language models. Readers can refer to the survey by~\cite{wu2022survey} for more introductions of protein language models. Recent attentions have been paid to pretrained models based on the 3D structure information. For instance, GearNet~\citep{zhang2022protein} built upon an invariant GNN with multi-type message passing leverages several pretraining objectives including contrastive learning between sequences and structures, distance/dihedral prediction, and residue type prediction. Other works like ProFSA~\citep{gao2023self} and DrugCLIP~\citep{gao2023drugclip} also utilize contrastive learning to learn SE(3)-invariant features, but focusing more on pocket pretraining, where the pocket-ligand interaction knowledge is incorporated as well. \cite{guo2022self} employs pretraining with the protein's tertiary structure, incorporating SE(3)-invariant features to ensure the efficient preservation of SE(3)-equivariance. PAAG~\citep{yuan2024annotation} enables multi-level alignment between protein sequence and textual annotation to capture the fine-grained motif inside the protein and successfully designs  proteins with functional domains.

\subsection{Tasks on Mol+Mol}
This subsection introduces the tasks with the input of ``molecule+molecule'', including liker design and chemical reaction prediction. 

\subsubsection{Linker Design}
Fragment-based molecule design requires to predict the linker, a small molecule, so that two or more molecular components can be combined into novel molecules with desirable properties. Linkers are of great importance in maintaining the proper orientation, flexibility, and stability of multi-domain proteins or fusion proteins.

\emph{Task definition:} The input consists of two or more unlinked molecular fragments, which are all represented as geometric graphs $\{\vec{\gG}_{i}\}_{i=1}^k$, and the model needs to learn an equivariant function $f_\theta$ whose output is a small molecule $\vec{\gG}_{L}$ used to link the fragments. Specifically, 
\begin{align}
    \vec{\mX}_L, \mH_L = f_\theta(\vec{\gG}_{1}, \vec{\gG}_{2},\ldots,\vec{\gG}_{k}).
\end{align}

\emph{Symmetry preserved:} If we impose rotation or translation operations on the input fragments simultaneously, the output coordinates should transform correspondingly while the atom features keep invariant.

\emph{Datasets:}
The linkers connecting molecules in ZINC~\citep{sterling2015zinc} can be computationally synthesized, similar to the methods employed by \cite{hussain2010computationally}. Conversely, CASF\citep{su2019comparative} offers experimentally validated molecules for linker design. In contrast to ZINC and CASF, which typically produce paired fragments, DiffLinker~\citep{igashov2024equivariant} generates a novel dataset comprising three or more fragments, drawing from GEOM~\citep{axelrod2022geom}.

\emph{Methods:} DeLinker~\citep{imrie2020deep} and 3DLinker~\citep{huang20223dlinker} employ VAE~\citep{kingma2013auto} to create the 3D structure of a linker. However, their capability is limited to linking only two fragments, rendering them ineffective when faced with an arbitrary number of fragments to link. In contrast, DiffLinker~\citep{igashov2024equivariant} has recently succeeded in addressing this challenge by harnessing an E($3$)-equivariant diffusion model configured to handle multiple fragments.

\subsubsection{Chemical Reaction Prediction}

In chemical reactions, identifying and characterizing transition state (TS) structures is crucial for understanding reaction mechanisms. This process entails locating the TS structure that minimizes the system's potential energy (PE) while adhering to specific constraints, such as SE(3) invariance.

\emph{Task definition: } Given a reactant $\vec{\gG}_\text{R}$ and a product $\vec{\gG}_\text{P}$, the objective is to generate the TS structure $\vec{\gG}_{\text{TS}}$ that optimizes the following objective: 
\begin{align}
    \vec{\gG}_{\text{TS}}^* = \argmin_{\vec{\gG}_{\text{TS}}}  \text{PE}(\vec{\gG}_{\text{TS}}|\vec{\gG}_\text{R}, \vec{\gG}_\text{P}),
\end{align}
where the function $\text{PE}(\cdot)$ returns the potential energy.

\emph{Symmetry preserved:} In general, the output, namely, the TS structure is invariant to any independent transformation (\emph{e.g.} rotation) imposed to each of the input structure. If the input and output are always fixed within the same 3D coordinate space, then this task is equivariant, namely, imposing the same transformation to the two input structures, the output TS is transformed in the same way. 

\emph{Datasets: } 
TSNet~\citep{jackson2021tsnet} has meticulously assembled a dataset called $\text{S}_\text{N}\text{2-TS}$, which contains structures of reactants, transition states (TS), and products pertinent to $\text{S}_\text{N}\text{2}$ reactions. Transition1x~\citep{schreiner2022transition1x} provides a resource of 9.6 million density functional theory (DFT) calculations encompassing forces and energies for molecular configurations across reaction pathways. This extensive dataset offers valuable information for training models for reaction prediction.

\emph{Methods: }
OA-ReactDiff~\citep{duan2023accurate} introduces a diffusion model to generate transition state (TS) structures. This model ensures SE(3)-equivariance of the score function by constructing local frames. Moreover, the equivariant backbone model is adapted to accommodate multiple objects. On the other hand, TSNet~\citep{jackson2021tsnet} employs the equivariant graph neural network (GNN) model TFN~\citep{thomas2018tensor} to predict TS structures. Initially, TFN is pretrained on extensive chemical data, such as QM9~\citep{ramakrishnan2014quantum}, to learn useful representations. It is then fine-tuned specifically for the task of predicting transition structures.

\subsection{Tasks on Mol+Protein}
The ``molecule+protein" tasks are well explored, such as ligand binding affinity prediction, protein-ligand docking, and pocket-based molecule sampling.

\subsubsection{Ligand Binding Affinity Prediction}
The task of predicting ligand binding affinity revolves around estimating the interaction strength between a protein (receptor) and a small molecule (ligand)~\citep{kong2023generalist}. Accurate predictions in this area offer significant advantages for designing and refining drug candidates. Additionally, they aid in prioritizing compounds for experimental evaluation, thereby streamlining the drug discovery process. 

\emph{Task definition:} With both the molecule and protein regarded as geometric graph $\vec{\gG}_m, \vec{\gG}_p$, the task aims to learn an efficient predictor $\phi_\theta$, which can predict the binding strength $s$ accurately:
\begin{align}
    s=\phi_\theta(\vec{\gG}_p, \vec{\gG}_m).
\end{align}

\emph{Symmetry preserved:} It is obvious that the binding affinity will not change under any transformation.

\emph{Datasets: } 
CrossDocked2020~\citep{francouer2020three} contains over 22 million posed ligand-receptor complexes and the corresponding binding affinity values, which are generated by docking ligands into multiple receptor structures from the same binding pocket. PDBbind~\citep{liu2017forging} provides accurate and reliable binding affinity data, allowing researchers to assess how well computational methods can predict the strength of binding between proteins and ligands.

\emph{Methods:} MaSIF~\citep{gainza2020deciphering} utilizes geodesic space to represent the protein surface, assigns geometric and chemical features to patches, and employs rotation invariance to process these features, facilitating predictions of protein-ligand interactions. ProtNet~\citep{wang2022learning} considers 3D protein presentations at various levels (\emph{e.g.}, amino-acid level, backbone level, and all-atoms level) to accomplish affinity prediction tasks. GET~\citep{kong2023generalist} extends this concept by unifying different levels universally for both molecule and protein representations. TargetDiff~\citep{guan2023d} introduces a diffusion process that gradually adds noise to coordinates and atom types. This process, guided by an SE(3)-equivariant graph neural network (GNN), incorporates binding free energy terms to steer generation towards high-affinity poses. HGIN~\citep{zhao2023geometric} constructs a hierarchical invariant graph model to predict changes in binding affinity resulting from protein mutations. BindNet~\citep{feng2023protein} designs two pretraining tasks utilizing Uni-Mol~\citep{zhou2023unimol} as the encoder to jointly learn protein and ligand interactions.

\subsubsection{Protein-Ligand Docking}
\label{sec:dpp}
This task works towards predicting the transformation, \emph{e.g.}, rotation and translation, imposed on protein and molecules so that they can dock together with the minimum root-mean-square-deviation.

\emph{Task definition:} %
Without loss of generality, we assume that the protein remains fixed while the position of the molecule transforms. 
By denoting the protein as $\vec{\gG}_p\coloneqq(\vec{\mX}_p,\mH_p)$ and the molecule as $\vec{\gG}_m\coloneqq(\vec{\mX}_m,\mH_m)$, respectively, the model needs to learn a prediction function $\phi_\theta$ that outputs the rotation matrix and translation vector (\emph{i.e.}, $\mR, \vec{\vt}$) by
\begin{align}
    \mR, \vec{\vt}=\phi_\theta(\vec{\mX}_p,\mH_p; \vec{\mX}_m, \mH_m).
\end{align}
With the predicted rotation $\mR$ and translation $\vec{\vt}$, we can dock the molecule towards the fixed protein. 

\emph{Symmetry preserved:} To make the final docked complex to be SE($3$)-equivariant, the predictor $\phi$ is supposed to meet the following independent SE($3$) constrains~\citep{stark2022equibind}:
\begin{align}
\label{eq:dpp}
    \mR'&=\mQ_m\mR\mQ_p^\top, \vec{\vt'}=\mQ_m\vec{\vt}-\mQ_m\mR\mQ_p^\top \vec{\vt}_p+\vec{\vt}_m, \\ 
    \nonumber
    \forall & \mQ_p,\mQ_m\in\text{SO}(3), \vec{\vt}_p, \vec{\vt}_m\in \sR^{3},
\end{align}
where $\mR', \vec{\vt'}$ are the predicted rotation matrix and translation vector after transforming the protein and the molecule, namely, $\mR', \vec{\vt'}=\phi_\theta(\vec{\mX}_p\mQ_p+\vec{\vt}_p,\mH_p; \vec{\mX}_m\mQ_m+\vec{\vt}_m, \mH_m)$.

\emph{Datasets:}
PDBbind~\citep{liu2017forging} stands out as the predominant dataset for Protein-Protein Docking, housing over 22 million poses resulting from the docking of ligands into their respective receptor structures. Typically, current methods segment the dataset based on chronological order, leveraging this organization for training and evaluation purposes.

\emph{Methods:} EquiBind~\citep{stark2022equibind} and TankBind~\citep{lu2022tankbind} have tackled the blind binding problem by leveraging equivariant graph neural networks. TankBind additionally introduces trigonometry constraints to enhance compound rationality. To further enhance performance, DiffDock~\citep{corso2023diffdock} proposes a diffusion process operating across three groups (T(3), SO(3), and SO(2)). In contrast, DESERT~\citep{long2022zero} offers a unique approach by initially outlining pocket shapes and then generating molecule structures to bind these pockets. This method alleviates the scarcity of experimental binding data and isn't reliant on predefined pocket-drug pairs. Recently, FABind~\citep{pei2023fabind} designs geometry-aware GNN layers and efficient interaction modules (\emph{e.g.} interfacial message passing) to unify pocket prediction and the docking stage, which leads to fast and accurate prediction. Further, Re-Dock~\citep{huang2024re} explores flexible docking by considering the gap between apo and holo conformations of the target protein, which enhances the practical utility.
 
\subsubsection{Pocket-Based Mol Sampling}
The technique of pocket-based molecular sampling aims at generating small molecules that have the potential to bind to a particular pocket on a protein or other biomolecular target. 

\emph{Task definition:} This target-aware design resorts to learn a generation model $p_\theta$ whose output is a new molecule $\vec{\gG}_m$ that can bind to a specific pocket $\vec{\gG}_p$:
\begin{align}
    \vec{\gG}_m \sim p_\theta(\vec{\gG}_m\mid\vec{\gG}_p).
\end{align}

\emph{Symmetry preserved:} It is an equivariant problem, implying that $p_\theta(\vec{\gG}_m\mid\vec{\gG}_p)=p_\theta(g\cdot\vec{\gG}_m\mid g\cdot\vec{\gG}_p)$ for any transformation $g$ of interest.

\emph{Datasets:}
CrossDocked2020~\citep{francouer2020three} serves as a substantial resource for sampling molecules based on docking pockets, containing approximately 22.5 million docked protein-ligand pairs.

\emph{Methods:} Pocket2Mol~\citep{peng2022pocket}, GraphBP~\citep{liu2022generating}, 
SBDD~\citep{luo2021a}, and FLAG~\citep{zhang2023molecule} adopt an autoregressive approach to generate molecules conditioned on binding sites, operating at the granularity of atoms or motifs. In contrast, TargetDiff~\citep{guan2023d} with a series following diffusion-based methods~\citep{lin2023d3fg, lin2022diffbp, pinheiro2024structure, qu2024molcraft, zhang2023molecule} diverges from this method by utilizing 3D equivariant diffusion in a non-autoregressive fashion. This approach enables the generation of all atoms simultaneously, resulting in higher efficiency. DESERT~\citep{long2022zero} further explores to first sketch the shape of the molecule according to the pocket, and then generates a molecule fitting in the shape. D3FG~\citep{lin2024functional} leverages a fragment-based diffusion to enhance the generative performance by decomposing molecules into functional groups and linkers.

\subsection{Tasks on Protein+Protein}
The ``protein+protein" tasks include protein interface prediction, protein-protein binding affinity prediction, protein-protein docking, antibody design that considers specifically the interaction between antibodies and antigens, and peptide design that aims at generating target-specific peptide.

\subsubsection{Protein Interface Prediction}
Biological processes often depend on interactions between biomolecules. This creates a need for predicting protein-protein interfaces, which involves identifying the regions on a protein's surface that are likely to participate in interactions with other proteins.

\emph{Task definition:} With the protein pair taken as two geometric graphs $\vec{\gG}_1, \vec{\gG}_2$, this task requires to learn a predictor $\phi_\theta$ that determines if the atoms on the protein belong to the interface. The output are interpreted as the atomic probabilities $\vp\in\sR^{N_1+N_2}$ of being located on the interface:
\begin{align}
    \vp=\phi_\theta(\vec{\gG}_1, \vec{\gG}_2).
\end{align}

\emph{Symmetry preserved:} Once the interaction proteins are selected, the atoms in the interface are deterministic no matter the rigid transformations on each partner, resulting in an invariant problem with respect to each protein:
\begin{align}
    \phi(\vec{\gG}_1, \vec{\gG}_2) = \phi(g_1\cdot \vec{\gG}_1, g_2\cdot \vec{\gG}_2), \forall g_1, g_2 \in \text{SE}(3).
\end{align}

\emph{Methods:} The methods dMaSIF~\citep{sverrisson2021fast} and SASNet~\citep{townshend2019end} operate via three-dimensional convolution on the protein 3D structures to keep rotation-invariance. Moreover, fed with more structure features such as distance, orientation and amide angle, DeepInteract~\citep{morehead2022geometric} adopts geometric transformer and achieves competitive performance as well.

\subsubsection{Binding Affinity Prediction}
Protein-protein interactions are fundamental to
bio-molecular activity and are crucial for many key functions in biological processes. Estimating the binding affinity between proteins not only aids in gaining a deeper understanding of protein mechanisms of action but also serves as the cornerstone for designing proteins with specific functions, such as highly specific antibodies and high-affinity ligands.

\emph{Task definition:} Given a pair of proteins that can be considered as geometric graphs $\vec{\gG}_1, \vec{\gG}_2$, this task requires learning a predictive function $\phi_\theta$, which can efficiently and accurately predict the binding strength $s$ between the pair of proteins:
\begin{align}
    s=\phi_\theta(\vec{\gG}_1, \vec{\gG}_2).
\end{align}

\emph{Symmetry preserved:} This is an invariant task because the binding strength $s$ remains unchanged under any translations or rotations applied to the pair of proteins.

\emph{Datasets: } 
PDBbind~\citep{wang2004pdbbind} dataset constitutes an assembly of complex structures, meticulously sourced from the Protein Data Bank (PDB), accompanied by binding affinities that have been quantified through rigorous experimental methods. Protein-Protein Affinity Benchmark Version 2~\citep{kastritis2011structure,vreven2015updates} encompasses a repertoire of 176 variegated protein-protein complexes, each accompanied by detailed affinity annotations. SKEMPI(Structural database of Kinetics and Energetics of Mutant Protein Interactions)~\citep{moal2012skempi} constitutes a curated database that delineates alterations in binding affinities and kinetic parameters consequent to mutagenesis. SKEMPI 2.0~\citep{jankauskaite2019skempi} represents the refined and augmented edition of the original SKEMPI database.

\emph{Methods:} mmCSM-PPI~\citep{rodrigues2021mmcsm} presents a binding affinity prediction method employing graph-based signatures that encapsulate protein structure's physico-chemical and geometric properties, augmented with complementary features to reflect various mechanisms. The Extra Trees model, trained with graph-based signatures and complementary features, yields promising results on the SKEMPI 2.0 dataset. GeoPPI~\citep{liu2021deep} utilizes the 3D conformations to ascertain a geometric representation that embodies the topological features of the protein structure through a self-supervised learning approach. Subsequently, these representations serve as inputs for gradient-boosting trees, facilitating the prediction of the variations in protein-protein binding affinity due to mutations. GET~\citep{kong2023generalist} introduces a bilevel design that ensures equivariance while unifying representations across different levels. GET achieves state-of-the-art performance in PDB dataset.

\subsubsection{Protein-Protein Docking}
We have investigated docking pose prediction between protein and molecule in~\cref{sec:dpp}. Here, we study the similar problem between protein and protein.

\emph{Task definition:} Assuming two proteins to be denoted as $\vec{\gG}_1=(\vec{\mX}_1,\mH_1), \vec{\gG}_2=(\vec{\mX}_2,\mH_2)$, respectively, the model needs to learn a prediction function $\phi_\theta$ to output the rotation matrix and translation vector (\emph{i.e.}, $\mR, \vec{\vt}$) by
\begin{align}
    \mR, \vec{\vt}=\phi_\theta(\vec{\mX}_1,\mH_1; \vec{\mX}_2, \mH_2).
\end{align}

\emph{Symmetry preserved:} This is identical to
~\cref{eq:dpp}.

\emph{Methods:} Equidock~\citep{ganea2022independent} uses
SE(3)-equivariant graph neural networks and optimal transport techniques to predict the transformation by aligning key points. HMR~\citep{wang2023learning} casts this task from 3D Euclidean space to 2D Riemannian manifold, keeping rotational invariant. DiffDock-PP~\citep{ketata2023diffdock} extends DiffDock~\citep{corso2023diffdock}, a diffusion generative model, to protein docking task and yields the state-of-the-art performance. Furthermore, in dMaSIF\citep{sverrisson2022physics}, an energy-based, SE(3)-equivariant model combined with physical priors is adopted to infer docking regions. Treating docking as an optimization problem, EBMDock~\citep{wu2024neural} employs geometric deep learning to extract features from protein residues and learns distance distributions between the residues involved in interfaces. Multimetric protein docking can be tackled by AlphaFold-Multimer~\citep{evans2021protein} and SyNDock~\citep{ji2023syndock}. Recently, ElliDock~\citep{yu2024rigid} predicts $\mathrm{SE}(3)$-equivariant elliptic paraboloids as the binding interface for protein pairs, and transfers the rigid protein-protein docking task into surface fitting while ensuring the same degree of freedom. There are also several works targeting at antibody-antigen docking, a subfield of protein docking. For instance, HSRN~\citep{jin2022antibody} proposes a hierarchical framework to handle docking in an iterative manner. By harnessing the capabilities of tFold-Ab~\citep{wu2024fast} and AlphaFold2~\citep{AlphaFold2021}, tFold-Ag~\citep{wu2024fast} generates antibody/antigen features and employs a docking module to predict complex structures with flexibility. 

\subsubsection{Antibody Design}
Antibodies are Y-shaped symmetric proteins produced by the immune system that recognize and bind to specific antigens. The design of antibodies mainly focuses on the variable domains consisting of a heavy chain and a light chain, with 3 Complementarity-Determining Regions (CDRs) and 4 framework regions interleaving on each chain. The 6 CDRs largely determine the binding specificity and affinity of the antibodies, especially CDR-H3 (\emph{i.e.} the 3rd CDR on the heavy chain), which is the main scope of the design.

\emph{Task definition:} Without loss of generality, we define the task as a conditional variant of structure and sequence co-design. More specifically, given the geometric graphs of the antigen $\vec{\gG}_A$, the heavy chain $\vec{\gG}_H$, and the light chain $\vec{\gG}_L$ with the CDRs missing, the model $\phi_\theta$ needs to fill in the geometric graph of the CDRs of interest $\vec{\gG}_C$:
\begin{align}
    \vec{\gG}_C =\phi_\theta(\vec{\gG}_A, \vec{\gG}_H, \vec{\gG}_L).
\end{align}

\emph{Symmetry preserved:} Apparently, the output CDRs $\vec{\gG}_C$ should be SE($3$)-equivariant with respect to the antigen:
\begin{align}
    g\cdot \vec{\gG}_C = \phi_\theta(g\cdot \vec{\gG}_A, g\cdot \vec{\gG}_H, g\cdot \vec{\gG}_L), \forall g \in \text{SE}(3)
\end{align}

\emph{Methods:} Antibody is of great significance in the field of therapeutics and biology, thus many works have dedicated to designing antibodies with desired binding specificity and affinity(\citep{kong2023conditional,kong2023end,jin2022iterative,shi2023protein,gao2022incorporating, verma2023abode, martinkus2023abdiffuser, luo2022antigenspecific}). RefineGNN~\citep{jin2022iterative} initiates the first attempt to design CDRs on the heavy chain only. Then MEAN~\citep{kong2023conditional} and DiffAb~\citep{luo2022antigenspecific} extend to the complete setting where the entire complex (\emph{i.e.} the antigen, the heavy chain and the light chain) without CDRs are given as contexts. Notably, MEAN~\citep{kong2023conditional} adopts GMN-like~\citep{huang2022equivariant} multi-channel architecture to encode the backbone atoms of the residues, and proposes an equivariant attention mechanism to capture interactions between different geometric components. Progressively, MEAN is upgraded to dyMEAN~\citep{kong2023end} which proposes a dynamic multi-channel encoder to capture the full-atom geometry of residues and tackles a more challenging setting where the entire structure and docking pose of the antibody needs to be generated instead of given as contexts. DiffAb~\citep{luo2022antigenspecific} proposes a diffusion generative model for antibody design. Similarly, AbDiffuser~\citep{martinkus2023abdiffuser} also adopts diffusion-based generative model, but steps forward to project each side chain into 4 pseudo-carbon atoms to capture the full-atom geometry and handles length change by placeholders in the sequence. ADesigner~\citep{tan2023cross} proposes a cross-gate MLP to facilitate the integration of sequences and structures. Unlike the aforementioned approaches, AbODE~\citep{verma2023abode} explores graph PDEs for antibody design. GeoAB~\citep{lin2024geoab} uses torsional prior knowledge with equivariant neural network focusing on bond lengths, bond angles and dihedrals. RADD~\citep{wu2024relation} introduce more node features, edge features, and edge relations to include more contextual and geometric information for designing the CDRs. Further,~\cite{gao2022incorporating} utilizes pretrained antibody language models to improve the quality of sequence-structure co-design, and tFold-Ab~\citep{wu2024fast} also employs a pretrained language model (\emph{i.e.} ESM-PPI), along with feature updating (\emph{i.e.} Evoformer-Single) and structure modules, to enable efficient and accurate prediction of antibody structures directly from sequence.

\subsubsection{Peptide Design}
Peptide, which consists of short sequences of amino acids, represents the intermediate modality between small molecules and proteins, and plays a critical role in various biological functions. Its unique position makes functional peptide design particularly appealing for both biological research and therapeutic applications~\citep{fosgerau2015peptide, lee2019comprehensive}.

\emph{Task definition:} Similar to antibody design, peptide design typically involves generating binding peptides for a given binding area on the target protein. Denoting the target as $\vec{\gG}_B$ and the peptide as $\vec{\gG}_P$, we can formalize the task as follows:
\begin{align}
    \vec{\gG}_P = \phi_\theta(\vec{\gG}_B).
\end{align}

\emph{Symmetry preserved:} Akin to antibody design, the output of the model requires to maintain invariance in the sequence distribution and equivariance in the structure distribution in terms of the $\mathrm{E}(3)$ group.

\emph{Datasets:} PepBDB~\citep{wen2019pepbdb} collects 13K protein-peptide complexes with peptides containing fewer than 50 residues from the Protein Data Bank~\citep{berman2000protein}. \cite{tsaban2022harnessing} curates a diverse and non-redundant dataset of 96 protein-peptide complexes, with peptides between 4 and 25 residues, which is referred to as the Long Non-Redundant (LNR) dataset. PepGLAD~\citep{kong2024full} further collects 6K non-redundant protein-peptide complexes, also featuring peptides between 4 and 25 residues, and partitions them based on the sequence identity of the receptors for training and validation, employing LNR as the test set.

\emph{Methods:} While conventional approaches rely on empirical energy functions to sample and optimize sequences and structures at the residue or  fragment level~\citep{bhardwaj2016accurate, cao2022design}, recent advances in geometric molecular design shed light on deep generative models. HelixGAN~\citep{xie2023helixgan} focuses on a sub-family of peptides with $\alpha$-helices. RFDiffusion~\citep{watson2023novo}, which is originally designed for protein generation, also explores supervised finetuning for target-specific peptide design. PepGLAD~\citep{kong2024full} takes a step further by tackling sequence-structure co-design with a geometric latent diffusion model.

\subsection{Tasks on Other Domains}
We briefly review the applications on other domains such as crystals and RNAs.

\subsubsection{Crystal Property Prediction}

In the realm of material science, the prediction of crystalline properties stands as a cornerstone for the innovation of new materials. Unlike molecules or proteins, which consist of a finite number of atoms, crystals are characterized by their periodic repetition throughout infinite 3D space. One of the main challenges lies in capturing this unique periodicity using geometric graph neural networks.

\emph{Task definition:} The infinite crystal structure is commonly simplified by its repeating unit, which is called a \emph{unit cell}, which is represented as $\vec{\gG}=(\vec{\mL}, \vec{\mX}, \mH)$, where $\vec{\mX}, \mH$ are coordinate matrix and feature matrix as defined before, and the additional matrix $\vec{\mL}=[\vec{\vl}_1, \vec{\vl}_2, \vec{\vl}_3]^\top\in\sR^{3\times 3}$ consists of three lattice vectors determining the periodicity of the crystal. The task is to predict the property $y\in\sR$ of the entire structure via the predictor $\phi_\theta$.
\begin{align}
    y=\phi_\theta(\vec{\mL}, \vec{\mX}, \mH).
\end{align}

\emph{Symmetry preserved:} The output of the predictor should be invariant with respect to several types of groups: 1) $\mathrm{E}(3)$-invariance of both the coordinates $\vec{\mX}$ and the lattice $\vec{\mL}$; 2) Periodic translation invariance of $\vec{\mX}$; 3) Cell choice invariance owing to periodicity, with details referred to~\citep{zeni2023mattergen}.

\emph{Datasets:} Materials Project (MP)~\citep{jain2013commentary} and JARVIS-DFT~\citep{choudhary2020joint} are two commonly-used datasets. In particular, MP is an open-access database containing more than 150K crystal structures with several properties collected by DFT calculation. JARVIS-DFT, part of the Joint Automated Repository for Various Integrated Simulations (JARVIS), is also calculated by DFT and provides more unique properties of materials like solar-efficiency and magnetic moment.

\emph{Methods:} To take the periodicity into consideration, CGCNN~\citep{xie2018crystal} proposes the multi-edge graph construction to model the interactions across the periodic boundaries. MEGNet~\citep{chen2019graph} additionally updates the global state
attributes during the message-passing procedure. ALIGNN~\citep{choudhary2021atomistic} composes two GNNs for both the atomic bond graph and its line graph to capture the interactions among atomic triplets. ECN~\citep{kaba2022equivariant} leverages space group symmetries into the GNNs for more powerful expressivity. Matformer~\citep{yan2022periodic} utilizes self-connecting edges to explicitly introduce the lattice matrix $\vec{\mL}$ into the transformer-based framework. To utilize the large amount of unlabeled data, Crystal Twins~\citep{magar2022crystal} applies two contrastive frameworks, Barlow Twins~\citep{zbontar2021barlow} and SimSiam~\citep{chen2021exploring} to pre-train the CGCNN models, and MMPT~\citep{yu2023crystal} proposes a mutex mask strategy to enforce the model to learn representations from two disjoint parts of the crystal.

\subsubsection{Crystal Generation}

Besides predicting the invariant properties of 3D crystals, the rapid progress of geometric graph neural networks has also paved the way to de novo material design, whose goal is to generate novel crystal structures beyond the existing databases.

\emph{Task definition:} Crystal generation methods commonly integrate geometric graph neural networks into deep generative frameworks, which aims to learn the distribution from a given dataset, allowing to generate new crystals through sampling from the learned distribution:
\begin{align}
    \vec{\mL}, \vec{\mX}, \mH\sim p_\theta(\vec{\mL}, \vec{\mX}, \mH).
\end{align}

\emph{Symmetry preserved:} Similar to the property prediction task, the learned distribution is also required to be invariant in terms of $\mathrm{E}(3)$ group and periodicity.

\emph{Datasets:} CDVAE~\citep{xie2021crystal} collects three datasets, named Perov-5~\citep{castelli2012new, castelli2012computational}, Carbon-24~\citep{carbon2020data} and MP-20~\citep{jain2013commentary} to evaluate the generative models on different crystal distributions. 

\emph{Methods:} CDVAE~\citep{xie2021crystal} incorporates a diffusion-based decoder into a VAE-based framework, by first predicting the lattice parameters from the latent space, and updating the atom types and coordinates according to the predicted lattice. SyMat~\citep{luo2023towards} refines this approach by generating atom types as permutation invariant sets and employing coordinate score-matching for the edges. DiffCSP~\citep{jiao2023crystal}, originally aiming at predicting crystal structures from given composition, also excels in generating structures from scratch. DiffCSP adopts the fractional coordinates $\mF=\vec{\mL}^{-1}\vec{\mX}$ instead of the Cartesian coordinates, and jointly generates the lattice matrix, atom types and coordinates via a diffusion-based framework. DiffCSP++~\citep{jiao2024space} extends DiffCSP with the conditions of lattice families and Wyckoff coordinates to maintain the space group constraints. Recently, MatterGen~\citep{zeni2023mattergen} further propels the joint diffusion method, and specializes the lattice diffusion process to be cubic-prior and rotation-fixed.

\subsubsection{RNA 3D Structure Ranking}
RNA, or ribonucleic acid, is a pivotal type of molecules that goes beyond its traditional role as a mere intermediary between DNA and protein synthesis. Its functionality heavily relies on its intricate three-dimensional structure, making the prediction and ranking of RNA's 3D conformation crucial. This structural complexity enables RNA to participate in gene regulation, cellular communication, and catalysis, underscoring its significance in fundamental life processes. As a result, RNA stands at the forefront of molecular biology and biotechnology research.

\emph{Task definition:} Here, we refer the ranking of 3D RNA structures to the task of identifying which structure most accurately reflecting the RNA's actual shape from a pool of imprecise ones. In other words, the score model $\phi_\theta$ is required to evaluate the root-mean-square deviation (RMSD) between each candidate 3D RNA structure represented by a geometric graph $\vec{\gG}$, and the ground truth:
\begin{align}
    s=\phi_\theta(\vec{\gG}).
\end{align}

\emph{Symmetry preserved:} This is obviously an invariant task because the RMSD value between the candidate structure and the ground truth remains impervious to any translations or rotations imposed on the candidate structure.

\emph{Methods:} ARES~\citep{raphael2021geometric} leverages e3nn~\citep{geiger2022e3nn} to model the 3D structure of RNA, ensuring equivariance and invariance during the update of atomic features. ARES then aggregates the features of all atoms to predict the RMSD value. In contrast, PaxNet~\citep{zhang2022physics} employs a two-layer multiplex graph to model the 3D structure of RNA. One layer captures local interactions, while the other focuses on non-local interactions. EquiRNA~\citep{li2025sizegeneralizable} introduces a hierarchical equivariant graph neural network with a size-insensitive K-nearest neighbor sampling strategy, aimed at solving the size generalization challenge through the reuse of nucleotide representations.

\emph{Datasets:} ARES~\citep{raphael2021geometric} uses a collection of 18K records from the FARFAR2-Classics dataset~\citep{das2007automated} as its training and validation sets. In addition, they have constructed two test sets: the first test set was selected from the FARFAR2-Puzzles dataset~\citep{das2007automated}; the second test set was curated based on certain criteria and built using the FARFAR2 rna denovo application. EquiRNA~\citep{li2025sizegeneralizable} introduces rRNAsolo, a new dataset for assessing size generalization in RNA structure evaluation. It covers a wider range of RNA sizes, more RNA types, and more recent RNA structures than existing datasets.

\section{Discussion and Future Prospect}
\label{sec:fp}

Whilst much progress has been made in this field, there are still a broad range of open research directions. We discuss several examples as follows.

\textbf{Geometric Graph Foundation Model.} 
Recent advancements in AI research, exemplified by the remarkable progress of models like the GPT series \citep{radford2018improving, radford2019language, brown2020language} and Gato~\citep{reed2022generalist}, have brought about substantial advantages by employing a unified foundational model across various tasks and domains. Foundation models diminish the necessity of manually crafting inductive biases for individual domains, amplifies the volume and variety of training data, and holds promise for further enhancement with increased data, computational resources, and model complexity. It is natural to mimic such success to geometric domain. However, it remains an interesting open question, especially considering the following design spaces. \textbf{1. Task space:} How to pretrain a large scale model that is generally beneficial to various downstream tasks? \textbf{2. Data space:} How to build a foundation model that can simultaneously extract rich information that spans across different types or scales of the geometric data? \textbf{3. Model space:} How to truly scale the model in terms of capacity and expressivity, such that more knowledge can be captured and stored in the model? Although some initial works (such as EPT~\citep{jiao2024equivariant}) manage to pretrain a unified model on small molecules and proteins, it still lacks a universal model that can tackle more kinds of input data and tasks.

\textbf{Effective Loop between Model Training and Real-World Experimental Verification.}
Unlike typical applications in vision and NLP, tasks in science usually require expensive labor, computational resources, and instruments to produce data, conduct verification, and record results. Existing research often adopts an open-loop style, where datasets are collected beforehand and proposed models are evaluated offline on these datasets. However, this approach presents two significant issues. Firstly, the constructed datasets are often small and insufficient for training geometric GNNs, especially for data-hungry foundational models equipped with large-scale parameters. Secondly, evaluating models solely on standalone datasets may fail to reflect feedback from the real world, resulting in less reliable evaluation of the model's true ability. These issues can be effectively addressed by training and testing geometric GNNs within a closed loop between model prediction and experimental verification. A notable example is provided by GNoME~\citep{merchant2023scaling}, which integrates an end-to-end pipeline consisting of graph network training, DFT computations, and autonomous laboratories for materials discovery and synthesis. It is expected that such a research paradigm will become increasingly important in future studies related to scientific applications.

\textbf{Integration with Large Language Models.} Large Language Models (LLMs) have been extensively shown to possess a wealth of knowledge, spanning various domains. Moreover, there has been a development of domain-specific Language Model Agents (LMAs) that exhibit high levels of expertise in specific areas~\citep{bran2023augmenting,liu2024agentbench}. Given that many of the tasks under discussion are intricately linked with the natural sciences, such as physics, biochemistry, and material science, which often require a deep understanding of domain-specific knowledge, it becomes compelling to enhance the existing knowledge base by integrating LLM agents into the training and evaluation pipeline of geometric Graph Neural Networks (GNNs). This integration holds promise for augmenting the capabilities of GNNs by leveraging the comprehensive knowledge representations offered by LLMs, thereby potentially improving the performance and robustness of these models in scientific applications. While there have been works leveraging LLMs for certain tasks such as molecule property prediction and drug design, they only operate on motifs~\citep{janakarajan2023large,liu2024conversational} or molecule graphs~\citep{zhang2023moleculegpt}. It still remains challenging to bridge them with geometric graph neural networks, enabling the pipeline to process 3D structural information and perform prediction and/or generation over 3D structures.

\textbf{Relaxation of Equivariance.} 
While equivariance is undeniably pivotal for bolstering data efficiency and promoting generalization across diverse datasets, it is noteworthy that rigidly adhering to equivariance principles can sometimes overly constrain the model, potentially compromising its performance. Thus, delving into methodologies that offer a degree of flexibility in relaxing equivariance constraints holds considerable significance. By exploring approaches that strike a balance between maintaining equivariance and accommodating adaptability, researchers can unlock avenues for enhancing the practical utility of models. Several pioneer studies~\citep{zheng2024relaxing,liu2024equivariant} try to relax the equivariance to a certain discrete point group and achieves a remarkable improvement on various dynamic physical systems, ranging from particle to vehicle dynamics. This exploration may not only enrich our understanding of model behavior but also pave the way for the development of more robust and versatile solutions with broader applicability.

\section{Conclusion}
\label{sec:conclusion}
In this survey, we conduct a systematic investigation of the progress in geometric Graph Neural Networks (GNNs), through the lens of data structures, models, and their applications. We specify geometric graph as the data structure, which generalizes the concept of graph in the presence of geometric information and permits the vital symmetry under certain transformations. We present geometric GNNs as the models, which consist of invariant GNNs, scalarization-based/high-degree steerable equivariant GNNs, and geometric graph transformers. We exhaustively discuss their applications through the taxonomy on the data and tasks, including both single instance and multi-instance tasks over domains in physics, biochemistry, and others like materials and RNAs.
We also discuss the challenges and the future potential directions of geometric GNNs.

\section{Acknowledgment}
\label{sec:acknowledgment}
This work was jointly supported by the following projects: 
The National Natural Science Foundation of China (No. 62376276, No. 62172422); 
Beijing Nova Program (No. 20230484278); 
the Fundamental Research Funds for the Central Universities, 
and the Research Funds of Renmin University of China (23XNKJ19);
Tencent AI Lab Rhino-Bird Focused Research Program.

\bibliographystyle{fcs}
\bibliography{Reference}

\begin{thebibliography}{100}

\bibitem{bronstein2021geometric}
Bronstein M~M, Bruna J, Cohen T, Veličković P.
\newblock {Geometric Deep Learning}: Grids, groups, graphs, geodesics, and gauges, 2021

\bibitem{schutt2017quantum}
Schütt K~T, Arbabzadah F, Chmiela S, Müller K~R, Tkatchenko A.
\newblock Quantum-chemical insights from deep tensor neural networks.
\newblock Nature Communications, 2017, 8(1)

\bibitem{Klicpera2020Directional}
Klicpera J, Groß J, Günnemann S.
\newblock Directional message passing for molecular graphs.
\newblock In: International Conference on Learning Representations.
\newblock 2020

\bibitem{klicpera2021gemnet}
Klicpera J, Becker F, G{\"u}nnemann S.
\newblock {GemNet}: Universal directional graph neural networks for molecules.
\newblock In: Annual Conference on Neural Information Processing Systems.
\newblock 2021

\bibitem{satorras2021en}
Satorras V~G, Hoogeboom E, Welling M.
\newblock E (n) equivariant graph neural networks.
\newblock In: International Conference on Machine Learning.
\newblock 2021

\bibitem{schutt2021equivariant}
Sch{\"u}tt K, Unke O, Gastegger M.
\newblock Equivariant message passing for the prediction of tensorial properties and molecular spectra.
\newblock In: International Conference on Machine Learning.
\newblock 2021,  9377--9388

\bibitem{thomas2018tensor}
Thomas N, Smidt T, Kearnes S, Yang L, Li~L, Kohlhoff K, Riley P.
\newblock Tensor field networks: Rotation-and translation-equivariant neural networks for 3d point clouds.
\newblock arXiv preprint arXiv:1802.08219, 2018

\bibitem{fuchs2020se}
Fuchs F, Worrall D, Fischer V, Welling M.
\newblock {SE(3)-Transformers}: 3d roto-translation equivariant attention networks.
\newblock In: Annual Conference on Neural Information Processing Systems.
\newblock 2020

\bibitem{brandstetter2022geometric}
Brandstetter J, Hesselink R, Pol v.~d E, Bekkers E~J, Welling M.
\newblock Geometric and physical quantities improve e(3) equivariant message passing.
\newblock In: International Conference on Learning Representations.
\newblock 2022

\bibitem{batzner20223}
Batzner S, Musaelian A, Sun L, Geiger M, Mailoa J~P, Kornbluth M, Molinari N, Smidt T~E, Kozinsky B.
\newblock E (3)-equivariant graph neural networks for data-efficient and accurate interatomic potentials.
\newblock Nature communications, 2022, 13(1): 2453

\bibitem{liao2023equiformer}
Liao Y~L, Smidt T.
\newblock Equiformer: Equivariant graph attention transformer for 3d atomistic graphs.
\newblock In: International Conference on Learning Representations.
\newblock 2023

\bibitem{minkyung2021rosettafold}
Baek M, DiMaio F, Anishchenko I, Dauparas J, others .
\newblock Accurate prediction of protein structures and interactions using a three-track neural network.
\newblock Science, 2021, 373(6557): 871--876

\bibitem{watson2023novo}
Watson J~L, Juergens D, Bennett N~R, others .
\newblock De novo design of protein structure and function with rfdiffusion.
\newblock Nature, 2023, 620(7976): 1089--1100

\bibitem{ingraham2023illuminating}
Ingraham J~B, Baranov M, Costello Z, Barber K~W, Wang W, Ismail A, Frappier V, Lord D~M, Ng-Thow-Hing C, Van~Vlack E~R, others .
\newblock Illuminating protein space with a programmable generative model.
\newblock Nature, 2023,  1--9

\bibitem{raphael2021geometric}
Townshend R~J~L, Eismann S, Watkins A~M, Rangan R, Karelina M, Das R, Dror R~O.
\newblock Geometric deep learning of rna structure.
\newblock Science, 2021, 373(6558): 1047--1051

\bibitem{corso2023diffdock}
Corso G, St{\"a}rk H, Jing B, Barzilay R, Jaakkola T~S.
\newblock {DiffDock}: Diffusion steps, twists, and turns for molecular docking.
\newblock In: International Conference on Learning Representations.
\newblock 2023

\bibitem{kong2023end}
Kong X, Huang W, Liu Y.
\newblock End-to-end full-atom antibody design.
\newblock In: International Conference on Machine Learning.
\newblock 23--29 Jul 2023,  17409--17429

\bibitem{gilmer2017neural}
Gilmer J, Schoenholz S~S, Riley P~F, Vinyals O, Dahl G~E.
\newblock Neural message passing for quantum chemistry.
\newblock In: International Conference on Machine Learning.
\newblock 2017

\bibitem{mcnutt2021gnina}
McNutt A~T, Francoeur P, Aggarwal R, Masuda T, Meli R, Ragoza M, Sunseri J, Koes D~R.
\newblock {GNINA 1.0}: molecular docking with deep learning.
\newblock Journal of cheminformatics, 2021, 13(1): 1--20

\bibitem{adolf2018rosettaantibodydesign}
Adolf-Bryfogle J, Kalyuzhniy O, Kubitz M, Weitzner B~D, Hu~X, Adachi Y, Schief W~R, Dunbrack~Jr R~L.
\newblock {RosettaAntibodyDesign (RAbD)}: A general framework for computational antibody design.
\newblock PLoS computational biology, 2018, 14(4): e1006112

\bibitem{ramakrishnan2014quantum}
Ramakrishnan R, Dral P~O, Rupp M, Lilienfeld v~O~A.
\newblock Quantum chemistry structures and properties of 134 kilo molecules.
\newblock Scientific Data, 2014, 1

\bibitem{liu2017forging}
Liu Z, Su~M, Han L, Liu J, Yang Q, Li~Y, Wang R.
\newblock Forging the basis for developing protein–ligand interaction scoring functions.
\newblock Accounts of Chemical Research, 2017, 50(2): 302--309

\bibitem{dunbar2014sabdab}
Dunbar J, Krawczyk K, Leem J, Baker T, Fuchs A, Georges G, Shi J, Deane C~M.
\newblock {SAbDab}: the structural antibody database.
\newblock Nucleic acids research, 2014, 42(D1): D1140--D1146

\bibitem{han2022geometrically}
Han J, Rong Y, Xu~T, Huang W.
\newblock Geometrically equivariant graph neural networks: A survey.
\newblock arXiv preprint arXiv:2202.07230, 2022

\bibitem{han2022learning}
Han J, Huang W, Ma~H, Li~J, Tenenbaum J~B, Gan C.
\newblock Learning physical dynamics with subequivariant graph neural networks.
\newblock In: Annual Conference on Neural Information Processing Systems.
\newblock 2022

\bibitem{sanchez2020learning}
Sanchez-Gonzalez A, Godwin J, Pfaff T, Ying R, Leskovec J, Battaglia P.
\newblock Learning to simulate complex physics with graph networks.
\newblock In: International Conference on Machine Learning.
\newblock 2020

\bibitem{kipf2018neural}
Kipf T, Fetaya E, Wang K~C, Welling M, Zemel R.
\newblock Neural relational inference for interacting systems.
\newblock In: International Conference on Machine Learning.
\newblock 2018

\bibitem{huang20223dlinker}
Huang Y, Peng X, Ma~J, Zhang M.
\newblock {3DLinker}: An e (3) equivariant variational autoencoder for molecular linker design.
\newblock In: International Conference on Machine Learning.
\newblock 2022,  9280--9294

\bibitem{guan2023d}
Guan J, Qian W~W, Peng X, Su~Y, Peng J, Ma~J.
\newblock 3d equivariant diffusion for target-aware molecule generation and affinity prediction.
\newblock In: International Conference on Learning Representations.
\newblock 2023

\bibitem{jing2022torsional}
Jing B, Corso G, Chang J, Barzilay R, Jaakkola T~S.
\newblock Torsional diffusion for molecular conformer generation.
\newblock In: N e u r IPS.
\newblock 2022

\bibitem{wu2023equivariant}
Liming W, Zhichao H, Jirui Y, Yu~R, Huang W.
\newblock Equivariant spatio-temporal attentive graph networks to simulate physical dynamics.
\newblock In: Annual Conference on Neural Information Processing Systems.
\newblock 2023

\bibitem{kong2023conditional}
Kong X, Huang W, Liu Y.
\newblock Conditional antibody design as 3d equivariant graph translation.
\newblock In: International Conference on Learning Representations.
\newblock 2023

\bibitem{senior2020improved}
Senior A~W, Evans R, Jumper J, Kirkpatrick J, Sifre L, Green T, Qin C, {\v{Z}}{\'\i}dek A, Nelson A~W, Bridgland A, others .
\newblock Improved protein structure prediction using potentials from deep learning.
\newblock Nature, 2020, 577(7792): 706--710

\bibitem{ocp_dataset}
Chanussot* L, Das* A, Goyal* S, others .
\newblock Open catalyst 2020 (oc20) dataset and community challenges.
\newblock ACS Catalysis, 2021

\bibitem{kong2024full}
Kong X, Huang W, Liu Y.
\newblock Full-atom peptide design with geometric latent diffusion.
\newblock arXiv preprint arXiv:2402.13555, 2024

\bibitem{duval2023hitchhiker}
Duval A, Mathis S~V, Joshi C~K, Schmidt V, Miret S, Malliaros F~D, Cohen T, Lio P, Bengio Y, Bronstein M.
\newblock A hitchhiker's guide to geometric gnns for 3d atomic systems.
\newblock arXiv preprint arXiv:2312.07511, 2023

\bibitem{xia2023systematic}
Xia J, Zhu Y, Du~Y, Liu Y, Li~S.
\newblock A systematic survey of chemical pre-trained models.
\newblock In: International Joint Conference on Artificial Intelligence.
\newblock 2023

\bibitem{guo2023graph}
Guo Z, Guo K, Nan B, Tian Y, Iyer R~G, Ma~Y, Wiest O, Zhang X, Wang W, Zhang C, Chawla N~V.
\newblock Graph-based molecular representation learning.
\newblock In: {International Joint Conference on Artificial Intelligence}.
\newblock 8 2023,  6638--6646.
\newblock Survey Track

\bibitem{atz2021geometric}
Atz K, Grisoni F, Schneider G.
\newblock Geometric deep learning on molecular representations.
\newblock Nature Machine Intelligence, 2021, 3(12): 1023--1032

\bibitem{zhang2023artificial}
Zhang X, Wang L, Helwig J, Luo Y, Fu~C, Xie Y, Liu M, Lin Y, Xu~Z, Yan K, others .
\newblock Artificial intelligence for science in quantum, atomistic, and continuum systems.
\newblock arXiv preprint arXiv:2307.08423, 2023

\bibitem{esteves2020theoretical}
Esteves C.
\newblock Theoretical aspects of group equivariant neural networks, 2020

\bibitem{cederberg2004course}
Cederberg J.
\newblock A course in modern geometries.
\newblock Springer Science \& Business Media, 2004

\bibitem{wu2020comprehensive}
Wu~Z, Pan S, Chen F, Long G, Zhang C, Philip S~Y.
\newblock A comprehensive survey on graph neural networks.
\newblock TNNLS, 2020, 32(1): 4--24

\bibitem{yuan2024index}
Yuan Z, Wei Z, Lv~F, Wen J~R.
\newblock Index-free triangle-based graph local clustering.
\newblock Frontiers of Computer Science, 2024, 18(3): 183404

\bibitem{C7SC02664A}
Wu~Z, Ramsundar B, Feinberg E, Gomes J, Geniesse C, Pappu A~S, Leswing K, Pande V.
\newblock {MoleculeNet}: a benchmark for molecular machine learning.
\newblock Chem. Sci., 2018, 9: 513--530

\bibitem{villar2021scalars}
Villar S, Hogg D~W, Storey-Fisher K, Yao W, Blum-Smith B.
\newblock Scalars are universal: Equivariant machine learning, structured like classical physics.
\newblock In: Annual Conference on Neural Information Processing Systems.
\newblock 2021

\bibitem{schutt2018schnet}
Sch{\"u}tt K~T, Sauceda H~E, Kindermans P~J, Tkatchenko A, M{\"u}ller K~R.
\newblock Schnet--a deep learning architecture for molecules and materials.
\newblock The Journal of Chemical Physics, 2018, 148(24): 241722

\bibitem{baek2023efficient}
Baek M, Anishchenko I, Humphreys I, Cong Q, Baker D, DiMaio F.
\newblock Efficient and accurate prediction of protein structure using rosettafold2.
\newblock bioRxiv, 2023,  2023--05

\bibitem{luo2023towards}
Luo Y, Liu C, Ji~S.
\newblock Towards symmetry-aware generation of periodic materials.
\newblock Annual Conference on Neural Information Processing Systems, 2024, 36

\bibitem{jiao2023crystal}
Jiao R, Huang W, Lin P, Han J, Chen P, Lu~Y, Liu Y.
\newblock Crystal structure prediction by joint equivariant diffusion.
\newblock Annual Conference on Neural Information Processing Systems, 2024, 36

\bibitem{huang2022equivariant}
Huang W, Han J, Rong Y, Xu~T, Sun F, Huang J.
\newblock Equivariant graph mechanics networks with constraints.
\newblock In: International Conference on Learning Representations.
\newblock 2022

\bibitem{gasteiger2020fast}
Gasteiger J, Giri S, Margraf J~T, G{\"u}nnemann S.
\newblock Fast and uncertainty-aware directional message passing for non-equilibrium molecules.
\newblock arXiv:2011.14115, 2020

\bibitem{zhu2023fastdimenet++}
Zhu F, Futrega M, Bao H, others .
\newblock {FastDimeNet++}: Training dimenet++ in 22 minutes.
\newblock In: Proceedings of the 52nd International Conference on Parallel Processing.
\newblock 2023,  274--284

\bibitem{finzi2020generalizing}
Finzi M, Stanton S, Izmailov P, Wilson A~G.
\newblock Generalizing convolutional neural networks for equivariance to lie groups on arbitrary continuous data.
\newblock In: International Conference on Machine Learning.
\newblock 2020

\bibitem{liu2022spherical}
Liu Y, Wang L, Liu M, Lin Y, Zhang X, Oztekin B, Ji~S.
\newblock Spherical message passing for 3d molecular graphs.
\newblock In: International Conference on Learning Representations.
\newblock 2022

\bibitem{wang2022comenet}
Wang L, Liu Y, Lin Y, Liu H, Ji~S.
\newblock {ComENet}: Towards complete and efficient message passing for 3d molecular graphs.
\newblock Annual Conference on Neural Information Processing Systems, 2022, 35: 650--664

\bibitem{li2024distance}
Li~Z, Wang X, Huang Y, Zhang M.
\newblock Is distance matrix enough for geometric deep learning?
\newblock Annual Conference on Neural Information Processing Systems, 2024, 36

\bibitem{li2024completeness}
Li~Z, Wang X, Kang S, Zhang M.
\newblock On the completeness of invariant geometric deep learning models.
\newblock arXiv preprint arXiv:2402.04836, 2024

\bibitem{yue2024a}
Yue A, Luo D, Xu~H.
\newblock A plug-and-play quaternion message-passing module for molecular conformation representation.
\newblock Proceedings of the AAAI Conference on Artificial Intelligence, 2024

\bibitem{du2022se}
Du~W, Zhang H, Du~Y, Meng Q, Chen W, Zheng N, Shao B, Liu T~Y.
\newblock {SE}(3) equivariant graph neural networks with complete local frames.
\newblock In: Proceedings of the 39th International Conference on Machine Learning.
\newblock 17--23 Jul 2022,  5583--5608

\bibitem{kofinas2021rototranslated}
Kofinas M, Nagaraja N~S, Gavves E.
\newblock Roto-translated local coordinate frames for interacting dynamical systems.
\newblock In: Annual Conference on Neural Information Processing Systems.
\newblock 2021

\bibitem{kofinas2023latent}
Kofinas M, Bekkers E~J, Nagaraja N~S, Gavves E.
\newblock Latent field discovery in interacting dynamical systems with neural fields.
\newblock In: Annual Conference on Neural Information Processing Systems.
\newblock 2023

\bibitem{kohler2019equivariant}
K{\"o}hler J, Klein L, No{\'e} F.
\newblock {Equivariant flows}: sampling configurations for multi-body systems with symmetric energies.
\newblock arXiv preprint arXiv:1910.00753, 2019

\bibitem{jing2021learning}
Jing B, Eismann S, Suriana P, Townshend R~J~L, Dror R.
\newblock Learning from protein structure with geometric vector perceptrons.
\newblock In: International Conference on Learning Representations.
\newblock 2021

\bibitem{han2022equivariant}
Han J, Huang W, Xu~T, Rong Y.
\newblock Equivariant graph hierarchy-based neural networks.
\newblock In: Annual Conference on Neural Information Processing Systems.
\newblock 2022

\bibitem{zhang2024improving}
Zhang Y, Cen J, Han J, Zhang Z, ZHOU J, Huang W.
\newblock Improving equivariant graph neural networks on large geometric graphs via virtual nodes learning.
\newblock In: International Conference on Machine Learning.
\newblock 2024

\bibitem{puny2021frame}
Puny O, Atzmon M, Smith E~J, Misra I, Grover A, Ben-Hamu H, Lipman Y.
\newblock Frame averaging for invariant and equivariant network design.
\newblock In: International Conference on Learning Representations.
\newblock 2021

\bibitem{duval2023faenet}
Duval A~A, Schmidt V, Hern{\'a}ndez-Garc{\i}a A, Miret S, Malliaros F~D, Bengio Y, Rolnick D.
\newblock Faenet: Frame averaging equivariant gnn for materials modeling.
\newblock In: International Conference on Machine Learning.
\newblock 2023,  9013--9033

\bibitem{du2024new}
Du~Y, Wang L, Feng D, Wang G, Ji~S, Gomes C~P, Ma~Z~M, others .
\newblock A new perspective on building efficient and expressive 3d equivariant graph neural networks.
\newblock Annual Conference on Neural Information Processing Systems, 2023, 36

\bibitem{aykent2023savenet}
Aykent S, Xia T.
\newblock Savenet: a scalable vector network for enhanced molecular representation learning.
\newblock Advances in Neural Information Processing Systems, 2023, 36: 42932--42949

\bibitem{wang2024enhancing}
Wang Y, Wang T, Li~S, He~X, Li~M, Wang Z, Zheng N, Shao B, Liu T~Y.
\newblock Enhancing geometric representations for molecules with equivariant vector-scalar interactive message passing.
\newblock Nature Communications, 2024, 15(1): 313

\bibitem{wang2023efficiently}
Wang Z, Liu G, Zhou Y, Wang T, Shao B.
\newblock Efficiently incorporating quintuple interactions into geometric deep learning force fields.
\newblock In: Thirty-seventh Conference on Neural Information Processing Systems.
\newblock 2023

\bibitem{cen2024high}
Cen J, Li~A, Lin N, Ren Y, Wang Z, Huang W.
\newblock Are high-degree representations really unnecessary in equivariant graph neural networks?
\newblock In: Annual Conference on Neural Information Processing Systems.
\newblock 2024

\bibitem{battiloro2025etnn}
Battiloro C, Karaismailo{\u{g}}lu E, Tec M, Dasoulas G, Audirac M, Dominici F.
\newblock E(n) equivariant topological neural networks.
\newblock In: The Thirteenth International Conference on Learning Representations.
\newblock 2025

\bibitem{li2025large}
Li~Z, Cen J, Su~B, Huang W, Xu~T, Rong Y, Zhao D.
\newblock Large language-geometry model: When llm meets equivariance.
\newblock arXiv preprint arXiv:2502.11149, 2025

\bibitem{anderson2019cormorant}
Anderson B, Hy~T~S, Kondor R.
\newblock Cormorant: Covariant molecular neural networks.
\newblock In: Annual Conference on Neural Information Processing Systems.
\newblock 2019

\bibitem{musaelian2023learning}
Musaelian A, Batzner S, Johansson A, Sun L, Owen C~J, Kornbluth M, Kozinsky B.
\newblock Learning local equivariant representations for large-scale atomistic dynamics.
\newblock Nature Communications, 2023, 14(1): 579

\bibitem{zitnick2022spherical}
Zitnick L, Das A, Kolluru A, Lan J, Shuaibi M, Sriram A, Ulissi Z, Wood B.
\newblock Spherical channels for modeling atomic interactions.
\newblock Annual Conference on Neural Information Processing Systems, 2022, 35: 8054--8067

\bibitem{passaro2023reducing}
Passaro S, Zitnick C~L.
\newblock Reducing {SO}(3) convolutions to {SO}(2) for efficient equivariant {GNN}s.
\newblock In: Proceedings of the 40th International Conference on Machine Learning.
\newblock 2023,  27420--27438

\bibitem{batatia2022mace}
Batatia I, Kovacs D~P, Simm G, Ortner C, Cs{\'a}nyi G.
\newblock {MACE}: Higher order equivariant message passing neural networks for fast and accurate force fields.
\newblock Annual Conference on Neural Information Processing Systems, 2022, 35: 11423--11436

\bibitem{ying2021transformers}
Ying C, Cai T, Luo S, Zheng S, Ke~G, He~D, Shen Y, Liu T~Y.
\newblock Do transformers really perform badly for graph representation?
\newblock Annual Conference on Neural Information Processing Systems, 2021

\bibitem{shi2022benchmarking}
Shi Y, Zheng S, Ke~G, Shen Y, You J, He~J, Luo S, Liu C, He~D, Liu T~Y.
\newblock Benchmarking graphormer on large-scale molecular modeling datasets.
\newblock arXiv preprint arXiv:2203.04810, 2022

\bibitem{tholke2022equivariant}
Th{\"o}lke P, Fabritiis G~D.
\newblock Equivariant transformers for neural network based molecular potentials.
\newblock In: International Conference on Learning Representations.
\newblock 2022

\bibitem{hutchinson2021lietransformer}
Hutchinson M~J, Le~Lan C, Zaidi S, Dupont E, Teh Y~W, Kim H.
\newblock Lietransformer: equivariant self-attention for lie groups.
\newblock In: International Conference on Machine Learning.
\newblock 2021

\bibitem{hsu2022learning}
Hsu C, Verkuil R, Liu J, Lin Z, Hie B, Sercu T, Lerer A, Rives A.
\newblock Learning inverse folding from millions of predicted structures.
\newblock In: International Conference on Machine Learning.
\newblock 2022,  8946--8970

\bibitem{liao2023equiformerv2}
Liao Y~L, Wood B~M, Das A, Smidt T.
\newblock {EquiformerV2}: Improved equivariant transformer for scaling to higher-degree representations.
\newblock In: International Conference on Learning Representations.
\newblock 2024

\bibitem{wang2024geometric}
Wang Y, Li~S, Wang T, Shao B, Zheng N, Liu T~Y.
\newblock Geometric transformer with interatomic positional encoding.
\newblock Annual Conference on Neural Information Processing Systems, 2024, 36

\bibitem{frank2024euclidean}
Frank J~T, Unke O~T, M{\"u}ller K~R, Chmiela S.
\newblock A euclidean transformer for fast and stable machine learned force fields.
\newblock Nature Communications, 2024, 15(1): 6539

\bibitem{aykent2025rethinking}
Aykent S, Xia T.
\newblock Gotennet: Rethinking efficient 3d equivariant graph neural networks.
\newblock In: The Thirteenth International Conference on Learning Representations.
\newblock 2025

\bibitem{jiao2024equivariant}
Jiao R, Kong X, Yu~Z, Huang W, Liu Y.
\newblock Equivariant pretrained transformer for unified geometric learning on multi-domain 3d molecules.
\newblock arXiv preprint arXiv:2402.12714, 2024

\bibitem{10.1093/bioinformatics/btac039}
Ma~H, Bian Y, Rong Y, Huang W, Xu~T, Xie W, Ye~G, Huang J.
\newblock {Cross-dependent graph neural networks for molecular property prediction}.
\newblock Bioinformatics, 2022, 38(7): 2003--2009

\bibitem{zhang2021nested}
Zhang M, Li~P.
\newblock Nested graph neural networks.
\newblock Advances in Neural Information Processing Systems, 2021, 34: 15734--15747

\bibitem{qin2022fast}
Qin S, Zhang X, Xu~H, Xu~Y.
\newblock Fast quaternion product units for learning disentangled representations in so(3).
\newblock TPAMI, 2022, 45(4): 4504--4520

\bibitem{zhu2018quaternion}
Zhu X, Xu~Y, Xu~H, Chen C.
\newblock Quaternion convolutional neural networks.
\newblock In: Proceedings of the European conference on computer vision (ECCV).
\newblock 2018,  631--647

\bibitem{zhang2020quaternion}
Zhang X, Qin S, Xu~Y, Xu~H.
\newblock Quaternion product units for deep learning on 3d rotation groups.
\newblock In: IEEE/CVF Conference on Computer Vision and Pattern Recognition (CVPR).
\newblock 2020,  7304--7313

\bibitem{joshi2023expressive}
Joshi C~K, Bodnar C, Mathis S~V, Cohen T, Liò P.
\newblock On the expressive power of geometric graph neural networks.
\newblock In: International Conference on Machine Learning.
\newblock 2023

\bibitem{gilmore2008lie}
Gilmore R.
\newblock Lie groups, physics, and geometry: an introduction for physicists, engineers and chemists.
\newblock Cambridge University Press, 2008

\bibitem{muller2006spherical}
M{\"u}ller C.
\newblock Spherical harmonics. volume~17.
\newblock Springer, 2006

\bibitem{griffiths2018introduction}
Griffiths D~J, Schroeter D~F.
\newblock Introduction to quantum mechanics.
\newblock Cambridge university press, 2018

\bibitem{weiler20183d}
Weiler M, Geiger M, Welling M, Boomsma W, Cohen T~S.
\newblock {3D Steerable CNNs}: Learning rotationally equivariant features in volumetric data.
\newblock Annual Conference on Neural Information Processing Systems, 2018, 31

\bibitem{ramachandran2017searching}
Ramachandran P, Zoph B, Le~Q~V.
\newblock Searching for activation functions, 2017

\bibitem{drautz2019atomic}
Drautz R.
\newblock Atomic cluster expansion for accurate and transferable interatomic potentials.
\newblock Physical Review B, 2019, 99(1): 014104

\bibitem{dusson2022atomic}
Dusson G, Bachmayr M, Cs{\'a}nyi G, Drautz R, Etter S, Oord v.~d C, Ortner C.
\newblock Atomic cluster expansion: Completeness, efficiency and stability.
\newblock Journal of Computational Physics, 2022, 454: 110946

\bibitem{bochkarev2022efficient}
Bochkarev A, Lysogorskiy Y, Menon S, Qamar M, Mrovec M, Drautz R.
\newblock Efficient parametrization of the atomic cluster expansion.
\newblock Physical Review Materials, 2022, 6(1): 013804

\bibitem{vaswani2017attention}
Vaswani A, Shazeer N, Parmar N, Uszkoreit J, Jones L, Gomez A~N, Kaiser {\L}, Polosukhin I.
\newblock Attention is all you need.
\newblock Annual Conference on Neural Information Processing Systems, 2017, 30

\bibitem{yuan2025survey}
Yuan C, Zhao K, Kuruoglu E~E, Wang L, Xu~T, Huang W, Zhao D, Cheng H, Rong Y.
\newblock A survey of graph transformers: Architectures, theories and applications.
\newblock arXiv preprint arXiv:2502.16533, 2025

\bibitem{hu2021ogb}
Hu~W, Fey M, Ren H, Nakata M, Dong Y, Leskovec J.
\newblock {OGB-LSC}: A large-scale challenge for machine learning on graphs.
\newblock In: Thirty-fifth Conference on Neural Information Processing Systems Datasets and Benchmarks Track (Round 2).
\newblock 2021

\bibitem{shuaibi2021rotation}
Shuaibi M, Kolluru A, Das A, Grover A, Sriram A, Ulissi Z, Zitnick C~L.
\newblock Rotation invariant graph neural networks using spin convolutions.
\newblock arXiv preprint arXiv:2106.09575, 2021

\bibitem{dym2020universality}
Dym N, Maron H.
\newblock On the universality of rotation equivariant point cloud networks.
\newblock In: International Conference on Learning Representations.
\newblock 2020

\bibitem{weisfeiler1968reduction}
Weisfeiler B, Leman A.
\newblock The reduction of a graph to canonical form and the algebra which appears therein.
\newblock nti, Series, 1968, 2(9): 12--16

\bibitem{lawrence2025improving}
Lawrence H, Portilheiro V, Zhang Y, Kaba S~O.
\newblock Improving equivariant networks with probabilistic symmetry breaking.
\newblock In: The Thirteenth International Conference on Learning Representations.
\newblock 2025

\bibitem{battaglia2016interaction}
Battaglia P, Pascanu R, Lai M, Jimenez~Rezende D, others .
\newblock Interaction networks for learning about objects, relations and physics.
\newblock Annual Conference on Neural Information Processing Systems, 2016, 29

\bibitem{sanchez2019hamiltonian}
Sanchez-Gonzalez A, Bapst V, Cranmer K, Battaglia P.
\newblock Hamiltonian graph networks with ode integrators.
\newblock arXiv preprint arXiv:1909.12790, 2019

\bibitem{guo2023newton}
Guo L, Wang W, Chen Z, Zhang N, Sun Z, Lai Y, Zhang Q, Chen H.
\newblock Newton--cotes graph neural networks: On the time evolution of dynamic systems.
\newblock Annual Conference on Neural Information Processing Systems, 2024, 36

\bibitem{allen2023graph}
Allen K~R, Guevara T~L, Rubanova Y, Stachenfeld K, Sanchez-Gonzalez A, Battaglia P, Pfaff T.
\newblock Graph network simulators can learn discontinuous, rigid contact dynamics.
\newblock In: Conference on Robot Learning.
\newblock 2023,  1157--1167

\bibitem{rubanova2022constraint}
Rubanova Y, Sanchez-Gonzalez A, Pfaff T, Battaglia P.
\newblock Constraint-based graph network simulator.
\newblock In: International Conference on Machine Learning.
\newblock 2022,  18844--18870

\bibitem{wu2022learning}
Wu~T, Wang Q, Zhang Y, Ying R, Cao K, Sosic R, Jalali R, Hamam H, Maucec M, Leskovec J.
\newblock Learning large-scale subsurface simulations with a hybrid graph network simulator.
\newblock In: Proceedings of the 28th ACM SIGKDD Conference on Knowledge Discovery and Data Mining.
\newblock 2022,  4184--4194

\bibitem{li2018learning}
Li~Y, Wu~J, Tedrake R, Tenenbaum J~B, Torralba A.
\newblock Learning particle dynamics for manipulating rigid bodies, deformable objects, and fluids.
\newblock In: International Conference on Learning Representations.
\newblock 2018

\bibitem{mrowca2018flexible}
Mrowca D, Zhuang C, Wang E, Haber N, Fei-Fei L, Tenenbaum J~B, Yamins D~L~K.
\newblock Flexible neural representation for physics prediction.
\newblock In: Annual Conference on Neural Information Processing Systems.
\newblock 2018

\bibitem{allen2023learning}
Allen K~R, Rubanova Y, Lopez-Guevara T, Whitney W~F, Sanchez-Gonzalez A, Battaglia P, Pfaff T.
\newblock Learning rigid dynamics with face interaction graph networks.
\newblock In: International Conference on Learning Representations.
\newblock 2023

\bibitem{xu2023eqmotion}
Xu~C, Tan R~T, Tan Y, Chen S, Wang Y~G, Wang X, Wang Y.
\newblock {EqMotion}: Equivariant multi-agent motion prediction with invariant interaction reasoning.
\newblock In: IEEE/CVF Conference on Computer Vision and Pattern Recognition (CVPR).
\newblock 2023,  1410--1420

\bibitem{liu2023physicsinspired}
Liu Y, Cheng J, Zhao H, Xu~T, Zhao P, Tsung F, Li~J, Rong Y.
\newblock Improving generalization in equivariant graph neural networks with physical inductive biases.
\newblock In: International Conference on Learning Representations.
\newblock 2024

\bibitem{coors2018spherenet}
Coors B, Condurache A~P, Geiger A.
\newblock {SphereNet}: Learning spherical representations for detection and classification in omnidirectional images.
\newblock In: Proceedings of the European conference on computer vision (ECCV).
\newblock 2018,  518--533

\bibitem{wangxiyuan2022graph}
Wang X, Zhang M.
\newblock Graph neural network with local frame for molecular potential energy surface.
\newblock In: LOG.
\newblock 2022,  19--1

\bibitem{luo2024enabling}
Luo S, Chen T, Krishnapriyan A~S.
\newblock Enabling efficient equivariant operations in the fourier basis via gaunt tensor products.
\newblock In: The Twelfth International Conference on Learning Representations.
\newblock 2024

\bibitem{kohler2020equivariant}
K{\"o}hler J, Klein L, Noe F.
\newblock {Equivariant Flows}: Exact likelihood generative learning for symmetric densities.
\newblock In: International Conference on Machine Learning.
\newblock 2020

\bibitem{xu2024equivariant}
Xu~M, Han J, Lou A, Kossaifi J, Ramanathan A, Azizzadenesheli K, Leskovec J, Ermon S, Anandkumar A.
\newblock Equivariant graph neural operator for modeling 3d dynamics.
\newblock arXiv preprint arXiv:2401.11037, 2024

\bibitem{schreiner2023implicit}
Schreiner M, Winther O, Olsson S.
\newblock Implicit transfer operator learning: Multiple time-resolution models for molecular dynamics.
\newblock Annual Conference on Neural Information Processing Systems, 2024, 36

\bibitem{midgley2024se}
Midgley L, Stimper V, Antor{\'a}n J, Mathieu E, Sch{\"o}lkopf B, Hern{\'a}ndez-Lobato J~M.
\newblock Se (3) equivariant augmented coupling flows.
\newblock Annual Conference on Neural Information Processing Systems, 2024, 36

\bibitem{han2024geometric}
Han J, Xu~M, Lou A, Ye~H, Ermon S.
\newblock Geometric trajectory diffusion models.
\newblock In: The Thirty-eighth Annual Conference on Neural Information Processing Systems.
\newblock 2024

\bibitem{raja2024stability}
Raja S, Amin I, Pedregosa F~a.
\newblock Stability-aware training of neural network interatomic potentials with differentiable boltzmann estimators.
\newblock arXiv preprint arXiv:2402.13984, 2024

\bibitem{amin2025towards}
Amin I, Raja , Krishnapriyan A~S.
\newblock Towards fast, specialized machine learning force fields: Distilling foundation models via energy hessians.
\newblock In: The Thirteenth International Conference on Learning Representations.
\newblock 2025

\bibitem{xu2022geodiff}
Xu~M, Yu~L, Song Y, Shi C, Ermon S, Tang J.
\newblock {GeoDiff}: A geometric diffusion model for molecular conformation generation.
\newblock In: International Conference on Learning Representations.
\newblock 2022

\bibitem{xu2023geometric}
Xu~M, Powers A~S, Dror R~O, Ermon S, Leskovec J.
\newblock Geometric latent diffusion models for 3d molecule generation.
\newblock In: International Conference on Machine Learning.
\newblock 2023,  38592--38610

\bibitem{xu2021an}
Xu~M, Wang W, Luo S, Shi C, Bengio Y, Gomez-Bombarelli R, Tang J.
\newblock An end-to-end framework for molecular conformation generation via bilevel programming.
\newblock In: Proceedings of the 38th International Conference on Machine Learning.
\newblock 18--24 Jul 2021,  11537--11547

\bibitem{shi2021learning}
Shi C, Luo S, Xu~M, Tang J.
\newblock Learning gradient fields for molecular conformation generation.
\newblock In: International Conference on Machine Learning.
\newblock 2021

\bibitem{gebauer2019symmetry}
Gebauer N, Gastegger M, Sch{\"u}tt K.
\newblock Symmetry-adapted generation of 3d point sets for the targeted discovery of molecules.
\newblock Annual Conference on Neural Information Processing Systems, 2019, 32

\bibitem{gebauer2022inverse}
Gebauer N~W, Gastegger M, Hessmann S~S, M{\"u}ller K~R, Sch{\"u}tt K~T.
\newblock Inverse design of 3d molecular structures with conditional generative neural networks.
\newblock Nature communications, 2022, 13(1): 973

\bibitem{huang2023mdm}
Huang L, Zhang H, Xu~T, Wong K~C.
\newblock Mdm: Molecular diffusion model for 3d molecule generation.
\newblock In: Proceedings of the AAAI Conference on Artificial Intelligence.
\newblock 2023,  5105--5112

\bibitem{peng2023moldiff}
Peng X, Guan J, Liu Q, Ma~J.
\newblock {M}ol{D}iff: Addressing the atom-bond inconsistency problem in 3{D} molecule diffusion generation.
\newblock In: International Conference on Machine Learning.
\newblock 23--29 Jul 2023,  27611--27629

\bibitem{luo2021predicting}
Luo S, Shi C, Xu~M, Tang J.
\newblock Predicting molecular conformation via dynamic graph score matching.
\newblock In: Annual Conference on Neural Information Processing Systems.
\newblock 2021

\bibitem{satorras2021enf}
Satorras V~G, Hoogeboom E, Fuchs F~B, Posner I, Welling M.
\newblock E(n) equivariant normalizing flows.
\newblock In: Annual Conference on Neural Information Processing Systems.
\newblock 2021

\bibitem{hoogeboom2022equivariant}
Hoogeboom E, Satorras V~G, Vignac C, Welling M.
\newblock Equivariant diffusion for molecule generation in 3d.
\newblock In: International Conference on Machine Learning.
\newblock 2022,  8867--8887

\bibitem{ganea2021geomol}
Ganea O~E, Pattanaik L, Coley C~W, Barzilay R, Jensen K, Green W, Jaakkola T~S.
\newblock {GeoMol}: Torsional geometric generation of molecular 3d conformer ensembles.
\newblock In: Annual Conference on Neural Information Processing Systems.
\newblock 2021

\bibitem{wang2023mperformer}
Wang F, Xu~H, Chen X, Lu~S, Deng Y, Huang W.
\newblock {MPerformer}: An se (3) transformer-based molecular perceptron.
\newblock In: Proceedings of the 32nd ACM International Conference on Information and Knowledge Management.
\newblock 2023,  2512--2522

\bibitem{bao2023equivariant}
Bao F, Zhao M, Hao Z, Li~P, Li~C, Zhu J.
\newblock Equivariant energy-guided {SDE} for inverse molecular design.
\newblock In: International Conference on Learning Representations.
\newblock 2023

\bibitem{zhu2022direct}
Zhu J, Xia Y, Liu C, Wu~L, Xie S, Wang Y, Wang T, Qin T, Zhou W, Li~H, Liu H, Liu T~Y.
\newblock Direct molecular conformation generation.
\newblock Transactions on Machine Learning Research, 2022

\bibitem{qiang2023coarse}
Qiang B, Song Y, Xu~M, Gong J, Gao B, Zhou H, Ma~W~Y, Lan Y.
\newblock Coarse-to-fine: a hierarchical diffusion model for molecule generation in 3d.
\newblock In: International Conference on Machine Learning.
\newblock 2023,  28277--28299

\bibitem{song2024equivariant}
Song Y, Gong J, Xu~M, Cao Z, Lan Y, Ermon S, Zhou H, Ma~W~Y.
\newblock Equivariant flow matching with hybrid probability transport for 3d molecule generation.
\newblock Annual Conference on Neural Information Processing Systems, 2024, 36

\bibitem{reidenbach2024coarsenconf}
Reidenbach D, Krishnapriyan A~S.
\newblock Coarsenconf: Equivariant coarsening with aggregated attention for molecular conformer generation.
\newblock Journal of Chemical Information and Modeling, 2024, 65(1): 22--30

\bibitem{song2024unified}
Song Y, Gong J, Zhou H, Zheng M, Liu J, Ma~W~Y.
\newblock Unified generative modeling of 3d molecules with bayesian flow networks.
\newblock In: The Twelfth International Conference on Learning Representations.
\newblock 2024

\bibitem{qu2024molcraft}
Qu~Y, Qiu K, Song Y, Gong J, Han J, Zheng M, Zhou H, Ma~W~Y.
\newblock Mol{CRAFT}: Structure-based drug design in continuous parameter space.
\newblock In: Forty-first International Conference on Machine Learning.
\newblock 2024

\bibitem{jiao2022energy}
Jiao R, Han J, Huang W, Rong Y, Liu Y.
\newblock Energy-motivated equivariant pretraining for 3d molecular graphs.
\newblock In: Proceedings of the AAAI Conference on Artificial Intelligence.
\newblock 2023,  8096--8104

\bibitem{liu2023molecular}
Liu S, Guo H, Tang J.
\newblock Molecular geometry pretraining with {SE}(3)-invariant denoising distance matching.
\newblock In: International Conference on Learning Representations.
\newblock 2023

\bibitem{liu2022pretraining}
Liu S, Wang H, Liu W, Lasenby J, Guo H, Tang J.
\newblock Pre-training molecular graph representation with 3d geometry.
\newblock In: International Conference on Learning Representations.
\newblock 2022

\bibitem{zaidi2023pretraining}
Zaidi S, Schaarschmidt M, Martens J, Kim H, Teh Y~W, Sanchez-Gonzalez A, Battaglia P, Pascanu R, Godwin J.
\newblock Pre-training via denoising for molecular property prediction.
\newblock In: International Conference on Learning Representations.
\newblock 2023

\bibitem{feng2022mgmae}
Feng J, Wang Z, Li~Y, Ding B, Wei Z, Xu~H.
\newblock {MGMAE}: Molecular representation learning by reconstructing heterogeneous graphs with a high mask ratio.
\newblock In: Proceedings of the 31st ACM International Conference on Information \& Knowledge Management.
\newblock 2022,  509--519

\bibitem{stark20223d}
St{\"a}rk H, Beaini D, Corso G, Tossou P, Dallago C, G{\"u}nnemann S, Li{\`o} P.
\newblock 3d infomax improves gnns for molecular property prediction.
\newblock In: International Conference on Machine Learning.
\newblock 2022,  20479--20502

\bibitem{zhou2023unimol}
Zhou G, Gao Z, Ding Q, Zheng H, Xu~H, Wei Z, Zhang L, Ke~G.
\newblock {Uni-Mol}: A universal 3d molecular representation learning framework.
\newblock In: International Conference on Learning Representations.
\newblock 2023

\bibitem{luo2023one}
Luo S, Chen T, Xu~Y, Zheng S, Liu T~Y, Wang L, He~D.
\newblock One transformer can understand both 2d \& 3d molecular data.
\newblock In: International Conference on Learning Representations.
\newblock 2023

\bibitem{pmlr-v202-liu23h}
Liu S, Du~W, Ma~Z~M, Guo H, Tang J.
\newblock A group symmetric stochastic differential equation model for molecule multi-modal pretraining.
\newblock In: International Conference on Machine Learning.
\newblock 23--29 Jul 2023,  21497--21526

\bibitem{ni2023sliced}
Ni~Y, Feng S, Ma~W~Y, Ma~Z~M, Lan Y.
\newblock {Sliced Denoising}: A physics-informed molecular pre-training method.
\newblock In: International Conference on Learning Representations.
\newblock 2023

\bibitem{feng2023fractional}
Feng S, Ni~Y, Lan Y, Ma~Z~M, Ma~W~Y.
\newblock Fractional denoising for 3d molecular pre-training.
\newblock In: International Conference on Machine Learning.
\newblock 2023,  9938--9961

\bibitem{liu2025denoisevae}
Liu Y, Chen J, Jiao R, Li~J, Huang W, Su~B.
\newblock Denoise{VAE}: Learning molecule-adaptive noise distributions for denoising-based 3d molecular pre-training.
\newblock In: The Thirteenth International Conference on Learning Representations.
\newblock 2025

\bibitem{wang2025molspectra}
Wang L, Liu S, Rong Y, Zhao D, Liu Q, Wu~S, Wang L.
\newblock Molspectra: Pre-training 3d molecular representation with multi-modal energy spectra.
\newblock In: The Thirteenth International Conference on Learning Representations.
\newblock 2025

\bibitem{wang2022lm}
Wang Z, Combs S~A, Brand R, Calvo M~R, Xu~P, Price G, Golovach N, Salawu E~O, Wise C~J, Ponnapalli S~P, others .
\newblock {LM-GVP}: an extensible sequence and structure informed deep learning framework for protein property prediction.
\newblock Scientific reports, 2022, 12(1): 6832

\bibitem{gligorijevic2021structure}
Gligorijevi{\'c} V, Renfrew P~D, Kosciolek T, Leman J~K, Berenberg D, Vatanen T, Chandler C, Taylor B~C, Fisk I~M, Vlamakis H, others .
\newblock Structure-based protein function prediction using graph convolutional networks.
\newblock Nature communications, 2021, 12(1): 3168

\bibitem{zhang2022protein}
Zhang Z, Xu~M, Jamasb A~R, Chenthamarakshan V, Lozano A, Das P, Tang J.
\newblock Protein representation learning by geometric structure pretraining.
\newblock In: International Conference on Learning Representations.
\newblock 2022

\bibitem{torng20173d}
Torng W, Altman R~B.
\newblock 3d deep convolutional neural networks for amino acid environment similarity analysis.
\newblock BMC bioinformatics, 2017, 18(1): 1--23

\bibitem{zhang2005tm}
Zhang Y, Skolnick J.
\newblock {TM}-align: a protein structure alignment algorithm based on the tm-score.
\newblock Nucleic acids research, 2005, 33(7): 2302--2309

\bibitem{eismann2020hierarchical}
Eismann S, Townshend R~J, Thomas N, Jagota M, Jing B, Dror R~O.
\newblock Hierarchical, rotation‐equivariant neural networks to select structural models of protein complexes.
\newblock Proteins: Structure, Function, and Bioinformatics, 2020, 89(5): 493–501

\bibitem{eismann2023protein}
Eismann S, Suriana P, Jing B, Townshend R~J, Dror R~O.
\newblock Protein model quality assessment using rotation-equivariant transformations on point clouds.
\newblock Proteins: Structure, Function, and Bioinformatics, 2023

\bibitem{chen20233d}
Chen C, Chen X, Morehead A, Wu~T, Cheng J.
\newblock 3d-equivariant graph neural networks for protein model quality assessment.
\newblock Bioinformatics, 2023, 39(1): btad030

\bibitem{tubiana2022scannet}
Tubiana J, Schneidman-Duhovny D, Wolfson H~J.
\newblock {ScanNet}: an interpretable geometric deep learning model for structure-based protein binding site prediction.
\newblock Nature Methods, 2022, 19(6): 730--739

\bibitem{zhang2023equipocket}
Zhang Y, Huang W, Wei Z, Yuan Y, Ding Z.
\newblock {EquiPocket}: an e (3)-equivariant geometric graph neural network for ligand binding site prediction.
\newblock arXiv preprint arXiv:2302.12177, 2023

\bibitem{meller2023predicting}
Meller A, Ward M~D, Borowsky J~H, Lotthammer J~M, others .
\newblock Predicting the locations of cryptic pockets from single protein structures using the pocketminer graph neural network.
\newblock Biophysical Journal, 2023, 122(3): 445a

\bibitem{ingraham2019generative}
Ingraham J, Garg V, Barzilay R, Jaakkola T.
\newblock Generative models for graph-based protein design.
\newblock Annual Conference on Neural Information Processing Systems, 2019, 32

\bibitem{tan2022generative}
Tan C, Gao Z, Xia J, Hu~B, Li~S~Z.
\newblock Generative de novo protein design with global context.
\newblock arXiv preprint arXiv:2204.10673, 2022

\bibitem{dauparas2022robust}
Dauparas J, Anishchenko I, Bennett N, Bai H, Ragotte R~J, Milles L~F, Wicky B~I, Courbet A, Haas d~R~J, Bethel N, others .
\newblock Robust deep learning--based protein sequence design using proteinmpnn.
\newblock Science, 2022, 378(6615): 49--56

\bibitem{gao2022pifold}
Gao Z, Tan C, Li~S~Z.
\newblock {PiFold}: Toward effective and efficient protein inverse folding.
\newblock In: International Conference on Learning Representations.
\newblock 2022

\bibitem{pmlr-v202-zheng23a}
Zheng Z, Deng Y, Xue D, Zhou Y, Ye~F, Gu~Q.
\newblock Structure-informed language models are protein designers.
\newblock In: Proceedings of the 40th International Conference on Machine Learning.
\newblock 23--29 Jul 2023,  42317--42338

\bibitem{gao2023kw}
Gao Z, Tan C, Chen X, Zhang Y, Xia J, Li~S, Li~S~Z.
\newblock {KW-Design}: Pushing the limit of protein deign via knowledge refinement.
\newblock In: International Conference on Learning Representations.
\newblock 2023

\bibitem{AlphaFold2021}
Jumper J, Evans R, Pritzel A, Green T, others .
\newblock Highly accurate protein structure prediction with {AlphaFold}.
\newblock Nature, 2021, 596(7873): 583--589

\bibitem{krishna2023generalized}
Krishna R, Wang J, Ahern W, Sturmfels P, Venkatesh P, Kalvet I, Lee G~R, Morey-Burrows F~S, Anishchenko I, Humphreys I~R, others .
\newblock Generalized biomolecular modeling and design with rosettafold all-atom.
\newblock bioRxiv, 2023,  2023--10

\bibitem{jing2023eigenfold}
Jing B, Erives E, Pao-Huang P, Corso G, Berger B, Jaakkola T~S.
\newblock {EigenFold}: Generative protein structure prediction with diffusion models.
\newblock In: ICLR 2023-Machine Learning for Drug Discovery workshop.
\newblock 2023

\bibitem{lin2023evolutionary}
Lin Z, Akin H, Rao R, Hie B, Zhu Z, Lu~W, Smetanin N, Verkuil R, Kabeli O, Shmueli Y, others .
\newblock Evolutionary-scale prediction of atomic-level protein structure with a language model.
\newblock Science, 2023, 379(6637): 1123--1130

\bibitem{fang2023method}
Fang X, Wang F, Liu L, He~J, Lin D, Xiang Y, Zhu K, Zhang X, Wu~H, Li~H, others .
\newblock A method for multiple-sequence-alignment-free protein structure prediction using a protein language model.
\newblock Nature Machine Intelligence, 2023, 5(10): 1087--1096

\bibitem{shi2023protein}
Shi C, Wang C, Lu~J, Zhong B, Tang J.
\newblock Protein sequence and structure co-design with equivariant translation.
\newblock In: International Conference on Learning Representations.
\newblock 2023

\bibitem{yue2025reqflow}
Yue A, Wang Z, Xu~H.
\newblock Reqflow: Rectified quaternion flow for efficient and high-quality protein backbone generation.
\newblock arXiv preprint arXiv:2502.14637, 2025

\bibitem{elnaggar2021prottrans}
Elnaggar A, Heinzinger M, Dallago C, Rehawi G, Wang Y, Jones L, Gibbs T, Feher T, Angerer C, Steinegger M, others .
\newblock Prottrans: Toward understanding the language of life through self-supervised learning.
\newblock IEEE transactions on pattern analysis and machine intelligence, 2021, 44(10): 7112--7127

\bibitem{chen2024xtrimopglm}
Chen B, Cheng X, Li~P, Geng Y~a, Gong J, Li~S, Bei Z, Tan X, Wang B, Zeng X, others .
\newblock {xTrimoPGLM}: unified 100b-scale pre-trained transformer for deciphering the language of protein.
\newblock arXiv preprint arXiv:2401.06199, 2024

\bibitem{ferruz2022protgpt2}
Ferruz N, Schmidt S, H{\"o}cker B.
\newblock Protgpt2 is a deep unsupervised language model for protein design.
\newblock Nature communications, 2022, 13(1): 4348

\bibitem{mansoor2021toward}
Mansoor S, Baek M, Madan U, Horvitz E.
\newblock Toward more general embeddings for protein design: Harnessing joint representations of sequence and structure.
\newblock bioRxiv, 2021,  2021--09

\bibitem{gao2023self}
Gao B, Jia Y, Mo~Y, Ni~Y, Ma~W~Y, Ma~Z~M, Lan Y.
\newblock Self-supervised pocket pretraining via protein fragment-surroundings alignment.
\newblock In: International Conference on Learning Representations.
\newblock 2023

\bibitem{wang2023multi}
Wang Z, Zhang Q, Shuang-Wei H, Yu~H, Jin X, Gong Z, Chen H.
\newblock Multi-level protein structure pre-training via prompt learning.
\newblock In: International Conference on Learning Representations.
\newblock 2023

\bibitem{gao2023drugclip}
Gao B, Qiang B, Tan H, Jia Y, Ren M, Lu~M, Liu J, Ma~W~Y, Lan Y.
\newblock {DrugCLIP}: Contrasive protein-molecule representation learning for virtual screening.
\newblock Annual Conference on Neural Information Processing Systems, 2024, 36

\bibitem{rives2019biological}
Rives A, Meier J, Sercu T, Goyal S, Lin Z, Liu J, Guo D, Ott M, Zitnick C~L, Ma~J, Fergus R.
\newblock Biological structure and function emerge from scaling unsupervised learning to 250 million protein sequences.
\newblock PNAS, 2019

\bibitem{guo2022self}
Guo Y, Wu~J, Ma~H, Huang J.
\newblock Self-supervised pre-training for protein embeddings using tertiary structures.
\newblock In: Proceedings of the AAAI Conference on Artificial Intelligence.
\newblock 2022,  6801--6809

\bibitem{yuan2024annotation}
Yuan C, Li~S, Ye~G, Zhang Y, Huang L~K, Huang W, Liu W, Yao J, Rong Y.
\newblock Annotation-guided protein design with multi-level domain alignment, 2024

\bibitem{igashov2024equivariant}
Igashov I, St{\"a}rk H, Vignac C, Schneuing A, Satorras V~G, Frossard P, Welling M, Bronstein M, Correia B.
\newblock Equivariant 3d-conditional diffusion model for molecular linker design.
\newblock Nature Machine Intelligence, 2024, 6(4): 417--427

\bibitem{imrie2020deep}
Imrie F, Bradley A~R, Schaar v.~d M, Deane C~M.
\newblock Deep generative models for 3d linker design.
\newblock Journal of Chemical Information and Modeling, 2020, 60(4): 1983--1995

\bibitem{duan2023accurate}
Duan C, Du~Y, Jia H, Kulik H~J.
\newblock Accurate transition state generation with an object-aware equivariant elementary reaction diffusion model.
\newblock arXiv preprint arXiv:2304.06174, 2023

\bibitem{jackson2021tsnet}
Jackson R, Zhang W, Pearson J.
\newblock {TSNet}: predicting transition state structures with tensor field networks and transfer learning.
\newblock Chemical Science, 2021, 12(29): 10022--10040

\bibitem{gainza2020deciphering}
Gainza P, Sverrisson F, Monti F, Rodola E, Boscaini D, Bronstein M, Correia B.
\newblock Deciphering interaction fingerprints from protein molecular surfaces using geometric deep learning.
\newblock Nature Methods, 2020, 17(2): 184--192

\bibitem{kong2023generalist}
Kong X, Huang W, Liu Y.
\newblock Generalist equivariant transformer towards 3d molecular interaction learning.
\newblock arXiv preprint arXiv:2306.01474, 2023

\bibitem{wang2022learning}
Wang L, Liu H, Liu Y, Kurtin J, Ji~S.
\newblock Learning hierarchical protein representations via complete 3d graph networks.
\newblock In: International Conference on Learning Representations.
\newblock 2022

\bibitem{zhao2023geometric}
Zhao K, Rong Y, Jiang B, Tang J, Zhang H, Yu~J~X, Zhao P.
\newblock Geometric graph learning for protein mutation effect prediction.
\newblock In: Proceedings of the 32nd ACM International Conference on Information and Knowledge Management.
\newblock 2023,  3412--3422

\bibitem{feng2023protein}
Feng S, Li~M, Jia Y, Ma~W~Y, Lan Y.
\newblock Protein-ligand binding representation learning from fine-grained interactions.
\newblock In: International Conference on Learning Representations.
\newblock 2023

\bibitem{jian2024general}
Jian Y, Wu~C, Reidenbach D, Krishnapriyan A~S.
\newblock General binding affinity guidance for diffusion models in structure-based drug design.
\newblock arXiv preprint arXiv:2406.16821, 2024

\bibitem{xue2025se}
Xue F, Zhang M, Li~S, Gao X, Wohlschlegel J~A, Huang W, Yang Y, Deng W.
\newblock Se (3)-equivariant ternary complex prediction towards target protein degradation.
\newblock arXiv preprint arXiv:2502.18875, 2025

\bibitem{stark2022equibind}
St{\"a}rk H, Ganea O, Pattanaik L, Barzilay D, Jaakkola T.
\newblock {E}qui{B}ind: Geometric deep learning for drug binding structure prediction.
\newblock In: International Conference on Machine Learning.
\newblock 17--23 Jul 2022,  20503--20521

\bibitem{lu2022tankbind}
Lu~W, Wu~Q, Zhang J, Rao J, Li~C, Zheng S.
\newblock {TANKB}ind: Trigonometry-aware neural networks for drug-protein binding structure prediction.
\newblock In: Annual Conference on Neural Information Processing Systems.
\newblock 2022

\bibitem{long2022zero}
Long S, Zhou Y, Dai X, Zhou H.
\newblock Zero-shot 3d drug design by sketching and generating.
\newblock Annual Conference on Neural Information Processing Systems, 2022, 35: 23894--23907

\bibitem{pei2023fabind}
Pei Q, Gao K, Wu~L, Zhu J, Xia Y, Xie S, Qin T, He~K, Liu T~Y, Yan R.
\newblock {FAB}ind: Fast and accurate protein-ligand binding.
\newblock In: Annual Conference on Neural Information Processing Systems.
\newblock 2023

\bibitem{huang2024re}
Huang Y, Zhang O, Wu~L, Tan C, Lin H, Gao Z, Li~S, Li~S, others .
\newblock {Re-Dock}: Towards flexible and realistic molecular docking with diffusion bridge.
\newblock arXiv preprint arXiv:2402.11459, 2024

\bibitem{peng2022pocket}
Peng X, Luo S, Guan J, Xie Q, Peng J, Ma~J.
\newblock {P}ocket2{M}ol: Efficient molecular sampling based on 3{D} protein pockets.
\newblock In: Proceedings of the 39th International Conference on Machine Learning.
\newblock 17--23 Jul 2022,  17644--17655

\bibitem{lin2022diffbp}
Lin H, Huang Y, Liu M, Li~X, Ji~S, Li~S~Z.
\newblock Diffbp: Generative diffusion of 3d molecules for target protein binding.
\newblock arXiv preprint arXiv:2211.11214, 2022

\bibitem{luo2021a}
Luo S, Guan J, Ma~J, Peng J.
\newblock A 3d generative model for structure-based drug design.
\newblock In: Annual Conference on Neural Information Processing Systems.
\newblock 2021

\bibitem{liu2022generating}
Liu M, Luo Y, Uchino K, Maruhashi K, Ji~S.
\newblock Generating 3d molecules for target protein binding.
\newblock In: International Conference on Machine Learning.
\newblock 2022,  13912--13924

\bibitem{zhang2023molecule}
Zhang Z, Min Y, Zheng S, Liu Q.
\newblock Molecule generation for target protein binding with structural motifs.
\newblock In: The Eleventh International Conference on Learning Representations.
\newblock 2023

\bibitem{lin2024functional}
Lin H, Huang Y, Zhang O, Liu Y, Wu~L, Li~S, Chen Z, Li~S~Z.
\newblock Functional-group-based diffusion for pocket-specific molecule generation and elaboration.
\newblock Annual Conference on Neural Information Processing Systems, 2024, 36

\bibitem{qiu2024structure}
Qiu K, Song Y, Yu~J, Ma~H, Cao Z, Zhang Z, Wu~Y, Zheng M, Zhou H, Ma~W~Y.
\newblock Structure-based molecule optimization via gradient-guided bayesian update.
\newblock arXiv e-prints, 2024,  arXiv--2411

\bibitem{pinheiro2024structure}
Pinheiro P~O, Jamasb A, Mahmood O, Sresht V, Saremi S.
\newblock Structure-based drug design by denoising voxel grids.
\newblock arXiv preprint arXiv:2405.03961, 2024

\bibitem{morehead2022geometric}
Morehead A, Chen C, Cheng J.
\newblock Geometric transformers for protein interface contact prediction.
\newblock In: International Conference on Learning Representation.
\newblock 2022

\bibitem{sverrisson2021fast}
Sverrisson F, Feydy J, Correia B~E, Bronstein M~M.
\newblock Fast end-to-end learning on protein surfaces.
\newblock In: Proceedings of the IEEE/CVF Conference on Computer Vision and Pattern Recognition.
\newblock 2021,  15272--15281

\bibitem{townshend2019end}
Townshend R, Bedi R, Suriana P, Dror R.
\newblock End-to-end learning on 3d protein structure for interface prediction.
\newblock Annual Conference on Neural Information Processing Systems, 2019, 32

\bibitem{rodrigues2021mmcsm}
Rodrigues C~H, Pires D~E, Ascher D~B.
\newblock {mmCSM-PPI}: predicting the effects of multiple point mutations on protein--protein interactions.
\newblock Nucleic Acids Research, 2021, 49(W1): W417--W424

\bibitem{liu2021deep}
Liu X, Luo Y, Li~P, Song S, Peng J.
\newblock Deep geometric representations for modeling effects of mutations on protein-protein binding affinity.
\newblock PLoS computational biology, 2021, 17(8): e1009284

\bibitem{ganea2022independent}
Ganea O~E, Huang X, Bunne C, Bian Y, Barzilay R, Jaakkola T~S, Krause A.
\newblock Independent {SE}(3)-equivariant models for end-to-end rigid protein docking.
\newblock In: International Conference on Learning Representations.
\newblock 2022

\bibitem{wang2023learning}
Wang Y, Shen Y, Chen S, Wang L, Fei Y, Zhou H.
\newblock Learning harmonic molecular representations on riemannian manifold.
\newblock In: International Conference on Learning Representations.
\newblock 2023

\bibitem{jin2022antibody}
Jin W, Barzilay R, Jaakkola T.
\newblock Antibody-antigen docking and design via hierarchical structure refinement.
\newblock In: International Conference on Machine Learning.
\newblock 2022,  10217--10227

\bibitem{ketata2023diffdock}
Ketata M~A, Laue C, Mammadov R, Stark H, Wu~M, Corso G, Marquet C, Barzilay R, Jaakkola T~S.
\newblock {DiffDock-PP}: Rigid protein-protein docking with diffusion models.
\newblock In: ICLR 2023-Machine Learning for Drug Discovery workshop.
\newblock 2023

\bibitem{ji2023syndock}
Ji~Y, Bian Y, Fu~G, Zhao P, Luo P.
\newblock {SyNDock}: N rigid protein docking via learnable group synchronization.
\newblock arXiv preprint arXiv:2305.15156, 2023

\bibitem{evans2021protein}
Evans R, O’Neill M, Pritzel A, Antropova N, Senior A, Green T, {\v{Z}}{\'\i}dek A, Bates R, Blackwell S, Yim J, others .
\newblock Protein complex prediction with alphafold-multimer.
\newblock BioRxiv, 2021,  2021--10

\bibitem{sverrisson2022physics}
Sverrisson F, Feydy J, Southern J, Bronstein M~M, Correia B~E.
\newblock Physics-informed deep neural network for rigid-body protein docking.
\newblock In: MLDD workshop of ICLR 2022.
\newblock 2022

\bibitem{yu2024rigid}
Yu~Z, Huang W, Liu Y.
\newblock Rigid protein-protein docking via equivariant elliptic-paraboloid interface prediction.
\newblock In: International Conference on Learning Representations.
\newblock 2024

\bibitem{wu2024neural}
Wu~H, Liu W, Bian Y, Wu~J, Yang N, Yan J.
\newblock Neural probabilistic protein-protein docking via a differentiable energy model.
\newblock In: The Twelfth International Conference on Learning Representations.
\newblock 2024

\bibitem{luo2022antigenspecific}
Luo S, Su~Y, Peng X, Wang S, Peng J, Ma~J.
\newblock Antigen-specific antibody design and optimization with diffusion-based generative models for protein structures.
\newblock In: Annual Conference on Neural Information Processing Systems.
\newblock 2022

\bibitem{jin2022iterative}
Jin W, Wohlwend J, Barzilay R, Jaakkola T~S.
\newblock Iterative refinement graph neural network for antibody sequence-structure co-design.
\newblock In: International Conference on Learning Representations.
\newblock 2022

\bibitem{gao2022incorporating}
Gao K, Wu~L, Zhu J, Peng T, Xia Y, He~L, Xie S, Qin T, Liu H, He~K, others .
\newblock Incorporating pre-training paradigm for antibody sequence-structure co-design.
\newblock bioRxiv, 2022,  2022--11

\bibitem{tan2023cross}
Tan C, Gao Z, Li~S~Z.
\newblock Cross-gate mlp with protein complex invariant embedding is a one-shot antibody designer.
\newblock arXiv e-prints, 2023,  arXiv--2305

\bibitem{verma2023abode}
Verma Y, Heinonen M, Garg V.
\newblock {A}b{ODE}: Ab initio antibody design using conjoined {ODE}s.
\newblock In: International Conference on Machine Learning.
\newblock 23--29 Jul 2023,  35037--35050

\bibitem{martinkus2023abdiffuser}
Martinkus K, Ludwiczak J, LIANG W~C, Lafrance-Vanasse J, Hotzel I, Rajpal A, Wu~Y, Cho K, Bonneau R, Gligorijevic V, others .
\newblock {AbDiffuser}: full-atom generation of in-vitro functioning antibodies.
\newblock In: Annual Conference on Neural Information Processing Systems.
\newblock 2023

\bibitem{wu2024fast}
Wu~F, Zhao Y, Wu~J, Jiang B, He~B, Huang L, Qin C, Yang F, Huang N, Xiao Y, others .
\newblock Fast and accurate modeling and design of antibody-antigen complex using tfold.
\newblock bioRxiv, 2024,  2024--02

\bibitem{lin2024geoab}
Lin H, Wu~L, Huang Y, Liu Y, Zhang O, Zhou Y, Sun R, Li~S~Z.
\newblock Geo{AB}: Towards realistic antibody design and reliable affinity maturation.
\newblock In: Forty-first International Conference on Machine Learning.
\newblock 2024

\bibitem{wu2024relation}
Wu~L, Lin H, Huang Y, Gao Z, Tan C, Liu Y, Wu~T, Li~S~Z.
\newblock Relation-aware equivariant graph networks for epitope-unknown antibody design and specificity optimization.
\newblock arXiv preprint arXiv:2501.00013, 2024

\bibitem{xie2023helixgan}
Xie X, Valiente P~A, Kim P~M.
\newblock Helixgan a deep-learning methodology for conditional de novo design of $\alpha$-helix structures.
\newblock Bioinformatics, 2023, 39(1): btad036

\bibitem{lin2024ppflow}
Lin H, Zhang O, Zhao H, Jiang D, Wu~L, Liu Z, Huang Y, Li~S~Z.
\newblock {PPFLOW}: Target-aware peptide design with torsional flow matching.
\newblock In: Forty-first International Conference on Machine Learning.
\newblock 2024

\bibitem{xie2018crystal}
Xie T, Grossman J~C.
\newblock Crystal graph convolutional neural networks for an accurate and interpretable prediction of material properties.
\newblock Physical review letters, 2018, 120(14): 145301

\bibitem{chen2019graph}
Chen C, Ye~W, Zuo Y, Zheng C, Ong S~P.
\newblock Graph networks as a universal machine learning framework for molecules and crystals.
\newblock Chemistry of Materials, 2019, 31(9): 3564--3572

\bibitem{choudhary2021atomistic}
Choudhary K, DeCost B.
\newblock Atomistic line graph neural network for improved materials property predictions.
\newblock npj Computational Materials, 2021, 7(1): 185

\bibitem{kaba2022equivariant}
Kaba O, Ravanbakhsh S.
\newblock Equivariant networks for crystal structures.
\newblock Annual Conference on Neural Information Processing Systems, 2022, 35: 4150--4164

\bibitem{yan2022periodic}
Yan K, Liu Y, Lin Y, Ji~S.
\newblock Periodic graph transformers for crystal material property prediction.
\newblock In: Annual Conference on Neural Information Processing Systems.
\newblock 2022

\bibitem{magar2022crystal}
Magar R, Wang Y, Barati~Farimani A.
\newblock Crystal twins: self-supervised learning for crystalline material property prediction.
\newblock npj Computational Materials, 2022, 8(1): 231

\bibitem{yu2023crystal}
Yu~H, Song Y, Hu~J, Guo C, Yang B.
\newblock A crystal-specific pre-training framework for crystal material property prediction.
\newblock arXiv preprint arXiv:2306.05344, 2023

\bibitem{song2024diffusion}
Song Z, Meng Z, King I.
\newblock A diffusion-based pre-training framework for crystal property prediction.
\newblock In: Proceedings of the AAAI Conference on Artificial Intelligence.
\newblock 2024,  8993--9001

\bibitem{xie2021crystal}
Xie T, Fu~X, Ganea O~E, Barzilay R, Jaakkola T~S.
\newblock Crystal diffusion variational autoencoder for periodic material generation.
\newblock In: International Conference on Learning Representations.
\newblock 2021

\bibitem{jiao2024space}
Jiao R, Huang W, Liu Y, Zhao D, Liu Y.
\newblock Space group constrained crystal generation.
\newblock arXiv preprint arXiv:2402.03992, 2024

\bibitem{zeni2023mattergen}
Zeni C, Pinsler R, Z{\"u}gner D, Fowler A, Horton M, Fu~X, Shysheya S, Crabb{\'e} J, Sun L, Smith J, others .
\newblock {MatterGen}: a generative model for inorganic materials design.
\newblock arXiv preprint arXiv:2312.03687, 2023

\bibitem{li2024powder}
Li~Q, Jiao R, Wu~L, Zhu T, Huang W, Jin S, Liu Y, Weng H, Chen X.
\newblock Powder diffraction crystal structure determination using generative models.
\newblock arXiv preprint arXiv:2409.04727, 2024

\bibitem{lin2024equivariant}
Lin P, Chen P, Jiao R, Mo~Q, Jianhuan C, Huang W, Liu Y, Huang D, Lu~Y.
\newblock Equivariant diffusion for crystal structure prediction.
\newblock In: Forty-first International Conference on Machine Learning.
\newblock 2024

\bibitem{millerflowmm}
Miller B~K, Chen R~T, Sriram A, Wood B~M.
\newblock Flowmm: Generating materials with riemannian flow matching.
\newblock In: Forty-first International Conference on Machine Learning.
\newblock 2024

\bibitem{wu2025a}
Wu~H, Song Y, Gong J, Cao Z, Ouyang Y, Zhang J, Zhou H, Ma~W~Y, Liu J.
\newblock A periodic bayesian flow for material generation.
\newblock In: The Thirteenth International Conference on Learning Representations.
\newblock 2025

\bibitem{zhang2022physics}
Zhang S, Liu Y, Xie L.
\newblock Physics-aware graph neural network for accurate rna 3d structure prediction.
\newblock arXiv preprint arXiv:2210.16392, 2022

\bibitem{li2025sizegeneralizable}
Li~Z, Cen J, Huang W, Wang T, Song L.
\newblock Size-generalizable {RNA} structure evaluation by exploring hierarchical geometries.
\newblock In: The Thirteenth International Conference on Learning Representations.
\newblock 2025

\bibitem{greff2022kubric}
Greff K, Belletti F, Beyer L, Doersch C, Du~Y, Duckworth D, Fleet D~J, Gnanapragasam D, Golemo F, Herrmann C, others .
\newblock Kubric: A scalable dataset generator.
\newblock In: Proceedings of the IEEE/CVF Conference on Computer Vision and Pattern Recognition.
\newblock 2022,  3749--3761

\bibitem{bear2021physion}
Bear D, Wang E, Mrowca D, others .
\newblock Physion: Evaluating physical prediction from vision in humans and machines.
\newblock In: Thirty-fifth Conference on Neural Information Processing Systems Datasets and Benchmarks Track (Round 1).
\newblock 2021

\bibitem{yu2016more}
Yu~K~T, Bauza M, Fazeli N, Rodriguez A.
\newblock More than a million ways to be pushed. a high-fidelity experimental dataset of planar pushing.
\newblock In: IROS.
\newblock 2016,  30--37

\bibitem{townshend2021atomd}
Townshend R~J~L, V{\"o}gele M, Suriana P~A, others .
\newblock {ATOM}3d: Tasks on molecules in three dimensions.
\newblock In: Annual Conference on Neural Information Processing Systems Datasets and Benchmarks Track.
\newblock 2021

\bibitem{xu2021learning}
Xu~M, Luo S, Bengio Y, Peng J, Tang J.
\newblock Learning neural generative dynamics for molecular conformation generation.
\newblock In: International Conference on Learning Representations.
\newblock 2020

\bibitem{chmiela2017machine}
Chmiela S, Tkatchenko A, Sauceda H~E, Poltavsky I, Sch{\"u}tt K~T, M{\"u}ller K~R.
\newblock Machine learning of accurate energy-conserving molecular force fields.
\newblock Science advances, 2017, 3(5): e1603015

\bibitem{oc22_dataset}
Tran R, Lan J, Shuaibi M, Wood B~M, Goyal S, Das A, Heras-Domingo J, Kolluru A, Rizvi A, Shoghi N, others .
\newblock The open catalyst 2022 (oc22) dataset and challenges for oxide electrocatalysts.
\newblock ACS Catalysis, 2023, 13(5): 3066--3084

\bibitem{seyler5108170molecular}
Seyler S, Beckstein O.
\newblock Molecular dynamics trajectory for benchmarking mdanalysis, 6 2017, 2017

\bibitem{lindorff2011fast}
Lindorff-Larsen K, Piana S, Dror R~O, Shaw D~E.
\newblock How fast-folding proteins fold.
\newblock Science, 2011, 334(6055): 517--520

\bibitem{axelrod2022geom}
Axelrod S, Gomez-Bombarelli R.
\newblock {GEOM}, energy-annotated molecular conformations for property prediction and molecular generation.
\newblock Scientific Data, 2022, 9(1): 185

\bibitem{wang2023automated}
Wang X, Zhao H, Tu~W~w, Yao Q.
\newblock Automated 3d pre-training for molecular property prediction.
\newblock In: Proceedings of the 29th ACM SIGKDD Conference on Knowledge Discovery and Data Mining.
\newblock 2023,  2419--2430

\bibitem{isert2022qmugs}
Isert C, Atz K, Jim{\'e}nez-Luna J, Schneider G.
\newblock Qmugs, quantum mechanical properties of drug-like molecules.
\newblock Scientific Data, 2022, 9(1): 273

\bibitem{ashburner2000gene}
Ashburner M, Ball C~A, Blake J~A, Botstein D, Butler H, Cherry J~M, Davis A~P, Dolinski K, Dwight S~S, Eppig J~T, others .
\newblock {Gene Ontology}: tool for the unification of biology.
\newblock Nature genetics, 2000, 25(1): 25--29

\bibitem{bairoch2000enzyme}
Bairoch A.
\newblock The enzyme database in 2000.
\newblock Nucleic acids research, 2000, 28(1): 304--305

\bibitem{orengo1997cath}
Orengo C~A, Michie A~D, Jones S, Jones D~T, Swindells M~B, Thornton J~M.
\newblock Cath--a hierarchic classification of protein domain structures.
\newblock Structure, 1997, 5(8): 1093--1109

\bibitem{xue2022multimodal}
Xue Y, Liu Z, Fang X, Wang F.
\newblock Multimodal pre-training model for sequence-based prediction of protein-protein interaction.
\newblock In: Machine Learning in Computational Biology.
\newblock 2022,  34--46

\bibitem{chandonia2019scope}
Chandonia J~M, Fox N~K, Brenner S~E.
\newblock {SCOPe}: classification of large macromolecular structures in the structural classification of proteins—extended database.
\newblock Nucleic acids research, 2019, 47(D1): D475--D481

\bibitem{heinzinger2023prostt5}
Heinzinger M, Weissenow K, Sanchez J~G, Henkel A, Steinegger M, Rost B.
\newblock {ProstT5}: Bilingual language model for protein sequence and structure.
\newblock bioRxiv, 2023,  2023--07

\bibitem{bepler2021learning}
Bepler T, Berger B.
\newblock Learning the protein language: Evolution, structure, and function.
\newblock Cell systems, 2021, 12(6): 654--669

\bibitem{rao2019evaluating}
Rao R, Bhattacharya N, Thomas N, Duan Y, Chen P, Canny J, Abbeel P, Song Y.
\newblock Evaluating protein transfer learning with tape.
\newblock Annual Conference on Neural Information Processing Systems, 2019, 32

\bibitem{varadi2022alphafold}
Varadi M, Anyango S, Deshpande M, others .
\newblock {AlphaFold Protein Structure Database}: massively expanding the structural coverage of protein-sequence space with high-accuracy models.
\newblock Nucleic acids research, 2022, 50(D1): D439--D444

\bibitem{gao2022alphadesign}
Gao Z, Tan C, Li~S~Z.
\newblock Alphadesign: A graph protein design method and benchmark on alphafolddb.
\newblock arXiv preprint arXiv:2202.01079, 2022

\bibitem{uniprot2023uniprot}
Consortium T~U.
\newblock {UniProt}: the universal protein knowledgebase in 2023.
\newblock Nucleic Acids Research, 2023, 51(D1): D523--D531

\bibitem{almagro2017deeploc}
Almagro~Armenteros J~J, S{\o}nderby C~K, S{\o}nderby S~K, Nielsen H, Winther O.
\newblock Deeploc: prediction of protein subcellular localization using deep learning.
\newblock Bioinformatics, 2017, 33(21): 3387--3395

\bibitem{steinegger2018clustering}
Steinegger M, S{\"o}ding J.
\newblock Clustering huge protein sequence sets in linear time.
\newblock Nature communications, 2018, 9(1): 2542

\bibitem{klausen2019netsurfp}
Klausen M~S, Jespersen M~C, Nielsen H, others .
\newblock {NetSurfP-2.0}: Improved prediction of protein structural features by integrated deep learning.
\newblock Proteins: Structure, Function, and Bioinformatics, 2019, 87(6): 520--527

\bibitem{xu2022peer}
Xu~M, Zhang Z, Lu~J, Zhu Z, Zhang Y, Chang M, Liu R, Tang J.
\newblock Peer: a comprehensive and multi-task benchmark for protein sequence understanding.
\newblock Annual Conference on Neural Information Processing Systems, 2022, 35: 35156--35173

\bibitem{kryshtafovych2019critical}
Kryshtafovych A, Schwede T, Topf M, Fidelis K, Moult J.
\newblock Critical assessment of methods of protein structure prediction (casp)—round xiii.
\newblock Proteins: Structure, Function, and Bioinformatics, 2019, 87(12): 1011--1020

\bibitem{berman2000protein}
Berman H~M, Westbrook J, Feng Z, Gilliland G, Bhat T~N, Weissig H, Shindyalov I~N, Bourne P~E.
\newblock The protein data bank.
\newblock Nucleic acids research, 2000, 28(1): 235--242

\bibitem{sterling2015zinc}
Sterling T, Irwin J~J.
\newblock Zinc 15 – ligand discovery for everyone.
\newblock Journal of Chemical Information and Modeling, 2015, 55(11): 2324--2337

\bibitem{su2019comparative}
Su~M, Yang Q, Du~Y, Feng G, Liu Z, Li~Y, Wang R.
\newblock Comparative assessment of scoring functions: The casf-2016 update.
\newblock Journal of Chemical Information and Modeling, 2019, 59(2): 895--913

\bibitem{schreiner2022transition1x}
Schreiner M, Bhowmik A, Vegge T, Busk J, Winther O.
\newblock Transition1x-a dataset for building generalizable reactive machine learning potentials.
\newblock Scientific Data, 2022, 9(1): 779

\bibitem{francouer2020three}
Francoeur P~G, Masuda T, Sunseri J, Jia A, Iovanisci R~B, Snyder I, Koes D~R.
\newblock Three-dimensional convolutional neural networks and a cross-docked data set for structure-based drug design.
\newblock Journal of Chemical Information and Modeling, 2020, 60(9): 4200--4215

\bibitem{morehead2021dips}
Morehead A, Chen C, Sedova A, Cheng J.
\newblock Dips-plus: The enhanced database of interacting protein structures for interface prediction.
\newblock Scientific Data, 2023, 10(1): 509

\bibitem{stark2006biogrid}
Stark C, Breitkreutz B~J, Reguly T, Boucher L, Breitkreutz A, Tyers M.
\newblock {BioGRID}: a general repository for interaction datasets.
\newblock Nucleic acids research, 2006, 34(suppl\_1): D535--D539

\bibitem{hallee2023protein}
Hallee L, Gleghorn J~P.
\newblock Protein-protein interaction prediction is achievable with large language models.
\newblock bioRxiv, 2023,  2023--06

\bibitem{vreven2015updates}
Vreven T, Moal I~H, Vangone A, Pierce B~G, others .
\newblock Updates to the integrated protein--protein interaction benchmarks: docking benchmark version 5 and affinity benchmark version 2.
\newblock Journal of molecular biology, 2015, 427(19): 3031--3041

\bibitem{jankauskaite2019skempi}
Jankauskait{\.e} J, Jim{\'e}nez-Garc{\'\i}a B, Dapk{\=u}nas J, Fern{\'a}ndez-Recio J, Moal I~H.
\newblock {SKEMPI 2.0}: an updated benchmark of changes in protein--protein binding energy, kinetics and thermodynamics upon mutation.
\newblock Bioinformatics, 2019, 35(3)

\bibitem{raybould2021cov}
Raybould M~I, Kovaltsuk A, Marks C, Deane C~M.
\newblock {CoV-AbDab}: the coronavirus antibody database.
\newblock Bioinformatics, 2021, 37(5): 734--735

\bibitem{wen2019pepbdb}
Wen Z, He~J, Tao H, Huang S~Y.
\newblock {PepBDB}: a comprehensive structural database of biological peptide--protein interactions.
\newblock Bioinformatics, 2019, 35(1): 175--177

\bibitem{lei2021deep}
Lei Y, Li~S, Liu Z, Wan F, Tian T, Li~S, Zhao D, Zeng J.
\newblock A deep-learning framework for multi-level peptide--protein interaction prediction.
\newblock Nature communications, 2021, 12(1): 5465

\bibitem{tsaban2022harnessing}
Tsaban T, Varga J~K, Avraham O, Ben-Aharon Z, Khramushin A, Schueler-Furman O.
\newblock Harnessing protein folding neural networks for peptide--protein docking.
\newblock Nature communications, 2022, 13(1): 176

\bibitem{jain2013commentary}
Jain A, Ong S~P, Hautier G, others .
\newblock Commentary: The materials project: A materials genome approach to accelerating materials innovation.
\newblock APL materials, 2013, 1(1): 011002

\bibitem{castelli2012new}
Castelli I~E, Landis D~D, Thygesen K~S, Dahl S, Chorkendorff I, Jaramillo T~F, Jacobsen K~W.
\newblock New cubic perovskites for one-and two-photon water splitting using the computational materials repository.
\newblock Energy \& Environmental Science, 2012, 5(10): 9034--9043

\bibitem{castelli2012computational}
Castelli I~E, Olsen T, Datta S, Landis D~D, Dahl S, Thygesen K~S, Jacobsen K~W.
\newblock Computational screening of perovskite metal oxides for optimal solar light capture.
\newblock Energy \& Environmental Science, 2012, 5(2): 5814--5819

\bibitem{carbon2020data}
Pickard C~J.
\newblock Airss data for carbon at 10gpa and the c+n+h+o system at 1gpa, 2020

\bibitem{choudhary2020joint}
Choudhary K, Garrity K~F, Reid A~C, DeCost B, Biacchi A~J, Hight~Walker A~R, Trautt Z, Hattrick-Simpers J, Kusne A~G, Centrone A, others .
\newblock The joint automated repository for various integrated simulations (jarvis) for data-driven materials design.
\newblock npj computational materials, 2020, 6(1): 173

\bibitem{choudhary2018machine}
Choudhary K, DeCost B, Tavazza F.
\newblock Machine learning with force-field-inspired descriptors for materials: Fast screening and mapping energy landscape.
\newblock Physical review materials, 2018, 2(8): 083801

\bibitem{watkins2020farfar2}
Watkins A~M, Rangan R, Das R.
\newblock {FARFAR2}: improved de novo rosetta prediction of complex global rna folds.
\newblock Structure, 2020, 28(8): 963--976

\bibitem{liu2024segno}
Liu Y, Cheng J, Zhao H, Xu~T, Zhao P, Tsung F, Li~J, Rong Y.
\newblock {SEGNO}: Generalizing equivariant graph neural networks with physical inductive biases.
\newblock In: The Twelfth International Conference on Learning Representations.
\newblock 2024

\bibitem{downs1989review}
Downs G~M, Gillet V~J, Holliday J~D, Lynch M~F.
\newblock Review of ring perception algorithms for chemical graphs.
\newblock Journal of chemical information and computer sciences, 1989, 29(3): 172--187

\bibitem{lipinski2012experimental}
Lipinski C~A, Lombardo F, Dominy B~W, Feeney P~J.
\newblock Experimental and computational approaches to estimate solubility and permeability in drug discovery and development settings.
\newblock Advanced drug delivery reviews, 2012, 64: 4--17

\bibitem{gowers2016mdanalysis}
Gowers R~J, Linke M, Barnoud J, Reddy T~J, Melo M~N, Seyler S~L, Domanski J, Dotson D~L, Buchoux S, Kenney I~M, others .
\newblock {MDAnalysis}: a python package for the rapid analysis of molecular dynamics simulations.
\newblock In: Proceedings of the 15th python in science conference.
\newblock 2016,  105

\bibitem{ronneberger2015u}
Ronneberger O, Fischer P, Brox T.
\newblock U-net: Convolutional networks for biomedical image segmentation.
\newblock In: Medical Image Computing and Computer-Assisted Intervention--MICCAI 2015: 18th International Conference, Munich, Germany, October 5-9, 2015, Proceedings, Part III 18.
\newblock 2015,  234--241

\bibitem{huang2020augmented}
Huang C~W, Dinh L, Courville A.
\newblock Augmented normalizing flows: Bridging the gap between generative flows and latent variable models.
\newblock arXiv preprint arXiv:2002.07101, 2020

\bibitem{liberti2014euclidean}
Liberti L, Lavor C, Maculan N, Mucherino A.
\newblock Euclidean distance geometry and applications.
\newblock SIAM review, 2014, 56(1): 3--69

\bibitem{kingma2013auto}
Kingma D~P, Welling M.
\newblock Auto-encoding variational bayes.
\newblock stat, 2014, 1050: 1

\bibitem{Wang2025DiffusionMF}
Wang L, Song C, Liu Z, Rong Y, Liu Q, Wu~S, Wang L.
\newblock Diffusion models for molecules: A survey of methods and tasks.
\newblock arXiv preprint arXiv:2502.09511, 2025

\bibitem{wang2019smiles}
Wang S, Guo Y, Wang Y, Sun H, Huang J.
\newblock Smiles-bert: large scale unsupervised pre-training for molecular property prediction.
\newblock In: Proceedings of the 10th ACM international conference on bioinformatics, computational biology and health informatics.
\newblock 2019,  429--436

\bibitem{hu2020strategies}
Hu~W, Liu B, Gomes J, Zitnik M, Liang P, Pande V, Leskovec J.
\newblock Strategies for pre-training graph neural networks.
\newblock In: International Conference on Learning Representations.
\newblock 2020

\bibitem{rong2020self}
Rong Y, Bian Y, Xu~T, Xie W, WEI Y, Huang W, Huang J.
\newblock Self-supervised graph transformer on large-scale molecular data.
\newblock In: Annual Conference on Neural Information Processing Systems.
\newblock 2020,  12559--12571

\bibitem{hu2020ogb}
Hu~W, Fey M, Zitnik M, Dong Y, Ren H, Liu B, Catasta M, Leskovec J.
\newblock Open graph benchmark: Datasets for machine learning on graphs.
\newblock Annual Conference on Neural Information Processing Systems, 2020, 33: 22118--22133

\bibitem{doi:10.1021/acs.jcim.7b00083}
Nakata M, Shimazaki T.
\newblock {PubChemQC Project}: A large-scale first-principles electronic structure database for data-driven chemistry.
\newblock Journal of Chemical Information and Modeling, 2017, 57(6): 1300--1308

\bibitem{pracht2020automated}
Pracht P, Bohle F, Grimme S.
\newblock Automated exploration of the low-energy chemical space with fast quantum chemical methods.
\newblock The journal for high quality research in physical chemistry, chemical physics and biophysical chemistry, 2020, 22(14): 7169--7192

\bibitem{hung2011protein}
Hung M~C, Link W.
\newblock Protein localization in disease and therapy.
\newblock Journal of cell science, 2011, 124(20): 3381--3392

\bibitem{dallago2021flip}
Dallago C, Mou J, Johnston K~E, Wittmann B, Bhattacharya N, Goldman S, Madani A, Yang K~K.
\newblock {FLIP}: Benchmark tasks in fitness landscape inference for proteins.
\newblock In: Thirty-fifth Conference on Neural Information Processing Systems Datasets and Benchmarks Track (Round 2).
\newblock 2021

\bibitem{krivak2015improving}
Kriv{\'a}k R, Hoksza D.
\newblock Improving protein-ligand binding site prediction accuracy by classification of inner pocket points using local features.
\newblock Journal of cheminformatics, 2015, 7(1): 1--13

\bibitem{le2009fpocket}
Le~Guilloux V, Schmidtke P, Tuffery P.
\newblock Fpocket: an open source platform for ligand pocket detection.
\newblock BMC bioinformatics, 2009, 10(1): 1--11

\bibitem{jimenez2017deepsite}
Jim{\'e}nez J, Doerr S, Mart{\'\i}nez-Rosell G, Rose A~S, De~Fabritiis G.
\newblock {DeepSite}: protein-binding site predictor using 3d-convolutional neural networks.
\newblock Bioinformatics, 2017, 33(19): 3036--3042

\bibitem{mylonas2021deepsurf}
Mylonas S~K, Axenopoulos A, Daras P.
\newblock {DeepSurf}: a surface-based deep learning approach for the prediction of ligand binding sites on proteins.
\newblock Bioinformatics, 2021, 37(12): 1681--1690

\bibitem{lin2022language}
Lin Z, Akin H, Rao R, Hie B, Zhu Z, Lu~W, Smetanin N, Santos~Costa d~A, Fazel-Zarandi M, Sercu T, Candido S, others .
\newblock Language models of protein sequences at the scale of evolution enable accurate structure prediction.
\newblock bioRxiv, 2022

\bibitem{suzek2015uniref}
Suzek B~E, Wang Y, Huang H, McGarvey P~B, Wu~C~H, Consortium U.
\newblock {UniRef clusters}: a comprehensive and scalable alternative for improving sequence similarity searches.
\newblock Bioinformatics, 2015, 31(6): 926--932

\bibitem{rao2020transformer}
Rao R, Meier J, Sercu T, Ovchinnikov S, Rives A.
\newblock Transformer protein language models are unsupervised structure learners.
\newblock In: International Conference on Learning Representations.
\newblock 2020

\bibitem{wu2022survey}
Wu~L, Huang Y, Lin H, Li~S~Z.
\newblock A survey on protein representation learning: Retrospect and prospect.
\newblock arXiv preprint arXiv:2301.00813, 2022

\bibitem{hussain2010computationally}
Hussain J, Rea C.
\newblock Computationally efficient algorithm to identify matched molecular pairs (mmps) in large data sets.
\newblock Journal of Chemical Information and Modeling, 2010, 50(3): 339--348

\bibitem{lin2023d3fg}
Lin H, Huang Y, Zhang O, Liu Y, Wu~L, Li~S, Chen Z, Li~S~Z.
\newblock Functional-group-based diffusion for pocket-specific molecule generation and elaboration.
\newblock In: Thirty-seventh Conference on Neural Information Processing Systems.
\newblock 2023

\bibitem{wang2004pdbbind}
Wang R, Fang X, Lu~Y, Wang S.
\newblock {The PDBbind database}: Collection of binding affinities for protein- ligand complexes with known three-dimensional structures.
\newblock Journal of medicinal chemistry, 2004, 47(12): 2977--2980

\bibitem{kastritis2011structure}
Kastritis P~L, Moal I~H, Hwang H, Weng Z, Bates P~A, Bonvin A~M, Janin J.
\newblock A structure-based benchmark for protein--protein binding affinity.
\newblock Protein Science, 2011, 20(3): 482--491

\bibitem{moal2012skempi}
Moal I~H, Fern{\'a}ndez-Recio J.
\newblock {SKEMPI}: a structural kinetic and energetic database of mutant protein interactions and its use in empirical models.
\newblock Bioinformatics, 2012, 28(20): 2600--2607

\bibitem{fosgerau2015peptide}
Fosgerau K, Hoffmann T.
\newblock Peptide therapeutics: current status and future directions.
\newblock Drug discovery today, 2015, 20(1): 122--128

\bibitem{lee2019comprehensive}
Lee A~C~L, Harris J~L, Khanna K~K, Hong J~H.
\newblock A comprehensive review on current advances in peptide drug development and design.
\newblock International journal of molecular sciences, 2019, 20(10): 2383

\bibitem{bhardwaj2016accurate}
Bhardwaj G, Mulligan V~K, Bahl C~D, Gilmore J~M, Harvey P~J, Cheneval O, Buchko G~W, Pulavarti S~V, Kaas Q, Eletsky A, others .
\newblock Accurate de novo design of hyperstable constrained peptides.
\newblock Nature, 2016, 538(7625): 329--335

\bibitem{cao2022design}
Cao L, Coventry B, Goreshnik I, Huang B, Sheffler W, Park J~S, Jude K~M, Markovi{\'c} I, Kadam R~U, Verschueren K~H, others .
\newblock Design of protein-binding proteins from the target structure alone.
\newblock Nature, 2022, 605(7910): 551--560

\bibitem{zbontar2021barlow}
Zbontar J, Jing L, Misra I, LeCun Y, Deny S.
\newblock Barlow twins: Self-supervised learning via redundancy reduction.
\newblock In: International Conference on Machine Learning.
\newblock 2021,  12310--12320

\bibitem{chen2021exploring}
Chen X, He~K.
\newblock Exploring simple siamese representation learning.
\newblock In: Proceedings of the IEEE/CVF conference on computer vision and pattern recognition.
\newblock 2021,  15750--15758

\bibitem{geiger2022e3nn}
Geiger M, Smidt T.
\newblock {e3nn}: Euclidean neural networks.
\newblock arXiv preprint arXiv:2207.09453, 2022

\bibitem{das2007automated}
Das R, Baker D.
\newblock Automated de novo prediction of native-like rna tertiary structures.
\newblock Proceedings of the National Academy of Sciences, 2007, 104(37): 14664--14669

\bibitem{radford2018improving}
Radford A, Narasimhan K, Salimans T, Sutskever I, others .
\newblock Improving language understanding by generative pre-training, 2018

\bibitem{radford2019language}
Radford A, Wu~J, Child R, Luan D, Amodei D, Sutskever I, others .
\newblock Language models are unsupervised multitask learners.
\newblock OpenAI blog, 2019, 1(8): 9

\bibitem{brown2020language}
Brown T, Mann B, Ryder N, Subbiah M, Kaplan J~D, Dhariwal P, Neelakantan A, Shyam P, Sastry G, Askell A, others .
\newblock Language models are few-shot learners.
\newblock Annual Conference on Neural Information Processing Systems, 2020, 33: 1877--1901

\bibitem{reed2022generalist}
Reed S, Zolna K, Parisotto E, Colmenarejo S~G, Novikov A, Barth-maron G, Gim{\'e}nez M, Sulsky Y, Kay J, Springenberg J~T, Eccles T, Bruce J, Razavi A, Edwards A, Heess N, Chen Y, Hadsell R, Vinyals O, Bordbar M, Freitas d~N.
\newblock A generalist agent.
\newblock Transactions on Machine Learning Research, 2022

\bibitem{merchant2023scaling}
Merchant A, Batzner S, Schoenholz S~S, Aykol M, Cheon G, Cubuk E~D.
\newblock Scaling deep learning for materials discovery.
\newblock Nature, 2023, 624(7990): 80--85

\bibitem{bran2023augmenting}
Bran A~M, Cox S, Schilter O, Baldassari C, White A, Schwaller P.
\newblock Augmenting large language models with chemistry tools.
\newblock In: NeurIPS 2023 AI for Science Workshop.
\newblock 2023

\bibitem{liu2024agentbench}
Liu X, Yu~H, Zhang H, Xu~Y, Lei X, Lai H, Gu~Y, Ding H, Men K, Yang K, Zhang S, Deng X, Zeng A, Du~Z, Zhang C, Shen S, Zhang T, Su~Y, Sun H, Huang M, Dong Y, Tang J.
\newblock {AgentBench}: Evaluating {LLM}s as agents.
\newblock In: International Conference on Learning Representations.
\newblock 2024

\bibitem{janakarajan2023large}
Janakarajan N, Erdmann T, Swaminathan S, Laino T, Born J.
\newblock Large language models in molecular discovery.
\newblock In: NeurIPS 2023 AI for Science Workshop.
\newblock 2023

\bibitem{liu2024conversational}
Liu S, Wang J, Yang Y, Wang C, Liu L, Guo H, Xiao C.
\newblock Conversational drug editing using retrieval and domain feedback.
\newblock In: International Conference on Learning Representations.
\newblock 2024

\bibitem{zhang2023moleculegpt}
Zhang W, Wang X, Nie W, Eaton J, Rees B, Gu~Q.
\newblock Molecule{GPT}: Instruction following large language models for molecular property prediction.
\newblock In: NeurIPS 2023 Workshop on New Frontiers of AI for Drug Discovery and Development.
\newblock 2023

\bibitem{zheng2024relaxing}
Zheng Z, Liu Y, Li~J, Yao J, Rong Y.
\newblock Relaxing continuous constraints of equivariant graph neural networks for physical dynamics learning.
\newblock In: KDD.
\newblock 2024

\bibitem{liu2024equivariant}
Liu Y, Zheng Z, Rong Y, Li~J.
\newblock Equivariant graph learning for high-density crowd trajectories modeling.
\newblock Transactions on Machine Learning Research, 2024

\end{thebibliography}

\end{document}